\documentclass[lettersize,journal]{IEEEtran}

\usepackage{tikz}
\usepackage{pgfplots}
\pgfplotsset{compat=1.18}  
\usepgfplotslibrary{groupplots}

\usepackage{float}

\usepackage{amsmath}
\usepackage{amssymb}

\usepackage[colorlinks=true,allcolors=blue]{hyperref}

\usepackage{amsmath,amsfonts}
\usepackage{algorithmic}
\usepackage{algorithm}
\usepackage{array}
\usepackage[caption=false,font=normalsize,labelfont=sf,textfont=sf]{subfig}
\usepackage{textcomp}
\usepackage{stfloats}
\usepackage{url}
\usepackage{verbatim}
\usepackage{graphicx}
\usepackage{cite}
\usepackage{lettrine}

\usepackage{cite}
\usepackage{amsmath}
\usepackage{booktabs}
\usepackage{multirow}

\usepackage{tabularx}

\newcolumntype{P}[1]{>{\raggedright\arraybackslash}p{#1}}
\newcolumntype{C}[1]{>{\centering\arraybackslash}p{#1}}
\usepackage{tikz}
\usepackage[europeanresistors,americaninductors]{circuitikz}
\usepackage{xcolor}

\usepackage{tikz}
\usepackage{pgfplots}
\pgfplotsset{compat=1.17}
\usetikzlibrary{external}
\usepackage{float}



\begin{document}

\title{
%
Fractional Order Federated Learning for    Battery Electric Vehicle Energy Consumption Modeling

}
\author{
    Mohammad Partohaghighi,
    Roummel Marcia,
    Bruce J. West,
    and YangQuan Chen,~\IEEEmembership{Member,~IEEE}
    \thanks{Mohammad Partohaghighi is with the Electrical Engineering and Computer Science, University of California at Merced, Merced, CA 95343 USA (e-mail: mpartohaghighi@ucmerced.edu).}
    \thanks{Roummel Marcia is with the Department of Applied Mathematics, University of California at Merced, Merced, CA 95343 USA (e-mail: rmarcia@ucmerced.edu).}
    \thanks{Bruce J. West is with the Department of Innovation and Research, North Carolina State University, Raleigh, NC 27695 USA (e-mail: brucejwest213@gmail.com).}
    \thanks{YangQuan Chen is with the Mechatronics, Embedded Systems and Automation (MESA) Lab, Department of Mechanical Engineering, School of Engineering, University of California, Merced, CA 95343 USA (e-mail: ychen53@ucmerced.edu).}
    \thanks{This work was supported in part by the U.S. Department of Commerce under Grant BS123456.}
    \thanks{Corresponding author: YangQuan Chen (e-mail: ychen53@ucmerced.edu).}
}
\markboth{Journal of \LaTeX\ Class Files,~Vol.~14, No.~8, August~2021}%
{Partohaghighi \MakeLowercase{\textit{et al.}}: Roughness-Informed Federated Learning For Energy Consumption Modeling of Battery Electric Vehicles}
\IEEEpubid{0000--0000/00\$00.00~\copyright~2021 IEEE}

\maketitle

\makeatletter
\renewcommand{\addcontentsline}[3]{%
}
\makeatother

\begin{abstract}
Federated learning on connected electric vehicles (BEVs) faces severe instability due to intermittent connectivity,
time-varying client participation, and pronounced client-to-client variation induced by diverse operating conditions.
Conventional FedAvg and many advanced methods can suffer from excessive drift and degraded convergence under these realistic constraints.
This work introduces Fractional-Order Roughness-Informed Federated Averaging (FO-RI-FedAvg), a lightweight and modular extension of FedAvg
that improves stability through two complementary client-side mechanisms: (i) adaptive roughness-informed proximal regularization, which
dynamically tunes the pull toward the global model based on local loss-landscape roughness, and (ii) non-integer-order local optimization,
which incorporates short-term memory to smooth conflicting update directions. The approach preserves standard FedAvg server aggregation,
adds only element-wise operations with amortizable overhead, and allows independent toggling of each component.
Experiments on two real-world BEV energy prediction datasets, VED and its extended version eVED, show that FO-RI-FedAvg achieves
improved accuracy and more stable convergence compared to strong federated baselines, particularly under reduced client participation.
\end{abstract}


\section{Introduction}
\label{sec:introduction}

Federated learning (FL) enables collaborative model training across a population of clients while keeping raw data local, a design that has become central to privacy-sensitive, large-scale learning on edge devices \cite{mcmahan2017fedavg,konecny2016federatedopt,kairouz2021advances,hard2018fedkeyboard}. In many real deployments, however, the clients are heterogeneous (in hardware, connectivity, environments, and behavior), participation is intermittent, and local datasets often deviate from classical independent and identically distributed assumptions, which fundamentally changes the optimization regime compared with centralized training \cite{zhao2018noniid,hsu2019measuring,kairouz2021advances}. Connected battery electric vehicles (BEVs) provide a canonical example: fleets generate rich telemetry (speed/acceleration profiles, grade, temperature, auxiliary loads, battery state, driver behavior), yet this data is privacy- and policy-constrained, often regulated or contractually protected, motivating on-vehicle learning with privacy-preserving aggregation \cite{bonawitz2017secureagg,gdpr2016,ccpa2018}. At the same time, BEVs operate under diverse vehicle variants, climates, road networks, and driving styles, yielding pronounced client-to-client variation in data distributions and learning dynamics \cite{yavasoglu2019range,zhang2020apenergy,petkevicius2021probabilistic,ullah2022evenergy}. Energy consumption modeling in particular is not merely a supervised task; it is a sequential, context-dependent system where long-range temporal effects (traffic regimes, repeated commute patterns, seasonal temperature, degradation) materially influence predictive accuracy and downstream decisions (range estimation, eco-routing, charging planning) \cite{yavasoglu2019range,zhang2020apenergy,huang2025sequenceapenergy,topic2019energies}.

Despite its simplicity and practical success, the Federated Averaging (FedAvg) paradigm is known to struggle under realistic client variability and intermittent participation \cite{mcmahan2017fedavg,kairouz2021advances}. When each client performs multiple local steps on a drifting local objective, the aggregated update can become biased toward frequently sampled or high-variance clients, leading to \emph{client drift}, slow convergence, and instability \cite{zhao2018noniid,hsu2019measuring,khaled2020tighterlocalsgd}. A substantial line of work proposes mechanisms to mitigate these effects, including proximal regularization \cite{li2018fedprox2}, control variates to reduce drift \cite{karimireddy2020scaffold}, normalized update schemes addressing objective inconsistency \cite{wang2020fednova}, adaptive server-side optimizers \cite{reddi2021fedopt}, and dynamic regularization strategies \cite{acar2021feddyn}. While these methods reduce some pathologies, two gaps remain salient for connected BEVs and similarly “messy” cyber-physical fleets: (i) many approaches still rely on \emph{integer-order} local optimization dynamics that are effectively short-memory (dominated by recent gradients), and (ii) they often lack an explicit, \emph{client-specific} indicator of when local optimization is entering a highly irregular regime where aggressive updates amplify divergence.

\paragraph{Integer-order vs.\ non-integer-order FL (terminology).}
Throughout this paper, we use \emph{integer-order} FL to refer to FedAvg-style training in which clients perform classical first-order local optimization (e.g., SGD), yielding effectively short-memory update dynamics.
In contrast, \emph{non-integer-order} FL in this work refers to our proposed FO-RI-FedAvg, which preserves the same federated protocol and server-side aggregation, but replaces standard client-side SGD with \emph{fractional-order} (i.e., non-integer-order) updates parameterized by an order \(\alpha\).
Thus, the distinction is entirely \emph{client-side}: we modify the local optimizer dynamics while keeping server aggregation unchanged, and recover the integer-order baseline as \(\alpha\to 1\).

This paper argues that BEV energy-consumption modeling makes these gaps unavoidable. Vehicle energy use is influenced by latent and slowly varying factors (battery aging, tire pressure, seasonal auxiliary load usage, driving style persistence), and thus learning benefits from \emph{long-memory} mechanisms that can integrate historical information in a principled way \cite{zhang2020apenergy,petkevicius2021probabilistic,huang2025sequenceapenergy}. Fractional calculus provides such a mechanism: fractional-order dynamics generalize classical differentiation to non-integer orders and naturally encode history dependence (“memory”) through nonlocal operators \cite{podlubny1999fractional,diethelm2010analysis,ortigueira2011fractional,monje2010fractionalcontrol}. Recent work has begun to explore fractional-order gradients and fractionalized optimization as practical algorithms with distinct implicit bias and convergence behavior \cite{shin2023fractional,yang2023isfogd}. In this work, we focus on the practically important regime \(0<\alpha\le 1\), where \(\alpha\) controls the strength of memory: smaller \(\alpha\) increases the influence of historical information, while \(\alpha\to 1\) recovers classical first-order behavior \cite{podlubny1999fractional,diethelm2010analysis,shin2023fractional}.

Long-memory alone is not sufficient in federated systems, because client-to-client variability also manifests through \emph{geometry}: different clients may experience qualitatively different local loss landscapes (flat vs.\ sharp, smooth vs.\ jagged), and the same nominal learning rate can be benign on one client and destabilizing on another. Modern generalization theory and practice have repeatedly shown that loss-landscape geometry (sharpness/flatness and related notions) influences stability and generalization \cite{hochreiter1997flat,keskar2017largebatch,dinh2017sharpgen}. Sharpness-aware training procedures further emphasize that geometry-aware updates can materially improve robustness \cite{foret2021sam,andriushchenko2022sam}. Complementarily, recent work proposes operational measures of landscape irregularity (``roughness'') that can be estimated from local loss probes and used as a diagnostic of optimization difficulty \cite{wu2021roughnessindex}. These insights suggest a natural federated question: can we \emph{control} client drift by adapting the strength of local regularization and update aggressiveness based on a per-client geometric indicator?

We answer this question by proposing \textbf{FO-RI-FedAvg} (\textbf{F}ractional-\textbf{O}rder \textbf{R}oughness-\textbf{I}nformed Federated Averaging), which integrates two ideas into a single, reproducible federated optimizer. First, each client performs local learning using a fractional-order update with order \(\alpha\in(0,1]\), injecting a tunable long-memory effect into the local dynamics \cite{podlubny1999fractional,diethelm2010analysis,shin2023fractional}. Second, each client computes a \emph{roughness index} \(\mathcal{I}_k\) (a scalar summary of local landscape irregularity) and uses it to modulate the strength of a drift-controlling regularizer and/or update damping, yielding \emph{client-adaptive} stability control rather than a single global setting \cite{wu2021roughnessindex,li2018fedprox2,acar2021feddyn}. Conceptually, FO-RI-FedAvg couples \emph{memory-aware} optimization (fractional order) with \emph{geometry-aware} control (roughness-informed regularization), targeting temporally dependent and participation-constrained regimes exemplified by connected BEV energy modeling \cite{yavasoglu2019range,zhang2020apenergy,petkevicius2021probabilistic,huang2025sequenceapenergy}.

\paragraph{Contributions.}
The main contributions of this work are:
\begin{itemize}
    \item \textbf{Algorithmic:} We introduce FO-RI-FedAvg, a new federated optimization method that combines fractional-order local dynamics (\(\alpha\in(0,1]\)) with roughness-informed client-specific stability control, designed for federated BEV learning under intermittent participation and realistic client variability \cite{mcmahan2017fedavg,li2018fedprox2,wu2021roughnessindex,shin2023fractional}.
    \item \textbf{Mechanistic:} We operationalize client roughness \(\mathcal{I}_k\) as a practical proxy for local landscape irregularity, bridging loss-geometry insights (sharpness/flatness) to federated drift mitigation \cite{hochreiter1997flat,keskar2017largebatch,foret2021sam,wu2021roughnessindex}.
    \item \textbf{Application:} We evaluate FO-RI-FedAvg on energy-consumption modeling for connected BEVs using VED and eVED, a domain that jointly exhibits privacy constraints, diverse operating conditions, and long-range temporal dependence \cite{bonawitz2017secureagg,gdpr2016,ccpa2018,zhang2020apenergy,petkevicius2021probabilistic,yavasoglu2019range,ullah2022evenergy,huang2025sequenceapenergy}.
\end{itemize}

\paragraph{Organization.}
Section~\ref{sec:methodology} formalizes the federated setting and presents FO-RI-FedAvg in detail, including the fractional-order client update, roughness estimation, and the server aggregation/control loop. Section~\ref{sec:system} provides the system framework for BEV energy consumption modeling. Section~\ref{experiments} reports empirical results with a focus on BEV energy-consumption prediction under intermittent participation and practical federated constraints. Limitations and future work can be found in Section~\ref{sec:limitations_futurework}. Finally, the conclusion section can be found in Section~\ref{sec:conclusion}.

\section{Methodology}
\label{sec:methodology}

This section formalizes the federated learning setting and introduces \textbf{FO-RI-FedAvg}
(\emph{Fractional-Order Roughness-Informed Federated Averaging}), a method that combines
(i) \emph{fractional-order} local update dynamics (defined in subsection~\ref{subsec:method_fractional}), which induces a lightweight
memory effect in the optimization trajectory \cite{samko1993fractional,caputo1967linear,wang2017fractionalbp,chen2017fogm},
and (ii) \emph{roughness-informed} control to mitigate client drift and improve stability under intermittent participation
and client-to-client variation, using client-side geometry probes inspired by loss-landscape and curvature diagnostics.
We additionally define an optional \emph{spectral stability} diagnostic that can be used to gate the same control loop
with minimal overhead.

\subsection{Federated learning setup and notation}
\label{subsec:method_setup}

We consider a standard federated learning (FL) system with a central server coordinating \(K\) clients
under partial participation \cite{bonawitz2019scale,li2020flspm,stich2019localsgd,haddadpour2019localdescent}.
Client \(k\in\{1,\dots,K\}\) holds a local dataset \(\mathcal{P}_k\) of size \(n_k=|\mathcal{P}_k|\),
and the total sample size is \(n=\sum_{k=1}^K n_k\).
Let \(\mathbf{w}\in\mathbb{R}^d\) denote the model parameters.

\paragraph{Local and global objectives.}
Each client minimizes a local empirical risk
\begin{equation}
F_k(\mathbf{w})
\triangleq
\frac{1}{n_k}\sum_{(x_i,y_i)\in \mathcal{P}_k}\ell(\mathbf{w};x_i,y_i),
\label{eq:local_obj}
\end{equation}
and the global FL objective is the data-weighted average
\begin{equation}
f(\mathbf{w})
\triangleq
\sum_{k=1}^{K}\frac{n_k}{n}F_k(\mathbf{w}).
\label{eq:global_obj}
\end{equation}

\paragraph{Communication rounds and partial participation.}
Training proceeds for \(T\) communication rounds indexed by \(t\in\{0,\dots,T-1\}\).
At each round, the server samples a subset \(S_t\subseteq\{1,\dots,K\}\) of participating clients,
with \(|S_t|=\max(\lceil C K\rceil,1)\), where \(C\in(0,1]\) is the participation fraction.
The server broadcasts the current global model \(\mathbf{w}_t\) to all \(k\in S_t\).
Each participating client performs local training for \(E\) local epochs (equivalently \(H\) local steps),
using mini-batches of size \(B\), and returns an updated model \(\mathbf{w}_{t+1}^k\) to the server.
The server aggregates by the standard data-weighted average \cite{bonawitz2019scale,li2020flspm}:
\begin{equation}
\mathbf{w}_{t+1}
=
\sum_{k\in S_t}\frac{n_k}{n_t}\mathbf{w}_{t+1}^k,
\qquad
n_t\triangleq \sum_{k\in S_t} n_k.
\label{eq:fedavg_agg}
\end{equation}

\paragraph{Client variability model.}
To capture client-to-client variation in local objectives, we use the bounded gradient dissimilarity condition:
there exists \(\zeta\ge 0\) such that for all \(\mathbf{w}\),
\begin{equation}
\frac{1}{K}\sum_{k=1}^{K}\left\|\nabla F_k(\mathbf{w})-\nabla f(\mathbf{w})\right\|_2^2
\le \zeta^2,
\label{eq:grad_dissim}
\end{equation}
which is widely used in theoretical analyses of local-update methods when client objectives deviate from the global objective
\cite{stich2019localsgd,haddadpour2019localdescent}.
We also assume each \(F_k\) is \(L\)-smooth (i.e., its gradient is \(L\)-Lipschitz), and that stochastic gradients have bounded variance.


\subsection{Fractional-order local update dynamics}
\label{subsec:method_fractional}

\paragraph{Motivation and scope.}
Fractional calculus generalizes integer-order differentiation and induces nonlocal (history-dependent) dynamics
\cite{samko1993fractional,caputo1967linear}.
Such operators model hereditary effects via power-law memory kernels, and have been explored as a mechanism for stabilizing
optimization in ill-conditioned or nonconvex regimes \cite{wang2017fractionalbp,chen2017fogm}.
In this work, we introduce a \emph{scalable} fractional-order modification for client-side training that:
(i) remains compatible with first-order federated optimization,
(ii) adds only lightweight constant-overhead element-wise operations per local step, and
(iii) reduces \emph{exactly} to standard SGD when $\alpha=1$ (Eqs.~\eqref{eq:frac_precond}--\eqref{eq:fo_update_generic}).
Here, $\alpha\in(0,1]$ denotes the fractional order used in the preconditioner \eqref{eq:frac_precond},
and $\mathcal{I}_k$ is the client-wise roughness index defined in \eqref{eq:ri_def}.
When integrated into FO-RI-FedAvg, setting $\alpha=1$ recovers the non-fractional roughness-informed variant (via \eqref{eq:fo_ri_update}),
and disabling the roughness control by setting $r(\mathcal{I}_k)=0$ in \eqref{eq:prox_obj} yields the FedAvg baseline.

\paragraph{Local update notation.}
Let $\mathbf{w}_{t,h}^k\in\mathbb{R}^d$ denote the client-$k$ local parameters after $h$ local steps within round $t$,
with $\mathbf{w}_{t,0}^k=\mathbf{w}_t$ being the received global model.
Let $b_{t,h}^k\subset \mathcal{P}_k$ be the mini-batch at step $h$ and define the stochastic gradient
\begin{equation}
\mathbf{g}_{t,h}^k
\triangleq
\nabla_{\mathbf{w}}\ell\!\left(\mathbf{w}_{t,h}^k; b_{t,h}^k\right).
\label{eq:stoch_grad}
\end{equation}

\paragraph{Standard local SGD (baseline).}
The standard client-side SGD update used in FedAvg is
\begin{equation}
\mathbf{w}_{t,h+1}^k
=
\mathbf{w}_{t,h}^k
-
\eta_t\,\mathbf{g}_{t,h}^k,
\label{eq:local_sgd}
\end{equation}
where $\eta_t>0$ is the (round-level) learning rate.

\paragraph{Fractional-order local SGD (FO-SGD): Caputo-inspired surrogate.}
We motivate our design from the \emph{Caputo fractional derivative}.
For a scalar function $f(t)$, its Caputo derivative of order $\alpha$ is
\begin{equation}
{}^C D_{t}^{\alpha} f(t)
=
\frac{1}{\Gamma(n-\alpha)}
\int_{0}^{t}
  \frac{
    f^{(n)}(\tau)
  }{
    (t-\tau)^{\alpha-n+1}
  }
\, d\tau,
\quad
n-1 < \alpha < n,
\label{eq:caputo_general}
\end{equation}
where $n$ is the smallest integer greater than $\alpha$, $\Gamma(\cdot)$ is the Gamma function,
and $\tau$ is the integration variable.
Equation~\eqref{eq:caputo_general} highlights the \emph{nonlocality} of fractional dynamics via a power-law weighted history.

\paragraph{Restriction to $n=1$ (i.e., $0<\alpha\le 1$).}
Although \eqref{eq:caputo_general} holds for general $n$, we restrict to $n=1$ so that $0<\alpha\le 1$.
This regime preserves a gradient-like descent structure and reduces exactly to standard SGD/FedAvg at $\alpha=1$.
In contrast, $n\ge 2$ (e.g., $1<\alpha\le 2$) introduces higher-order inertial dynamics and typically requires additional state variables
and tighter stability control; moreover, it would involve higher classical derivatives not directly accessible in first-order stochastic training.
Accordingly, we adopt $n=1$ as a scalable fractional-order extension and leave $n\ge2$ as future work.

Setting $n=1$ yields the specialized Caputo form
\begin{equation}
{}^{C}D_{t}^{\alpha} f(t)
=
\frac{1}{\Gamma(1-\alpha)}
\int_{0}^{t}
\frac{f'(\tau)}{(t-\tau)^{\alpha}}
\, d\tau,
\qquad 0<\alpha\le 1.
\label{eq:caputo_alpha_01}
\end{equation}

\paragraph{Leading-order scaling and discrete surrogate.}
A common implementable route uses a fractional Taylor-series expansion around $t_0$:
\begin{equation}
D_{t}^{\alpha} f(t)
=
\sum_{i=1}^{\infty}
\frac{f^{(i)}(t_0)}{\Gamma(i+1-\alpha)}
(t-t_0)^{i-\alpha},
\qquad 0<\alpha\le 1.
\label{eq:frac_taylor_series}
\end{equation}
Retaining the leading term ($i=1$) yields the characteristic factor $1/\Gamma(2-\alpha)$ and a lag dependence proportional to $(t-t_0)^{1-\alpha}$.
Motivated by this scaling, we inject a fractional-order \emph{effect} into SGD through a constant-overhead discrete surrogate.

\paragraph{Discrete fractional preconditioning (implementable FO-SGD proxy).}
We define an element-wise \emph{fractional preconditioner} from the most recent parameter displacement with fractional order $ 0<\alpha\le 1,$:
\begin{equation}
\mathbf{p}_{t,h}^k
\triangleq
\frac{1}{\Gamma(2-\alpha)}
\left(\left|\mathbf{w}_{t,h}^k-\mathbf{w}_{t,h-1}^k\right|+\delta\mathbf{1}\right)^{\odot(1-\alpha)},
\qquad
\label{eq:frac_precond}
\end{equation}
where $\delta>0$ is a numerical stabilizer,
$\mathbf{1}$ is the all-ones vector, and $\odot$ denotes element-wise operations.

The resulting FO-SGD local update is
\begin{equation}
\mathbf{w}_{t,h+1}^k
=
\mathbf{w}_{t,h}^k
-
\eta_t
\left(\mathbf{g}_{t,h}^k \odot \mathbf{p}_{t,h}^k\right).
\label{eq:fo_update_generic}
\end{equation}

\paragraph{Interpretation (fractional memory effect).}
Equation~\eqref{eq:frac_precond} should be interpreted as a \emph{Caputo-inspired surrogate} rather than an exact evaluation of
\eqref{eq:caputo_alpha_01}. Specifically, we map the lag term $(t-t_0)^{1-\alpha}$ to a discrete displacement proxy,
replacing $(t-t_0)$ by the most recent coordinate-wise parameter change
$\left|\mathbf{w}_{t,h}^k-\mathbf{w}_{t,h-1}^k\right|$.
This yields a constant-overhead mechanism that injects a fractional-order memory \emph{effect} into first-order stochastic updates.

\paragraph{Reduction to standard SGD.}
When $\alpha=1$, $\Gamma(1)=1$ and $(|\Delta|+\delta)^{0}=\mathbf{1}$ element-wise, hence
$\mathbf{p}_{t,h}^k=\mathbf{1}$ and \eqref{eq:fo_update_generic} reduces exactly to the baseline SGD update \eqref{eq:local_sgd}.

\paragraph{Initialization at the first local step.}
Since $\mathbf{w}_{t,h-1}^k$ is undefined at $h=0$, we set $\mathbf{p}_{t,0}^k=\mathbf{1}$ and apply the fractional preconditioner for $h\ge 1$,
so that the first local step coincides with standard SGD and numerical sensitivity to $\delta$ is avoided.

\paragraph{Practical safeguards.}
To avoid extreme coordinate-wise scaling when $|\mathbf{w}_{t,h}^k-\mathbf{w}_{t,h-1}^k|$ is very small or very large,
we optionally clip the preconditioner:
\begin{equation}
\mathbf{p}_{t,h}^k
\leftarrow
\mathrm{clip}\!\left(\mathbf{p}_{t,h}^k;\; p_{\min},p_{\max}\right),
\label{eq:precond_clip}
\end{equation}
where $\mathrm{clip}(\cdot)$ is applied element-wise and $0<p_{\min}\le p_{\max}$.

\paragraph{Computational cost.}
The preconditioner in \eqref{eq:frac_precond} adds only $O(d)$ element-wise operations per local step and requires storing one previous local iterate
$\mathbf{w}_{t,h-1}^k$ (i.e., $O(d)$ additional memory), preserving the scalability of standard FL training.
\subsection{Roughness and spectral diagnostics for stability-aware control}
\label{subsec:method_diagnostics}

FO-RI-FedAvg modulates local training using client-specific stability signals.
We define (i) a \emph{geometric roughness index} computed from local loss landscape slices,
motivated by empirical and theoretical evidence that loss geometry/curvature relates to optimization stability
\cite{li2018losslandscape,ghorbani2019hessian},
and (ii) an optional \emph{spectral stability} indicator derived from client weights \cite{martin2021implicit}.

\subsubsection{Geometric roughness index (client-wise)}
\label{subsubsec:roughness_index}

Fix a round \(t\) and client \(k\in S_t\).
Let \(\mathbf{w}_t\) be the broadcast model. We probe the local objective \(F_k\) in a neighborhood of \(\mathbf{w}_t\)
along \(M\) random directions.
For \(i\in\{1,\dots,M\}\), sample \(\mathbf{d}_i\sim\mathcal{N}(\mathbf{0},\mathbf{I}_d)\) and normalize
\begin{equation}
\mathbf{d}_i \leftarrow \frac{\mathbf{d}_i}{\|\mathbf{d}_i\|_2}.
\label{eq:dir_norm}
\end{equation}
Define a 1D slice over \(s\in[-\ell,\ell]\) by
\begin{equation}
\varphi_i(s)\triangleq F_k(\mathbf{w}_t+s\mathbf{d}_i),
\label{eq:slice_def}
\end{equation}
where \(\ell>0\) sets the probe radius.

We discretize \(s\) using \(m+1\) evenly spaced points
\(s_j=-\ell + j\cdot \frac{2\ell}{m}\) for \(j=0,\dots,m\).
The (discrete) total variation of \(\varphi_i\) is
\begin{equation}
\mathrm{TV}(\varphi_i)
\triangleq
\sum_{j=0}^{m-1}\left|\varphi_i(s_{j+1})-\varphi_i(s_j)\right|.
\label{eq:tv_def}
\end{equation}
Let the amplitude on the slice be
\begin{equation}
A_i \triangleq \max_{0\le j\le m}\varphi_i(s_j)-\min_{0\le j\le m}\varphi_i(s_j),
\label{eq:amp_def}
\end{equation}
and define the normalized total variation
\begin{equation}
T_i \triangleq \frac{1}{2\ell}\cdot\frac{\mathrm{TV}(\varphi_i)}{A_i+\epsilon_A},
\label{eq:ntv_def}
\end{equation}
where \(\epsilon_A>0\) prevents division by zero.

Finally, the client roughness index is the coefficient-of-variation statistic
\begin{equation}
\mathcal{I}_k
\triangleq
\frac{\mathrm{std}(\{T_i\}_{i=1}^{M})}{\mathrm{mean}(\{T_i\}_{i=1}^{M})+\epsilon_T},
\label{eq:ri_def}
\end{equation}
where \(\epsilon_T>0\) is a small stabilizer.
Larger \(\mathcal{I}_k\) indicates a more oscillatory local landscape around \(\mathbf{w}_t\),
which is consistent with harder-to-optimize, more irregular local geometry \cite{li2018losslandscape,ghorbani2019hessian}.

\paragraph{Practical estimator.}
Computing \(\varphi_i(s_j)\) exactly can be expensive; in practice we estimate it using a fixed probe mini-batch
\(b^{k}_{\mathrm{probe}}\subset \mathcal{P}_k\) of size \(B_{\mathrm{probe}}\), i.e.,
\(\varphi_i(s)\approx \ell(\mathbf{w}_t+s\mathbf{d}_i; b^{k}_{\mathrm{probe}})\).
This yields a low-overhead per-round diagnostic.

\subsubsection{Optional spectral stability indicator (client-wise)}
\label{subsubsec:spectral_indicator}

To complement geometric roughness, we optionally compute a lightweight spectral diagnostic on a designated weight matrix
\(\mathbf{W}(\mathbf{w})\) extracted from the model (e.g., the final linear layer).
Following weight-matrix spectral regularization perspectives \cite{martin2021implicit}, we define the \emph{spectral flatness ratio}
\begin{equation}
\kappa_k
\triangleq
\frac{\|\mathbf{W}(\mathbf{w}_t)\|_2}{\|\mathbf{W}(\mathbf{w}_t)\|_F+\epsilon_F},
\label{eq:spectral_flatness}
\end{equation}
where \(\|\cdot\|_2\) is the spectral norm, \(\|\cdot\|_F\) is the Frobenius norm,
and \(\epsilon_F>0\) stabilizes the denominator.
A larger \(\kappa_k\) indicates stronger anisotropy (energy concentrated in a few singular directions),
which we use as a conservative proxy for update-direction fragility.

In practice, \(\|\mathbf{W}\|_2\) can be estimated with a small fixed number of power iterations,
while \(\|\mathbf{W}\|_F\) is computed in a single pass.

\subsection{FO-RI-FedAvg: integrating fractional dynamics with roughness-informed control}
\label{subsec:method_forifedavg}

FO-RI-FedAvg modifies local training in two coupled ways:
(i) it adds a \emph{roughness-controlled proximal regularizer} to reduce local-update drift under heterogeneity,
and (ii) it replaces standard SGD steps with the fractional-order update \eqref{eq:fo_update_generic}--\eqref{eq:frac_precond}.

\subsubsection{Roughness-controlled proximal objective}
\label{subsubsec:prox_objective}

At round \(t\), client \(k\in S_t\) minimizes a regularized objective
\begin{equation}
\widetilde{F}_{k,t}(\mathbf{w})
\triangleq
F_k(\mathbf{w})
+
\frac{\lambda_t\, r(\mathcal{I}_k)}{2}\,\|\mathbf{w}-\mathbf{w}_t\|_2^2,
\label{eq:prox_obj}
\end{equation}
where \(\lambda_t>0\) is a server-chosen base regularization strength and
\(r(\mathcal{I}_k)\ge 0\) is a monotone roughness response function.
A simple choice is \(r(\mathcal{I}_k)=\mathrm{clip}(\mathcal{I}_k; I_{\min}, I_{\max})\),
or a smooth saturating map such as
\begin{equation}
r(\mathcal{I}_k)=\frac{\mathcal{I}_k}{\mathcal{I}_k+\tau_I},
\label{eq:ri_response}
\end{equation}
with \(\tau_I>0\).
This increases the proximal pull toward \(\mathbf{w}_t\) for rough clients, mitigating drift.

The stochastic gradient of \(\widetilde{F}_{k,t}\) at \(\mathbf{w}_{t,h}^k\) is
\begin{equation}
\widetilde{\mathbf{g}}_{t,h}^k
\triangleq
\nabla_{\mathbf{w}}\ell(\mathbf{w}_{t,h}^k; b_{t,h}^k)
+
\lambda_t\, r(\mathcal{I}_k)\,(\mathbf{w}_{t,h}^k-\mathbf{w}_t).
\label{eq:prox_grad}
\end{equation}

\subsubsection{Fractional-order roughness-informed local step}
\label{subsubsec:fo_ri_step}

FO-RI-FedAvg updates local parameters via
\begin{equation}
\mathbf{w}_{t,h+1}^k
=
\mathbf{w}_{t,h}^k
-
\eta_t
\left(\widetilde{\mathbf{g}}_{t,h}^k \odot \mathbf{p}_{t,h}^k\right),
\label{eq:fo_ri_update}
\end{equation}
with \(\mathbf{p}_{t,h}^k\) defined in \eqref{eq:frac_precond}.
Optionally, we incorporate the spectral indicator by gating the preconditioner:
\begin{equation}
\mathbf{p}_{t,h}^k
\leftarrow
\frac{1}{1+\beta_\kappa \kappa_k}\,\mathbf{p}_{t,h}^k,
\label{eq:spectral_gate}
\end{equation}
where \(\beta_\kappa\ge 0\) controls the strength of spectral damping.
Setting \(\beta_\kappa=0\) disables spectral gating.

\paragraph{Learning rate schedule.}
We use either a constant \(\eta_t=\eta\) or a diminishing schedule such as
\(\eta_t=\eta_0/\sqrt{t+1}\), which is standard for stabilizing stochastic optimization.

\paragraph{Ablations and reductions.}
FO-RI-FedAvg reduces to:
(i) a roughness-informed proximal local-SGD variant when \(\alpha=1\) (since \(\mathbf{p}_{t,h}^k=\mathbf{1}\)),
(ii) a fractional local-update FL variant when \(r(\mathcal{I}_k)=0\),
and (iii) a standard local-SGD FL baseline when \(\alpha=1\) and \(r(\mathcal{I}_k)=0\) \cite{stich2019localsgd,haddadpour2019localdescent}.

\subsection{Algorithm specification}
\label{subsec:method_algorithm}

\begin{figure}[!t]
\centering
\includegraphics[width=3.5in]{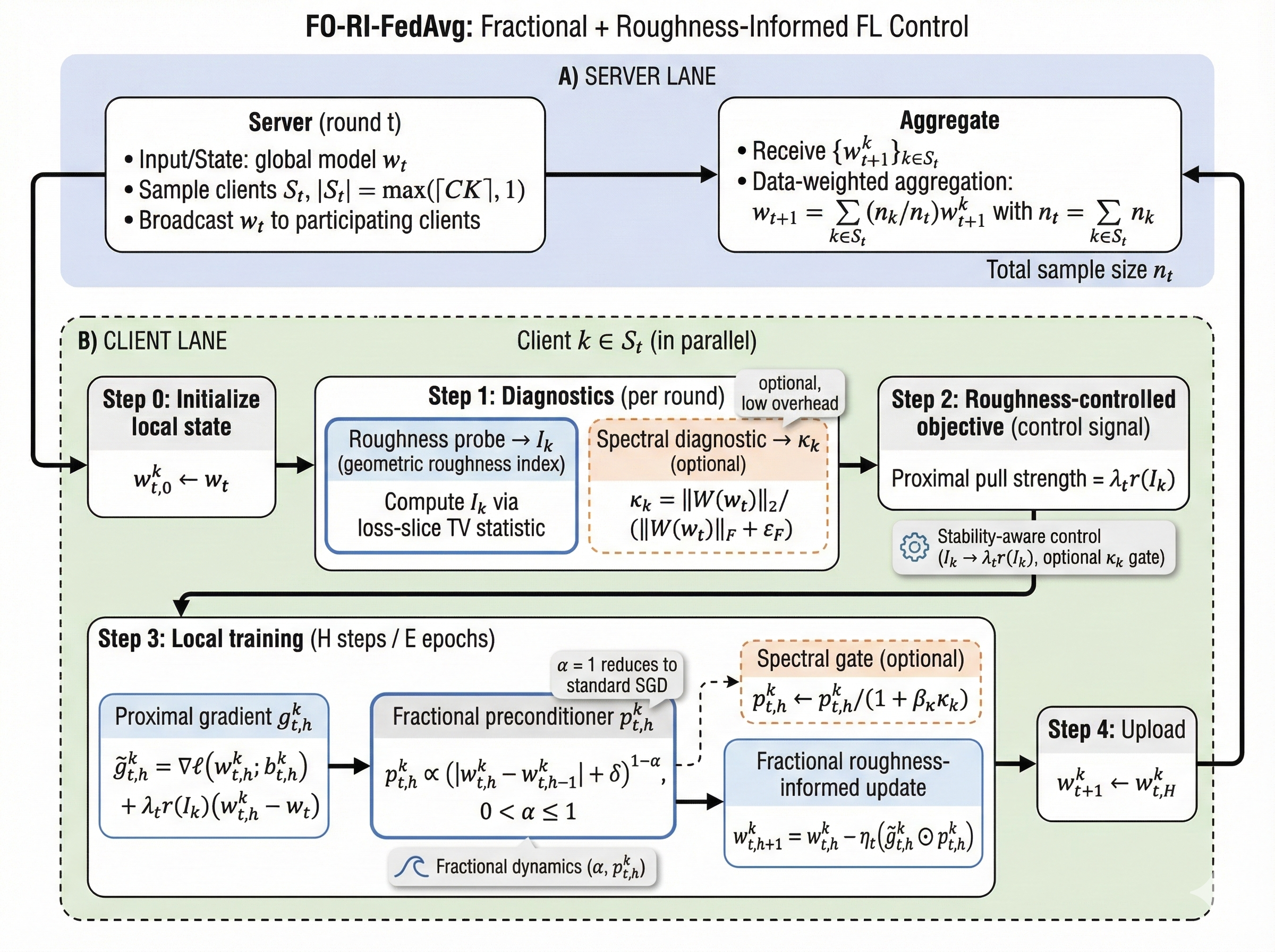}
\caption{\textbf{Pipeline of FO-RI-FedAvg (Fractional-Order Roughness-Informed Federated Averaging).}
}\label{fig:fo_ri_fedavg}
\end{figure}

Algorithm~\ref{alg:fo_ri_fedavg} and Figure~\ref{fig:fo_ri_fedavg} provide the full FO-RI-FedAvg procedure, including
(i) client diagnostics, (ii) fractional roughness-informed local updates, and (iii) server aggregation.

\begin{algorithm}[t]
\caption{FO-RI-FedAvg (Fractional-Order Roughness-Informed Federated Averaging)}
\label{alg:fo_ri_fedavg}
\begin{algorithmic}[1]
\STATE \textbf{Input:} $K$ clients; rounds $T$; participation fraction $C$; init. $\mathbf{w}_0\in\mathbb{R}^d$;
local steps $H$ (or epochs $E$); mini-batch size $B$;
fractional order $\alpha\in(0,1]$; stabilizer $\delta>0$;
schedules $\{\eta_t\}_{t=0}^{T-1}$, $\{\lambda_t\}_{t=0}^{T-1}$;
roughness params $(M,\ell,m,B_{\mathrm{probe}})$; stabilizers $(\epsilon_A,\epsilon_T)$;
optional spectral gate $\beta_\kappa\ge 0$, $\epsilon_F>0$.
\STATE \textbf{Output:} Global model $\mathbf{w}_T$.

\FOR{$t=0$ \TO $T-1$}
    \STATE $S_t \gets$ random subset of $\{1,\dots,K\}$ with $|S_t|=\max(\lceil CK\rceil,1)$
    \STATE Broadcast $\mathbf{w}_t$ to all $k\in S_t$
    \FOR{\textbf{each} $k\in S_t$ \textbf{in parallel}}
        \STATE $\mathbf{w}_{t,k} \gets \mathbf{w}_t$
        \STATE $b_{\mathrm{probe}}^k \gets$ sample from $\mathcal{P}_k$ with $|b_{\mathrm{probe}}^k|=B_{\mathrm{probe}}$
        \STATE $\mathcal{I}_k \gets$ compute roughness index via Eqs.~\eqref{eq:dir_norm}--\eqref{eq:ri_def} on $b_{\mathrm{probe}}^k$
        \IF{$\beta_\kappa>0$}
            \STATE $\kappa_k \gets$ compute spectral flatness via Eq.~\eqref{eq:spectral_flatness} on $\mathbf{W}(\mathbf{w}_t)$
        \ELSE
            \STATE $\kappa_k \gets 0$
        \ENDIF

        \STATE \emph{Initialize (first local step, standard proximal SGD):}
        \STATE $b_{t,0}^k \gets$ sample from $D_k$ with $|b_{t,0}^k|=B$
        \STATE $\widetilde{\mathbf{g}}_{t,0}^k \gets$ proximal gradient via Eq.~\eqref{eq:prox_grad}
        \STATE $\mathbf{w}_{t,k} \gets \mathbf{w}_{t,k} - \eta_t\,\widetilde{\mathbf{g}}_{t,0}^k$

        \FOR{$h=1$ \TO $H-1$}
            \STATE $b_{t,h}^k \gets$ sample from $D_k$ with $|b_{t,h}^k|=B$
            \STATE $\widetilde{\mathbf{g}}_{t,h}^k \gets$ proximal gradient via Eq.~\eqref{eq:prox_grad}
            \STATE $\mathbf{p}_{t,h}^k \gets$ fractional preconditioner via Eq.~\eqref{eq:frac_precond}
            \IF{$\beta_\kappa>0$}
                \STATE $\mathbf{p}_{t,h}^k \gets \mathbf{p}_{t,h}^k /(1+\beta_\kappa\kappa_k)$ \COMMENT{Eq.~\eqref{eq:spectral_gate}}
            \ENDIF
            \STATE $\mathbf{p}_{t,h}^k \gets$ clip$(\mathbf{p}_{t,h}^k)$ \COMMENT{optional, Eq.~\eqref{eq:precond_clip}}
            \STATE $\mathbf{w}_{t,k} \gets \mathbf{w}_{t,k} - \eta_t\left(\widetilde{\mathbf{g}}_{t,h}^k \odot \mathbf{p}_{t,h}^k\right)$ \COMMENT{Eq.~\eqref{eq:fo_ri_update}}
        \ENDFOR
        \STATE $\mathbf{w}_{t+1,k} \gets \mathbf{w}_{t,k}$
    \ENDFOR

    \STATE $n_t \gets \sum_{k\in S_t} n_k$
    \STATE $\mathbf{w}_{t+1} \gets \sum_{k\in S_t} (n_k/n_t)\,\mathbf{w}_{t+1,k}$ \COMMENT{Eq.~\eqref{eq:fedavg_agg}}
\ENDFOR
\STATE \textbf{return} $\mathbf{w}_T$
\end{algorithmic}
\end{algorithm}

\subsection{Computational and communication complexity}
\label{subsec:method_complexity}

\paragraph{Communication.}
FO-RI-FedAvg preserves the standard FL communication pattern: each selected client transmits one model vector
\(\mathbf{w}_{t+1}^k\in\mathbb{R}^d\) and receives \(\mathbf{w}_t\), yielding \(O(d)\) communication per participating client per round
\cite{bonawitz2019scale,li2020flspm}. Secure aggregation can be used as an orthogonal system layer if needed;
for dropout-robust secure aggregation with near-linear scaling, one option is Turbo-Aggregate \cite{so2021turboaggregate}.

\paragraph{Local compute (training).}
Local training cost is essentially that of local SGD: \(H\) stochastic gradient evaluations plus element-wise vector ops.
The fractional update adds only \(O(d)\) element-wise operations per step (absolute value, power, and multiplication).

\paragraph{Diagnostics overhead.}
Roughness estimation uses \(M(m+1)\) probe loss evaluations on a probe batch of size \(B_{\mathrm{probe}}\),
for cost \(O(M(m+1)\cdot \mathrm{cost\_forward}(B_{\mathrm{probe}}))\) per participating client per round
(or amortized if computed every \(R\) rounds).
The optional spectral diagnostic requires computing \(\|\mathbf{W}\|_F\) in one pass and estimating \(\|\mathbf{W}\|_2\)
via a small fixed number of power iterations, which is typically negligible relative to backprop for deep models \cite{martin2021implicit}.



\section{System Framework for BEV Energy Consumption Modeling}
\label{sec:system}

This section instantiates the federated learning formulation of Section~\ref{sec:methodology}
in a concrete cyber--physical setting: energy consumption modeling for a connected fleet of
battery electric vehicles (BEVs). We define (i) the system structure and privacy boundary,
(ii) the physical provenance of the energy signal used to generate supervised labels, and
(iii) the data organization, preprocessing, and windowed learning samples that form each
client dataset \(\mathcal{P}_k\). This section does \emph{not} re-derive FO-RI-FedAvg updates,
objectives, or convergence statements; those are fixed in Section~\ref{sec:methodology}.

\subsection{System Framework}
\label{subsec:SystemFramework}

\subsubsection{System Structure}
\label{subsubsec:SystemStructure}

\paragraph{Federated clients as vehicles.}
We consider a fleet of \(K\) connected BEVs, each acting as a federated client indexed by
\(k\in\{1,\dots,K\}\), consistent with Section~\ref{subsec:method_setup}.
Client \(k\) retains raw telemetry locally (e.g., detailed speed profiles and location traces)
and participates intermittently in communication rounds \(t\in\{0,\dots,T-1\}\) under partial
participation (fraction \(C\in(0,1]\)) as in Section~\ref{subsec:method_setup}.
Only model parameters/updates are transmitted; raw trip data never leaves the vehicle.
Secure aggregation can be used as an orthogonal protection layer when required \cite{bonawitz2017secureagg}.

\paragraph{Trip-level sensing and local retention.}
Client \(k\) logs \(Q_k\) trips indexed by \(q\in\{1,\dots,Q_k\}\).
Each trip is sampled at a fixed period \(\Delta\tau>0\). We denote continuous time within a trip
by \(\tau\) (to avoid collision with the FL round index \(t\)), and the corresponding discrete
sample index by \(r\in\{1,\dots,N_{k,q}\}\), where \(\tau_r \triangleq r\Delta\tau\).
At each sample \(r\), the vehicle records a raw telemetry vector
\(\mathbf{z}_{k,q}^{r}\in\mathbb{R}^{d_z}\) from onboard sensors and the BMS/CAN bus.
Typical channels include speed and acceleration, road grade/elevation proxies, ambient temperature,
auxiliary load proxies, and (when available) battery/drive signals such as pack power.

\paragraph{Feature construction (local).}
From \(\mathbf{z}_{k,q}^{r}\), the client constructs a feature vector
\(\mathbf{x}_{k,q}^{r}\in\mathbb{R}^{d_x}\) using only locally available signals and optional
locally computed enrichments. These features are not uploaded; the federated interface is only
through the learning objective in Eq.~\eqref{eq:local_obj} via samples \((x_i,y_i)\in\mathcal{P}_k\).

\subsubsection{Learning Target: Windowed Energy Consumption}
\label{subsubsec:LearningTarget}

\paragraph{Per-sample energy increment.}
The supervised learning target is derived from electrical energy consumption computed from pack power.
Let \(P_{\mathrm{pack}}(\tau)\) denote pack electrical power (W). The per-sample energy increment is
\begin{equation}
\label{eq:energy_increment_sf}
e_{k,q}^{r}
\triangleq
\int_{\tau_{r-1}}^{\tau_r} P_{\mathrm{pack}}(\xi)\, d\xi
\;\approx\;
P_{\mathrm{pack}}(\tau_r)\,\Delta\tau,
\qquad \tau_r=r\Delta\tau,
\end{equation}
where the discretization is consistent with standard numerical quadrature at telemetry rate.
This notation avoids overloading \(E\), which denotes local epochs in Section~\ref{subsec:method_setup}.

\paragraph{Windowing and supervised examples.}
To capture temporal dependence in BEV energy dynamics, we form fixed-length windows of
\(L_{\mathrm{win}}\) samples (so that \(L_{\mathrm{win}}\Delta\tau\) corresponds to a desired horizon, e.g., 60 seconds).
We use \(L_{\mathrm{win}}\) (not \(L\)) to avoid collision with the smoothness constant \(L\) in Section~\ref{subsec:method_setup}.
For each \(r\ge L_{\mathrm{win}}\), define the windowed input matrix
\begin{equation}
\label{eq:window_input_sf}
\mathbf{X}_{k,q}^{r}
\triangleq
\big[\mathbf{x}_{k,q}^{r-L_{\mathrm{win}}+1},\dots,\mathbf{x}_{k,q}^{r}\big]
\in\mathbb{R}^{L_{\mathrm{win}}\times d_x},
\end{equation}
and the windowed energy label
\begin{equation}
\label{eq:window_label_sf}
Y_{k,q}^{r}
\triangleq
\sum_{j=r-L_{\mathrm{win}}+1}^{r} e_{k,q}^{j}
\in\mathbb{R}.
\end{equation}

\paragraph{Mapping to the FL dataset \(\mathcal{P}_k\).}
Each supervised example is a pair \((x_i,y_i)\) as in Eq.~\eqref{eq:local_obj}, instantiated as
\[
x_i \equiv \mathbf{X}_{k,q}^{r},\qquad
y_i \equiv Y_{k,q}^{r},
\]
where the sample identity \(i\) corresponds to a specific \((q,r)\) within client \(k\).
Thus, the client-local dataset is
\begin{equation}
\label{eq:Pk_systemframework}
\mathcal{P}_k
\triangleq
\Big\{ \big(\mathbf{X}_{k,q}^{r},\, Y_{k,q}^{r}\big)\;:\;
q=1,\dots,Q_k,\;\; r=L_{\mathrm{win}},\dots,N_{k,q} \Big\},
\end{equation}
with \(n_k=|\mathcal{P}_k|\) matching Section~\ref{subsec:method_setup}.
This section defines only the system/data interface; the optimization and aggregation are fixed by
Section~\ref{sec:methodology}.

\subsubsection{Vehicle Specifications and Energy Modeling}
\label{subsec:VehicleSpecifications}

We consider a connected BEV fleet with natural heterogeneity across vehicle configurations,
auxiliary loads, battery state-of-health, and driving environments. When direct pack power
telemetry is available, we compute labels using Eq.~\eqref{eq:energy_increment_sf} without
requiring a battery model. When only partial signals are available (or to ensure physical
consistency under controlled heterogeneity), we generate reference signals using a standard
component-based BEV simulation pipeline (e.g., AVL CRUISE\texttrademark~M) \cite{avl_cruise_m}.
The physical modeling below is included \emph{only} to document label provenance and does not
constitute a contribution in electrochemical modeling.

\paragraph{Pack power definition.}
Pack electrical power is defined as
\begin{equation}
\label{eq:pack_power}
P_{\mathrm{pack}}(\tau) \triangleq V_{\mathrm{pack}}(\tau)\, I_{\mathrm{pack}}(\tau),
\end{equation}
where \(V_{\mathrm{pack}}(\tau)\) and \(I_{\mathrm{pack}}(\tau)\) denote pack terminal voltage (V)
and current (A). The energy labels \(e_{k,q}^{r}\) and \(Y_{k,q}^{r}\) follow from
Eqs.~\eqref{eq:energy_increment_sf}--\eqref{eq:window_label_sf}.

\paragraph{Equivalent-circuit battery model (optional, for reference generation).}
If \(V_{\mathrm{pack}}, I_{\mathrm{pack}}\), or \(P_{\mathrm{pack}}\) are not directly observed, we compute them using
a first-order equivalent-circuit model (one RC pair). Let \(\mathrm{SoC}(\tau)\in[0,1]\) be state-of-charge,
\(Q_{\mathrm{cell}}>0\) be cell capacity (Ah), and \(I_{\mathrm{cell}}(\tau)\) be cell current (A).
SoC is computed by Coulomb counting:
\begin{equation}
\label{eq:soc_system}
\mathrm{SoC}(\tau)
=
\mathrm{SoC}(0)
-
\frac{1}{Q_{\mathrm{cell}}}\int_{0}^{\tau} I_{\mathrm{cell}}(\xi)\, d\xi .
\end{equation}
Let \(V_{\mathrm{cell}}(\tau)\) be terminal cell voltage, \(V_{\mathrm{oc}}(\mathrm{SoC})\) the open-circuit voltage curve,
\(R_0(\mathrm{SoC})\) the series resistance, and \((R_1(\mathrm{SoC}),\,\mathsf{C}_1(\mathrm{SoC}))\) the transient RC parameters.
We write \(\mathsf{C}_1\) (not \(C_1\)) to avoid collision with the participation fraction \(C\) in Section~\ref{subsec:method_setup}.
The ECM dynamics are:
\begin{equation}
\label{eq:ecm_system}
\begin{split}
V_{\mathrm{cell}}(\tau)
&=
V_{\mathrm{oc}}\!\left(\mathrm{SoC}(\tau)\right)
-
V_1(\tau)
-
R_0\!\left(\mathrm{SoC}(\tau)\right)\,\\
\times I_{\mathrm{cell}}(\tau),
\\[0.6em]
\mathsf{C}_1\!\left(\mathrm{SoC}(\tau)\right)\,
\frac{d V_1(\tau)}{d\tau}
&=
I_{\mathrm{cell}}(\tau)
-
\frac{V_1(\tau)}
{R_1\!\left(\mathrm{SoC}(\tau)\right)} .
\end{split}
\end{equation}

For a pack with \(n_s\) cells in series and \(n_p\) in parallel, we use the standard mapping
\[
V_{\mathrm{pack}}(\tau)=n_s V_{\mathrm{cell}}(\tau),\qquad
I_{\mathrm{pack}}(\tau)=n_p I_{\mathrm{cell}}(\tau),
\]
and compute \(P_{\mathrm{pack}}(\tau)\) via Eq.~\eqref{eq:pack_power}.
Cell imbalance and degradation dynamics are neglected for tractability.

\begin{table}[!t]
\caption{Symbol provenance and units for BEV energy labeling.}
\label{tab:provenance_symbols}
\centering
\footnotesize
\setlength{\tabcolsep}{4pt}
\renewcommand{\arraystretch}{1.08}
\begin{tabular}{@{}p{0.26\linewidth}p{0.70\linewidth}@{}}
\toprule
\textbf{Symbol} & \textbf{Meaning / units / provenance} \\
\midrule
$\tau,\Delta\tau$ &
Trip time (s) and sampling period (s); distinct from FL rounds $t$. \\
$P_{\mathrm{pack}}(\tau)$ &
Pack electrical power (W), telemetry or simulation \cite{avl_cruise_m}. \\
$V_{\mathrm{pack}}(\tau), I_{\mathrm{pack}}(\tau)$ &
Pack voltage (V) and current (A), telemetry or ECM. \\
$e_{k,q}^{r}$ &
Per-sample energy increment (J or Wh), Eq.~\eqref{eq:energy_increment_sf}. \\
$Y_{k,q}^{r}$ &
Window energy label (J or Wh), Eq.~\eqref{eq:window_label_sf}. \\
$\mathrm{SoC}(\tau)$ &
State-of-charge in $[0,1]$, Eq.~\eqref{eq:soc_system}. \\
$V_{\mathrm{oc}}(\cdot)$ &
Open-circuit voltage curve (V), lookup versus SoC. \\
$R_0(\cdot),R_1(\cdot)$ &
Resistances ($\Omega$), SoC-dependent parameters. \\
$\mathsf{C}_1(\cdot)$ &
Capacitance (F), SoC-dependent parameter; distinct from participation fraction $C$. \\
\bottomrule
\end{tabular}
\end{table}

\subsubsection{Data Analysis and Processing}
\label{subsec:DataAnalysis}

This subsection specifies dataset provenance and the preprocessing steps that construct each client dataset
\(\mathcal{P}_k\) used in Section~\ref{sec:methodology}.
Our real-world base is the Vehicle Energy Dataset (VED) \cite{oh2019ved}, which provides privacy-protected
time-series signals suitable for energy modeling. When richer context (e.g., map-based elevation and speed limits)
is required, we optionally use the extended VED (eVED) enrichment pipeline \cite{zhang2022eved}.
For controlled scenario generation (without reporting results here), we support simulator-based augmentation using
traffic/driving simulators such as SUMO and CARLA \cite{krajzewicz2012sumo,dosovitskiy2017carla}.

\paragraph{Preprocessing (vehicle-local).}
All preprocessing is performed locally on each client to preserve privacy boundaries.
We resample raw channels to a uniform \(\Delta\tau\), synchronize timestamps, handle missing values
(interpolation or deletion depending on gap length), and filter obvious sensor glitches.
Features are normalized using per-client statistics computed on the local training split (e.g., robust z-score),
to prevent leakage of fleet-wide distributional information.

\paragraph{Feature construction and dimensionality.}
The base feature vector \(\mathbf{x}_{k,q}^{r}\in\mathbb{R}^{d_x}\) is built from raw telemetry
\(\mathbf{z}_{k,q}^{r}\) and may include derived nonlinear transforms (e.g., polynomial terms of speed)
to reflect known nonlinearities in traction power demand. Dimensionality reduction (e.g., PCA) or model-based
feature selection may be applied locally, without altering the supervised dataset definition.

\begin{table}[!t]
\caption{Dataset source and structure (illustrative; full training uses the complete corpus).}
\label{tab:dataset_source}
\centering
\footnotesize
\setlength{\tabcolsep}{4pt}
\renewcommand{\arraystretch}{1.08}
\begin{tabular}{@{}p{0.28\linewidth}p{0.68\linewidth}@{}}
\toprule
\textbf{Property} & \textbf{Description} \\
\midrule
Primary source &
VED (Ann Arbor / Michigan, USA) \cite{oh2019ved}. \\
Optional enrichment &
eVED (map/context augmentation) \cite{zhang2022eved}. \\
Signals &
Privacy-protected time-series telemetry including energy/fuel-related measurements. \\
Windowing &
$L_{\mathrm{win}}$ samples per window; label is window-sum energy (Eq.~\eqref{eq:window_label_sf}). \\
\bottomrule
\end{tabular}
\end{table}

\begin{table}[!t]
\caption{Local preprocessing and feature construction (client-side).}
\label{tab:preprocessing_features}
\centering
\footnotesize
\setlength{\tabcolsep}{4pt}
\renewcommand{\arraystretch}{1.08}
\begin{tabular}{@{}p{0.30\linewidth}p{0.66\linewidth}@{}}
\toprule
\textbf{Process} & \textbf{Details} \\
\midrule
Resampling/sync &
Uniform $\Delta\tau$, channel alignment, and missing-data handling. \\
Normalization &
Per-client robust normalization on the local training split. \\
Windowing &
$\mathbf{X}_{k,q}^{r}\in\mathbb{R}^{L_{\mathrm{win}}\times d_x}$ (Eq.~\eqref{eq:window_input_sf}). \\
Targets &
$Y_{k,q}^{r}\in\mathbb{R}$ as window energy (Eq.~\eqref{eq:window_label_sf}). \\
Derived features &
Nonlinear transforms/interactions computed locally as needed. \\
\bottomrule
\end{tabular}
\end{table}



\section{Experiments}
\label{experiments}

\subsection{Experimental Goals and Claim-to-Evidence Map}
\label{subsec:claim_evidence_map}

We organize the experimental section around explicit claims (C1--C7) that directly correspond to the components defined in
Section~\ref{sec:methodology}. For each claim, we pre-specify (i) the exact evidence (metrics, figures, and tables) used for evaluation,
(ii) the required baselines, and (iii) where the evidence appears in subsequent subsections. No results are presented here; this subsection
serves as a roadmap for falsifiable and reproducible evaluation.


\begin{table*}[t]
\centering
\caption{\textbf{Claim-to-evidence map for FO-RI-FedAvg.} Each claim is supported by planned figures/tables with explicit metrics and empirical evidence.}
\label{tab:claim_evidence_map}
\footnotesize
\setlength{\tabcolsep}{6pt}
\renewcommand{\arraystretch}{1.08}
\begin{tabular}{@{}p{0.07\linewidth} p{0.33\linewidth} p{0.56\linewidth}@{}}
\toprule
\textbf{Claim} &
\textbf{What the claim asserts} &
\textbf{Evidence (metrics + figure/table references)} \\
\midrule
\textbf{C1} &
\textbf{Drift mitigation under partial participation.}
Roughness-controlled proximal training (Eqs.~\eqref{eq:prox_obj}--\eqref{eq:prox_grad} with \(r(\mathcal{I}_k)\)) reduces client drift and stabilizes aggregation when fewer clients participate per round. &
\textbf{Drift metrics:}
per-round client update magnitude \(\|\mathbf{w}_{t+1}^k-\mathbf{w}_t\|_2\),
dispersion across \(k\in S_t\),
update variance across clients,
rounds-to-stability.
\textbf{Figures:}
Figs.~\ref{fig:drift_ved_mean}, \ref{fig:drift_eved_mean}, \ref{fig:drift_cv_ved},
\ref{fig:participation_drift_ved}. \\
\midrule
\textbf{C2} &
\textbf{Fractional-order stability/convergence for \(0<\alpha\le1\).}
Fractional preconditioning (Eqs.~\eqref{eq:fo_update_generic}--\eqref{eq:frac_precond}) yields more stable and/or faster convergence than integer-order updates and recovers SGD at \(\alpha=1\). &
\textbf{Convergence metrics:}
test/validation error vs.\ communication rounds,
rounds-to-threshold,
final utility.
\textbf{Figures and tables:}
Tabs.~\ref{tab:bev_main_results}, \ref{tab:rounds_to_threshold};
Figs.~\ref{fig:convergence_ved}, \ref{fig:convergence_eved},
\ref{fig:bestsofar_ved}, \ref{fig:bestsofar_eved}. \\
\midrule
\textbf{C3} &
\textbf{Low amortized overhead with detailed breakdown.}
FO-RI-FedAvg realizes fractional-order behavior via a Taylor-series-based Caputo-inspired surrogate,
implemented as lightweight element-wise preconditioning with $O(d)$ extra work and $O(d)$ memory per local step. &
\textbf{Overhead analysis:}
end-to-end round time relative to baselines (Table~\ref{tab:overhead_time_round}),
and component-wise breakdown separating local training from diagnostics
(Table~\ref{tab:overhead_breakdown}).
Fractional preconditioning is consistently lightweight, while roughness probing is the dominant non-training cost.
\textbf{Tables:}
Tabs.~\ref{tab:overhead_time_round}, \ref{tab:overhead_breakdown}. \\
\midrule
\textbf{C4} &
\textbf{Reductions and ablations.}
FO-RI-FedAvg reduces to RI-FedAvg when \(\alpha=1\), to FO-FedAvg when \(r(\mathcal{I}_k)=0\), and to FedAvg when both conditions hold. &
\textbf{Ablation grid:}
controlled toggles for \(\alpha\) and \(r(\mathcal{I}_k)\),
verification of performance ordering and mechanism attribution.
\textbf{Figures and tables:}
Figs.~\ref{fig:ablate_lambda_rmse}, \ref{fig:ablate_safeguards};
Tabs.~\ref{tab:ablation_summary}, \ref{tab:ablate_probe_freq}.\\
\midrule
\textbf{C5} &
\textbf{Spectral gating improves robustness in low-participation or noisy-client regimes.}
Spectral damping via \(\kappa_k\) stabilizes training under anisotropy/overconfidence and can be disabled with \(\beta_\kappa=0\). &
\textbf{Stress tests:}
low participation settings and noisy-client injections;
comparison of gated vs.\ ungated variants.
\textbf{Figure and table:}
Fig.~\ref{fig:ablate_alpha_rmse} +
Tab.~\ref{tab:ablate_spectral_points}. \\
\midrule
\textbf{C6} &
\textbf{BEV energy modeling as a canonical long-memory FL domain.}
On BEV telemetry, FO-RI-FedAvg achieves consistent utility gains and improved stability aligned with system characteristics. &
\textbf{Utility and stratified analysis:}
RMSE/MAE/MAPE on window energy \(Y_{k,q}^r\).
\textbf{Figures and tables:}
Figs.~\ref{fig:vehicletype_rmse_ved}, \ref{fig:vehicletype_rmse_eved};
Tabs.~\ref{tab:vehicletype_drift_ved}, \ref{tab:vehicletype_drift_eved}. \\
\midrule
\textbf{C7} &
\textbf{Scalability to different client populations (50--200 clients).}
FO-RI-FedAvg maintains low error and exhibits graceful degradation as the number of clients grows significantly. &
\textbf{Scaling metrics:}
RMSE vs.\ number of clients, \textbf{Table:}
Tab.~\ref{tab:bev_scaling_clients}.\\
\bottomrule
\end{tabular}
\end{table*}

\paragraph{Evidence rules (evaluation protocol).}
All reported metrics are computed at the \emph{communication-round} level to reflect FL efficiency under client participation.
We report mean and uncertainty (e.g., mean\(\pm\)std or 95\% confidence intervals) over multiple random seeds and client-sampling realizations.
Baselines are tuned under a comparable hyperparameter budget, with all tuning ranges and selections disclosed.
Client construction follows vehicle-defined partitions, and participation is controlled via the fraction \(C\); all methods share identical
model architectures, preprocessing, and trip-disjoint splits unless a baseline definition requires otherwise.
For overhead, we report both (i) absolute wall-clock time per round and (ii) a normalized breakdown separating local training
from diagnostic computation, with probe-frequency sensitivity when roughness diagnostics are amortized.

Unless stated otherwise, each table/figure aggregates results over multiple independent runs (different random seeds and client-sampling realizations).
Due to stochasticity and separate executions, minor numerical differences across tables may occur even under the same nominal settings.



\subsection{Experimental Setup}
\label{subsec:exp_setup}

This subsection specifies the datasets, client construction, model architectures, and training protocol used to evaluate
FO-RI-FedAvg in the BEV energy modeling setting of Section~\ref{sec:system}, under the federated optimization formulation of
Section~\ref{sec:methodology}. No results are reported here.
We explicitly evaluate robustness to partial participation by sweeping the participation fraction \(C\) in
Section~\ref{subsec:sensitivity_participation}.

\subsubsection{Datasets and Tasks}
\label{subsubsec:exp_datasets_tasks}

\paragraph{Primary BEV datasets (VED/eVED).}
Our primary real-world corpus is the Vehicle Energy Dataset (VED) \cite{oh2019ved}, which provides large-scale vehicle
telemetry and energy-related signals collected in real driving conditions with privacy-preserving de-identification.
When richer context features are needed (e.g., calibrated GPS and map-derived attributes such as elevation and speed limits),
we use the extended VED (eVED) enrichment pipeline \cite{zhang2022eved}. The BEV instantiation and labeling pipeline follow
Section~\ref{sec:system}.

\noindent\textbf{Scope note.}
Our experimental evaluation focuses on VED and eVED due to the limited availability of public real-world BEV telemetry corpora;
broader validation on additional federated telemetry sources is left for future work.

\paragraph{Prediction task.}
Given a windowed input \(\mathbf{X}_{k,q}^{r}\in\mathbb{R}^{L_{\mathrm{win}}\times d_x}\), the goal is to predict the
window energy label \(Y_{k,q}^{r}\in\mathbb{R}\) (Eq.~\eqref{eq:window_label_sf}), where
\(Y_{k,q}^{r}=\sum_{j=r-L_{\mathrm{win}}+1}^{r} e_{k,q}^{j}\) and \(e_{k,q}^{j}\) is derived from pack power
(Eq.~\eqref{eq:energy_increment_sf}). We evaluate forecasting performance using standard regression metrics (reported later),
computed at the communication-round level.

\paragraph{Splits and leakage prevention.}
To avoid temporal and identity leakage in time-series telemetry, we perform splits at the \emph{trip} level.
Within each client \(k\), trips are partitioned into train/validation/test sets with disjoint trip IDs, ensuring that no windows
from the same trip appear in multiple splits. Clients correspond to individual vehicles, reflecting realistic federated deployments.
For cross-vehicle generalization checks, we additionally consider a \emph{vehicle-held-out} split in which a subset of clients
is reserved for testing (no participation in training), while maintaining identical feature construction and normalization rules.

\begin{table}[H]
\centering
\caption{\textbf{Dataset and task specification.} Windowed inputs and energy labels follow Section~\ref{sec:system}.}
\label{tab:dataset_setup}
\setlength{\tabcolsep}{3pt}
\renewcommand{\arraystretch}{1.08}
\begin{tabularx}{\columnwidth}{p{0.26\columnwidth} X}
\toprule
\textbf{Item} & \textbf{Specification} \\
\midrule
Datasets & VED \cite{oh2019ved} and eVED \cite{zhang2022eved}. \\
Task & Predict window energy \(Y_{k,q}^{r}\) from window features \(\mathbf{X}_{k,q}^{r}\) (Eqs.~\eqref{eq:window_input_sf}--\eqref{eq:window_label_sf}). \\
Windowing & Fixed length \(L_{\mathrm{win}}\) samples; sampling period \(\Delta\tau\) (Section~\ref{sec:system}). \\
Splits & Trip-level splits within each client; optional vehicle-held-out evaluation for generalization. \\
Privacy boundary & Raw telemetry remains local; only model parameters/updates are transmitted (Section~\ref{sec:system}). \\
\bottomrule
\end{tabularx}
\end{table}

\subsubsection{Client Construction and Participation}
\label{subsubsec:exp_partitioning}

\paragraph{Client construction (natural non-IID).}
Clients correspond to individual vehicles, each contributing trip windows generated under its own operating conditions.
This naturally induces non-IID client data distributions due to differences in routes, climates, vehicle variants,
and driving behaviors. Unless otherwise stated, we use these natural client partitions throughout all experiments
without introducing additional synthetic non-IID protocols.

\paragraph{Partial participation.}
At each communication round \(t\), the server samples a participating client set \(S_t\) with
\(|S_t|=\max(\lceil CK\rceil,1)\) as in Section~\ref{subsec:method_setup}. For participation sensitivity experiments,
we sweep \(C\in\{0.1,0.2,0.3,0.5\}\) while keeping the model architecture and the local training budget
(local steps/epochs and batch size) unchanged across methods.

\subsubsection{Models}
\label{subsubsec:exp_models}

\paragraph{Primary predictor (default backbone).}
Our main BEV energy predictor maps the windowed input
\(\mathbf{X}_{k,q}^{r}\in\mathbb{R}^{L_{\mathrm{win}}\times d_x}\) to a scalar prediction
\(\widehat{Y}_{k,q}^{r}\in\mathbb{R}\).
Unless explicitly stated otherwise, \textbf{all previously reported results in this paper}
(main BEV results, convergence and efficiency, client churn, and overhead analyses)
use an \textbf{edge-friendly ANN regressor} as the default backbone:
a \textbf{two-hidden-layer MLP} with ReLU activations trained on the windowed feature tensor.
For consistency with the architecture-generalization study
(Section~\ref{subsec:arch_generalization}), we use \(n=60\) time steps in the input window.

\paragraph{Architecture robustness checks.}
To verify that observed improvements are not backbone-specific, we evaluate FO-RI-FedAvg
and all baselines under identical federation settings while swapping the neural predictor.
Specifically, we consider three backbones:
(i) \textbf{ANN} (default), operating on the windowed input representation;
(ii) \textbf{GRU}, a single-layer GRU encoder ingesting the sequence
\((X_v^{r-n+1},\dots,X_v^{r})\) and producing a regression output via a linear head;
(iii) \textbf{LSTM}, a single-layer LSTM with the same hidden size and the same regression head.
Recurrent models capture temporal dependence directly, whereas the MLP relies on the
windowed feature tensor to encode history.

\paragraph{Loss and output.}
Across all methods, baselines, and architectures, we use the same regression objective
for \(\ell(\mathbf{w};x_i,y_i)\) (Eq.~\eqref{eq:local_obj}), chosen consistently across experiments
(e.g., MSE or Huber), and report RMSE/MAE as primary evaluation metrics.

\begin{table}[!t]
\caption{Model configurations. Within each experiment, all methods share the same backbone; architecture-robustness studies compare backbones under identical federation settings.}
\label{tab:model_setup}
\centering
\footnotesize
\setlength{\tabcolsep}{3pt}
\renewcommand{\arraystretch}{1.08}
\begin{tabular}{@{}p{0.34\linewidth}p{0.62\linewidth}@{}}
\toprule
\textbf{Backbone} & \textbf{Input handling} \\
\midrule
ANN (2 hidden layers, ReLU) &
Windowed input tensor (with $n=60$ time steps) mapped to $\widehat{Y}_{k,q}^{r}$ via a lightweight MLP regressor. \\
GRU (1 layer) + linear head &
Sequential encoder over $(X_v^{r-n+1},\dots,X_v^{r})$; regression via linear head. \\
LSTM (1 layer) + linear head &
Sequential encoder over $(X_v^{r-n+1},\dots,X_v^{r})$; regression via linear head. \\
\bottomrule
\end{tabular}
\end{table}

\subsubsection{Training Protocol}
\label{subsubsec:exp_training_protocol}

\paragraph{Federated optimization loop.}
All methods follow the same FL loop and aggregation in Eq.~\eqref{eq:fedavg_agg} with \(T\) rounds, \(K\) clients,
participation fraction \(C\), local epochs \(E\) (equivalently \(H\) local steps), batch size \(B\), and learning-rate schedule
\(\{\eta_t\}_{t=0}^{T-1}\) (Section~\ref{subsec:method_setup}). Unless otherwise stated, we evaluate checkpoints at a fixed cadence
(e.g., every round or every few rounds) using the validation split defined above, and we use a consistent early-stopping rule
across methods (e.g., best validation score with a patience window).

\paragraph{FO-RI-FedAvg hyperparameters and diagnostics cadence.}
FO-RI-FedAvg uses fractional order \(\alpha\in(0,1]\), stabilizer \(\delta>0\), and (optionally) preconditioner clipping
(Eq.~\eqref{eq:precond_clip}). Roughness diagnostics use \(M\) random directions, probe radius \(\ell\), grid size \(m\),
probe batch size \(B_{\mathrm{probe}}\), and stabilizers \(\epsilon_A,\epsilon_T\) (Section~\ref{subsec:method_diagnostics}).
Diagnostics can be computed every round or amortized (e.g., every \(R\) rounds) with the same schedule across relevant methods.
If spectral gating is enabled, we set \(\beta_\kappa\ge 0\) and compute \(\kappa_k\) as in Eq.~\eqref{eq:spectral_flatness}.

\paragraph{Baseline fairness and tuning budget.}
All baselines (FedAvg, FedProx, SCAFFOLD, RI-FedAvg, FO-FedAvg) use the same data splits, same architecture, and the same
client sampling schedule. Hyperparameters are tuned under a comparable budget (same number of trials and comparable ranges),
and the selected configurations are reported alongside results.

\begin{table}[!t]
\caption{Training protocol and hyperparameter ranges. Concrete values are fixed per experiment and disclosed with results; ranges here define the tuning budget.}
\label{tab:training_hparams}
\centering
\footnotesize
\setlength{\tabcolsep}{3pt}
\renewcommand{\arraystretch}{1.08}
\begin{tabular}{@{}p{0.33\linewidth}p{0.63\linewidth}@{}}
\toprule
\textbf{Parameter} & \textbf{Specification / range} \\
\midrule
Rounds / clients &
$T$ rounds; $K$ clients (vehicles). \\
Participation &
$C\in(0,1]$; $|S_t|=\max(\lceil CK\rceil,1)$. \\
Local compute &
Local epochs $E\in\{1,\dots,5\}$ or steps $H$; batch size $B\in[16,256]$. \\
Learning rate &
$\eta_t\in\{\eta,\ \eta_0/\sqrt{t+1}\}$, with $\eta_0\in[10^{-4},10^{-1}]$. \\
Prox.\ regularization &
$\lambda_t\in\{\lambda,\ \text{schedule}\}$, $\lambda\in[10^{-4},10^{0}]$; response $r(\mathcal{I}_k)$ as in Eq.~\eqref{eq:ri_response}. \\
Fractional order &
$\alpha\in\{1.0,0.9,0.7,0.5,0.3\}$ (restricted to $0<\alpha\le 1$). \\
Stabilizers &
$\delta\in[10^{-8},10^{-4}]$; $\epsilon_A,\epsilon_T,\epsilon_F$ small positive constants. \\
Roughness probe &
$M\in[5,20]$, $\ell\in[10^{-4},10^{-2}]$, $m\in[20,200]$, $B_{\mathrm{probe}}\in[16,256]$. \\
Spectral gate (opt.) &
$\beta_\kappa\in[0,10]$; $\kappa_k$ via Eq.~\eqref{eq:spectral_flatness}. \\
\bottomrule
\end{tabular}
\end{table}

\paragraph{External baselines (prior work).}

\paragraph{(B1) FedAvg.}
Federated Averaging (FedAvg) is the standard FL baseline that performs multiple steps of local SGD on each participating client
and aggregates the resulting client models (or updates) by a data-size-weighted average as in Eq.~\eqref{eq:fedavg_agg} \cite{mcmahan2017fedavg}.

\paragraph{(B2) FedProx.}
FedProx mitigates client drift under statistical heterogeneity by adding a \emph{proximal} regularization term to each client’s local objective,
penalizing deviation from the broadcast model \(\mathbf{w}_t\) (typically \(\frac{\mu}{2}\|\mathbf{w}-\mathbf{w}_t\|_2^2\)).
This stabilizes local training when client objectives differ and can also improve robustness to varying local work \cite{li2018fedprox2}.

\paragraph{(B3) SCAFFOLD.}
SCAFFOLD reduces client drift by using \emph{control variates} (server and client correction terms) that debias local SGD updates,
so that the expected local update better matches the global objective’s update direction under heterogeneous client objectives \cite{karimireddy2020scaffold}.

\paragraph{(B4) FedNova \cite{wang2020fednova}.}
FedNova addresses \emph{objective inconsistency} arising from heterogeneous local computation (e.g., different numbers of local steps/epochs)
by \emph{normalizing} client updates with respect to their effective local progress before aggregation. This yields an aggregation rule that is more consistent
across clients with unequal local work \cite{wang2020fednova}.

\paragraph{(B5) FedAdam \cite{reddi2021fedopt}.}
FedAdam (FedOpt family) is a server-side adaptive optimizer baseline: clients perform local SGD, the server forms an aggregated update direction
(from the weighted average of client model deltas), and then applies an Adam-style moment update to the global model.
We tune the server learning rate and Adam hyperparameters \((\beta_1,\beta_2,\epsilon)\) along with the client local learning rate \cite{reddi2021fedopt}.

\paragraph{(B6) MOON \cite{li2021moon}.}
MOON (model-contrastive federated learning) augments local training with a \emph{model-level contrastive} objective on learned representations:
it encourages the current client model to stay close to the global model (positive) while contrasting against prior client models (negatives),
thereby reducing representation drift under non-IID data. We tune the contrastive weight (e.g., \(\mu\)) and temperature (e.g., \(\tau\)) in addition to the local learning rate \cite{li2021moon}.

\paragraph{(B7) FedDyn \cite{acar2021feddyn}.}
FedDyn targets heterogeneity-induced bias by adding a \emph{dynamic regularization} term to the local objectives (involving a quadratic penalty around \(\mathbf{w}_t\)
and a linear term that is updated across rounds). This mechanism aims to align stationary points of the modified local objectives with those of the global objective,
improving stability under non-IID client objectives. We tune the method-specific regularization strength (e.g., \(\alpha\)) under the same tuning budget \cite{acar2021feddyn}.

\paragraph{(B8) FedEL \cite{zhang2025fedel}.}
FedEL (\emph{Federated Elastic Learning}) targets \emph{system heterogeneity/stragglers} by using a window-based training process that progressively trains different parts of the model:
it slides a window over the network and dynamically selects important tensors to train under a coordinated runtime budget.
FedEL additionally includes a tensor-importance adjustment module to harmonize local and global tensor importance and reduce bias under data heterogeneity \cite{zhang2025fedel}.

\paragraph{Internal variants (ours; \emph{not} baselines).}
The following methods are \textbf{component-isolating variants of FO-RI-FedAvg}, included to attribute gains to (i) roughness-informed control and
(ii) fractional-order dynamics, rather than to compare against prior work.

\paragraph{(V1) RI-FedAvg (\(\alpha{=}1\); integer-order roughness-informed).}
RI-FedAvg uses the same client roughness index \(\mathcal{I}_k\) (Section~\ref{subsec:method_diagnostics}) to modulate a proximal pull toward \(\mathbf{w}_t\),
but performs \emph{standard} (non-fractional) local SGD updates (equivalently, FO-RI-FedAvg with \(\alpha=1\)).
We tune the base regularization schedule \(\{\lambda_t\}\), the response map \(r(\mathcal{I}_k)\), and \(\{\eta_t\}\).
This variant isolates the contribution of roughness-informed control without fractional-order dynamics.

\paragraph{(V2) FO-FedAvg (\(r(\mathcal{I}_k){=}0\); fractional-order without roughness control).}
FO-FedAvg activates fractional-order local update dynamics but removes roughness-informed control by setting \(r(\mathcal{I}_k)=0\)
(and \(\lambda_t=0\)) in Eq.~\eqref{eq:prox_obj}, while retaining the fractional preconditioning in
Eqs.~\eqref{eq:fo_update_generic}--\eqref{eq:frac_precond} \cite{wang2017fractionalbp,shin2023fractional}.
We tune \(\alpha\in(0,1]\), \(\delta\), optional preconditioner clipping bounds \((p_{\min},p_{\max})\), and \(\{\eta_t\}\).
This variant isolates the contribution of fractional memory effects without roughness-aware drift control.

\paragraph{(Ours) FO-RI-FedAvg (full method).}
Our full method combines (i) roughness-controlled proximal stabilization via \(\lambda_t\,r(\mathcal{I}_k)\) in Eq.~\eqref{eq:prox_obj} and
(ii) fractional-order preconditioned local updates via Eqs.~\eqref{eq:fo_ri_update}--\eqref{eq:frac_precond},
optionally with spectral gating (Eq.~\eqref{eq:spectral_gate}).
This setting corresponds to the complete algorithm described in Algorithm~\ref{alg:fo_ri_fedavg} and Figure~\ref{fig:fo_ri_fedavg}.

\begin{table*}[!t]
\centering
\caption{\textbf{Algorithm summary (external baselines + our variants).} All methods share the same model, data splits, and client sampling schedule; differences are in the update/aggregation/control mechanisms.}
\label{tab:baseline_summary}
\setlength{\tabcolsep}{6pt}
\renewcommand{\arraystretch}{1.12}
\begin{tabular}{@{}p{0.24\linewidth} p{0.72\linewidth}@{}}
\toprule
\textbf{Method} & \textbf{Key mechanism} \\
\midrule

\multicolumn{2}{@{}l}{\textbf{External baselines (prior work)}} \\
\midrule
FedAvg \cite{mcmahan2017fedavg} &
Multiple steps of local SGD on clients + data-size-weighted model averaging (Eq.~\eqref{eq:fedavg_agg}). \\
\midrule
FedProx \cite{li2018fedprox2} &
Adds a proximal regularizer (typically \(\frac{\mu}{2}\|\mathbf{w}-\mathbf{w}_t\|_2^2\)) to anchor local training to \(\mathbf{w}_t\), reducing client drift under non-IID heterogeneity. \\
\midrule
SCAFFOLD \cite{karimireddy2020scaffold} &
Uses server/client control variates to debias local SGD updates (variance-reduction style correction), mitigating drift from heterogeneous objectives. \\
\midrule
FedNova \cite{wang2020fednova} &
Normalizes client updates by effective local progress (e.g., number of local steps) before aggregation to address objective inconsistency under heterogeneous local work. \\
\midrule
FedAdam \cite{reddi2021fedopt} &
FedOpt server-side adaptive optimization: apply Adam-style moment updates to the aggregated client update direction (weighted average of client deltas). \\
\midrule
MOON \cite{li2021moon} &
Model-contrastive regularization on representations: pull local model toward the global model (positive) and contrast against previous local models (negatives) to reduce representation drift under non-IID data. \\
\midrule
FedDyn \cite{acar2021feddyn} &
Dynamic regularization of local objectives (quadratic anchoring around \(\mathbf{w}_t\) plus a cross-round updated linear term) to reduce heterogeneity-induced bias and stabilize training. \\
\midrule
FedEL \cite{zhang2025fedel} &
System-heterogeneity/straggler method: window-based progressive training that selects important tensors under a coordinated runtime budget, plus tensor-importance adjustment to harmonize local/global importance. \\

\midrule

\multicolumn{2}{@{}l}{\textbf{Internal variants (ours)}} \\
\midrule
RI-FedAvg (\(\alpha=1\)) &
Roughness index \(\mathcal{I}_k\) modulates proximal strength (Eq.~\eqref{eq:prox_obj}); integer-order local SGD (no fractional preconditioning). \\
\midrule
FO-FedAvg (\(r(\mathcal{I}_k)=0\)) &
Fractional-order preconditioning (\(0<\alpha\le 1\)) via Eqs.~\eqref{eq:fo_update_generic}--\eqref{eq:frac_precond} without roughness control (\(\lambda_t=0\)). \\
\midrule
\textbf{FO-RI-FedAvg} &
Full method: roughness-controlled proximal stabilization (Eq.~\eqref{eq:prox_obj}) + fractional preconditioned local updates (Eq.~\eqref{eq:fo_ri_update}), optionally with spectral gating (Eq.~\eqref{eq:spectral_gate}). \\
\bottomrule
\end{tabular}
\end{table*}


\subsection{Evaluation Metrics}
\label{subsec:eval_metrics}

This subsection defines the evaluation metrics used throughout the Experiments section.
All metrics are computed at the \emph{communication-round} level (round index \(t\)) to reflect
federated efficiency under partial participation, and are reported with uncertainty over multiple
random seeds and client-sampling realizations (as specified in \S\ref{subsec:claim_evidence_map}).

\subsubsection{Evaluation sets and aggregation}
\label{subsubsec:eval_sets}

For each client \(k\), let \(\mathcal{T}_k\) denote its local test set (trip-disjoint from training),
constructed from \(\mathcal{P}_k\) as described in \S\ref{subsubsec:exp_datasets_tasks}. Each example is a
window-label pair \((\mathbf{X}_{k,q}^{r}, Y_{k,q}^{r})\), where \(\mathbf{X}_{k,q}^{r}\) is the input feature window
(e.g., a fixed-length time window of driving signals) and \(Y_{k,q}^{r}\) is the corresponding supervised energy label
over that window. The indices have the following interpretation: \(k\) is the client/vehicle index, \(q\) is the trip
(or trajectory segment) index within client \(k\), and \(r\) is the window endpoint (time step) within trip \(q\).
The model prediction at round \(t\) is \(\widehat{Y}_{k,q}^{r}(t)\), obtained by evaluating the current global model
\(\mathbf{w}_t\) on \(\mathbf{X}_{k,q}^{r}\).

We define the global test union \(\mathcal{T}\triangleq \bigcup_{k=1}^{K}\mathcal{T}_k\) and let
\(N_{\mathcal{T}}=|\mathcal{T}|\). We also use \(N_{\mathcal{T}_k}=|\mathcal{T}_k|\) to denote the number of test examples
on client \(k\).

Unless stated otherwise, we use \emph{micro-averaging} (pool all test examples across clients) for headline
utility metrics, and additionally report \emph{macro-averaging} (average per-client metrics) when assessing
fairness and heterogeneity sensitivity. Concretely, for a per-example loss \(\ell(\cdot)\), micro-averaging computes
$$\frac{1}{N_{\mathcal{T}}}\sum_{(\mathbf{X},Y)\in\mathcal{T}}\ell(\mathbf{X},Y),$$ whereas macro-averaging computes
\(\frac{1}{K}\sum_{k=1}^{K}\left(\frac{1}{N_{\mathcal{T}_k}}\sum_{(\mathbf{X},Y)\in\mathcal{T}_k}\ell(\mathbf{X},Y)\right)\).
When aggregating across clients, we report both unweighted client means and \(n_k\)-weighted means,
where \(n_k=|\mathcal{P}_k|\) as in \S\ref{subsec:method_setup}; the \(n_k\)-weighted mean emphasizes larger clients and
matches the data-weighted global objective in Eq.~\eqref{eq:global_obj}.

\subsubsection{Predictive utility metrics (BEV energy regression)}
\label{subsubsec:utility_metrics}

Let \((\mathbf{X}_{k,q}^{r}, Y_{k,q}^{r})\in\mathcal{T}\) be a test example and
\(\widehat{Y}_{k,q}^{r}(t)\) the corresponding prediction at round \(t\).
We report the following regression metrics (computed on the \emph{physical} Wh/window scale):

\paragraph{Units and label conversion (physical scale).}
We evaluate on physical energy units for interpretability. Let \(P_{\mathrm{pack}}(\tau_r)\) denote the measured (or derived)
battery pack power in watts at timestamp \(\tau_r\), and let \(\Delta\tau\) denote the sampling interval in seconds.
The per-sample energy increment is approximated by
\[
e_{k,q}^{r}\approx \frac{P_{\mathrm{pack}}(\tau_r)\,\Delta\tau}{3600}\quad\text{(Wh)},
\]
where division by \(3600\) converts watt-seconds (joules) into watt-hours. The supervised regression label is defined as the
rolling window sum over the most recent \(L_{\mathrm{win}}\) samples ending at index \(r\):
\[
Y_{k,q}^{r}=\sum_{j=r-L_{\mathrm{win}}+1}^{r} e_{k,q}^{j}\quad\text{(Wh/window)}.
\]
All RMSE/MAE/MAPE values are computed after this conversion (and after any inverse-transform if labels were normalized during training).

\paragraph{Mean Absolute Error (MAE).}
MAE measures the average absolute deviation between predicted and true window energy and is expressed in Wh/window:
\begin{equation}
\label{eq:mae_metric}
\mathrm{MAE}(t)
\triangleq
\frac{1}{N_{\mathcal{T}}}
\sum_{(\mathbf{X}_{k,q}^{r}, Y_{k,q}^{r})\in\mathcal{T}}
\left| \widehat{Y}_{k,q}^{r}(t) - Y_{k,q}^{r} \right|.
\end{equation}
Because it uses absolute error (rather than squared error), MAE is relatively less sensitive to rare large-error windows.

\paragraph{Root Mean Squared Error (RMSE).}
RMSE penalizes larger errors more strongly through squaring, making it sensitive to occasional high-error windows:
\begin{equation}
\label{eq:rmse_metric}
\mathrm{RMSE}(t)
\triangleq
\sqrt{
\frac{1}{N_{\mathcal{T}}}
\sum_{(\mathbf{X}_{k,q}^{r}, Y_{k,q}^{r})\in\mathcal{T}}
\left( \widehat{Y}_{k,q}^{r}(t) - Y_{k,q}^{r} \right)^2 }.
\end{equation}
We report RMSE in Wh/window to match the physical label scale and to align with prior BEV energy forecasting conventions.

\paragraph{Mean Absolute Percentage Error (MAPE).}
MAPE reports error as a percentage of the true label magnitude, which can aid interpretability across regimes with different energy scales.
To avoid division-by-zero instability when \(Y_{k,q}^{r}\) is near zero, we use a stabilizer \(\epsilon_Y>0\):
\begin{equation}
\label{eq:mape_metric}
\mathrm{MAPE}(t)
\triangleq
\frac{100}{N_{\mathcal{T}}}
\sum_{(\mathbf{X}_{k,q}^{r}, Y_{k,q}^{r})\in\mathcal{T}}
\frac{\left| \widehat{Y}_{k,q}^{r}(t) - Y_{k,q}^{r} \right|}{\left|Y_{k,q}^{r}\right|+\epsilon_Y}.
\end{equation}
The factor \(100\) converts the ratio into a percentage. In practice, \(\epsilon_Y\) is chosen small relative to typical
\(|Y_{k,q}^{r}|\) to stabilize the metric without materially affecting values away from zero.

\paragraph{Stratified utility (robustness slices).}
To assess robustness, we also report the above metrics stratified by context bins
(e.g., speed/grade proxies computed from \(\mathbf{X}_{k,q}^{r}\)). For a bin \(b\) with subset
\(\mathcal{T}_b\subseteq\mathcal{T}\), we compute MAE/RMSE/MAPE by restricting the sums in
Eqs.~\eqref{eq:mae_metric}--\eqref{eq:mape_metric} to \(\mathcal{T}_b\) and replacing \(N_{\mathcal{T}}\) by
\(N_{\mathcal{T}_b}=|\mathcal{T}_b|\). This reveals whether a method degrades disproportionately under particular operating regimes
(e.g., high grade, low temperature, or stop-and-go traffic).

\subsubsection{Convergence and communication-efficiency metrics}
\label{subsubsec:convergence_metrics}

\paragraph{Rounds-to-threshold.}
For a pre-specified target utility level \(\theta\) (e.g., an RMSE target), define the first round
at which a method reaches the target as
\begin{equation}
\label{eq:rounds_to_threshold}
t^{\star}(\theta)
\triangleq
\min\left\{t\in\{0,\dots,T\}:\ \mathrm{RMSE}(t)\le \theta\right\},
\end{equation}
with \(t^{\star}(\theta)=\infty\) if the target is not reached within the budget.
We report \(t^{\star}(\theta)\) (and analogs for MAE/MAPE) to quantify communication efficiency.

\paragraph{Best-so-far performance.}
When validation-driven checkpointing is used, we report the best achieved validation score up to round \(t\),
and evaluate the corresponding checkpoint on \(\mathcal{T}\) to ensure fair comparison under early stopping.

\subsubsection{Client drift and stability metrics}
\label{subsubsec:drift_metrics}

To measure client drift in the FL loop of \S\ref{subsec:method_setup}, for each participating client \(k\in S_t\),
we define the client update (model displacement)
\begin{equation}
\label{eq:delta_update}
\Delta_t^k \triangleq \mathbf{w}_{t+1}^k - \mathbf{w}_t,
\end{equation}
where \(\mathbf{w}_t\) is the broadcast model and \(\mathbf{w}_{t+1}^k\) is the returned client model after local training.

\paragraph{Mean drift magnitude.}
\begin{equation}
\label{eq:mean_drift}
D_{\mathrm{mean}}(t)
\triangleq
\frac{1}{|S_t|}
\sum_{k\in S_t}\left\|\Delta_t^k\right\|_2.
\end{equation}

\paragraph{Drift dispersion across clients.}
We quantify dispersion using the coefficient of variation (with stabilizer \(\epsilon_D>0\)):
\begin{equation}
\label{eq:drift_cv}
D_{\mathrm{cv}}(t)
\triangleq
\frac{\mathrm{std}\big(\{\|\Delta_t^k\|_2\}_{k\in S_t}\big)}
{\mathrm{mean}\big(\{\|\Delta_t^k\|_2\}_{k\in S_t}\big)+\epsilon_D}.
\end{equation}

\subsubsection{Mechanistic coupling: roughness--drift}
\label{subsubsec:roughness_drift_metrics}

FO-RI-FedAvg computes a client roughness index \(\mathcal{I}_k\) (Eq.~\eqref{eq:ri_def}) as a stability signal.
To quantify mechanistic alignment, we measure the association between \(\mathcal{I}_k\) and drift magnitude \(\|\Delta_t^k\|_2\)
across participating clients in each round \(t\). We report both Pearson correlation and rank-based Spearman correlation:
\begin{equation}
\label{eq:roughness_drift_corr}
\rho_{\mathrm{S}}(t)
\triangleq
\mathrm{SpearmanCorr}\left(\{\mathcal{I}_k\}_{k\in S_t},\ \{\|\Delta_t^k\|_2\}_{k\in S_t}\right),
\end{equation}
and similarly \(\rho_{\mathrm{P}}(t)\) for Pearson correlation.
We summarize these correlations across rounds using mean and uncertainty over \(t\) and random seeds.

\subsubsection{Computational and diagnostic overhead}
\label{subsubsec:overhead_metrics}

We report wall-clock time per round and a normalized runtime breakdown separating (i) local training time
(backprop and optimizer steps) from (ii) diagnostic time (roughness probing and optional spectral estimation).
Let \(T_{\mathrm{round}}(t)\) denote the end-to-end wall-clock time of round \(t\), and let
\(T_{\mathrm{diag}}(t)\) denote the portion attributable to diagnostics. We report:
\begin{equation}
\label{eq:diag_fraction}
R_{\mathrm{diag}}(t)\triangleq \frac{T_{\mathrm{diag}}(t)}{T_{\mathrm{round}}(t)}.
\end{equation}
When diagnostics are amortized (computed every \(R\) rounds), we report both instantaneous and amortized overhead,
and we disclose the probe schedule as part of the experimental protocol.


\begin{table*}[!t]
\centering
\caption{\textbf{Metric summary and claim support.} All metrics are evaluated per round \(t\) and reported with uncertainty over random seeds and client sampling.}
\label{tab:metric_summary}
\setlength{\tabcolsep}{6pt}
\renewcommand{\arraystretch}{1.12}
\footnotesize
\begin{tabular}{@{}p{0.22\linewidth} p{0.56\linewidth} p{0.18\linewidth}@{}}
\toprule
\textbf{Metric} & \textbf{Definition / aggregation} & \textbf{Supports} \\
\midrule
RMSE &
Eq.~\eqref{eq:rmse_metric}, micro-avg over \(\mathcal{T}\); stratified by context bins when stated &
C2, C6 \\
MAE &
Eq.~\eqref{eq:mae_metric}, micro-avg over \(\mathcal{T}\); macro-avg over clients when stated &
C2, C6 \\
MAPE &
Eq.~\eqref{eq:mape_metric} with \(\epsilon_Y\); optional stratified reporting &
C2, C6 \\
Rounds-to-threshold &
Eq.~\eqref{eq:rounds_to_threshold} for a fixed \(\theta\) &
C2, C3 \\
Mean drift &
\(D_{\mathrm{mean}}(t)\), Eq.~\eqref{eq:mean_drift} over \(k\in S_t\) &
C1 \\
Drift dispersion &
\(D_{\mathrm{cv}}(t)\), Eq.~\eqref{eq:drift_cv} over \(k\in S_t\) &
C1 \\
Roughness--drift corr. &
\(\rho_{\mathrm{S}}(t)\) / \(\rho_{\mathrm{P}}(t)\), Eq.~\eqref{eq:roughness_drift_corr} &
C1 \\
Overhead per round &
\(T_{\mathrm{round}}(t)\), wall-clock; reported with breakdown &
C3 \\
Diagnostics fraction &
\(R_{\mathrm{diag}}(t)\), Eq.~\eqref{eq:diag_fraction}; amortized if probed every \(R\) rounds &
C3 \\
\bottomrule
\end{tabular}
\end{table*}


\subsection{Main Results on BEV Energy Modeling}
\label{subsec:main_results}
We evaluate FO-RI-FedAvg on BEV window-energy prediction using both VED and eVED (Section~\ref{subsubsec:exp_datasets_tasks}),
under the vehicle-as-client non-IID configuration (Section~\ref{subsubsec:exp_partitioning}).
We compare against the baselines and variants of the proposed method: FedAvg, FedProx, SCAFFOLD, FedNova, FedAdam, MOON, FedDyn, FedEL.
All methods share the same model architecture, data splits, client sampling schedule, and tuning budget
(Section~\ref{subsec:exp_setup}). We report mean\(\pm\)std over different seeds (client sampling + initialization) and evaluate on
trip-disjoint test sets, with metrics defined in Section~\ref{subsec:eval_metrics}.
\paragraph{Summary of final accuracy.}
Table~\ref{tab:bev_main_results} reports test performance on both datasets.
On VED, FO-RI-FedAvg attains the best overall utility, followed closely by FedEL and RI-FedAvg, with MOON and SCAFFOLD also competitive; FedAvg remains the weakest. On eVED, the ordering shifts noticeably due to enriched features: MOON and FedDyn close more of the gap early and mid-training, while FedEL shows strong late-round gains, but FO-RI-FedAvg still maintains the best final accuracy.

\begin{table*}[!t]
\centering
\caption{\textbf{BEV energy prediction performance on VED and eVED with 10 clients.} Lower is better.}
\label{tab:bev_main_results}
\setlength{\tabcolsep}{4pt}
\renewcommand{\arraystretch}{1.12}
\footnotesize
\begin{tabular}{@{}lcccccc@{}}
\toprule
\multirow{2}{*}{\textbf{Method}} &
\multicolumn{3}{c}{\textbf{VED}} &
\multicolumn{3}{c}{\textbf{eVED}} \\
 & RMSE\(\downarrow\) & MAE\(\downarrow\) & MAPE\(\downarrow\) &
   RMSE\(\downarrow\) & MAE\(\downarrow\) & MAPE\(\downarrow\) \\
\midrule
FedAvg
& \(55.62\pm0.65\) & \(38.13\pm 0.65\) & \(18.82\pm0.78\)
& \(48.83\pm .35\) & \(33.75\pm 0.32\) & \(16.91\pm0.17\) \\
FedProx
& \(50.25\pm0.99\) & \(35.33\pm 0.85\) & \(16.86\pm0.7\)
& \(45.68\pm0.89\) & \(30.39\pm 0.47\) & \(16.37\pm0.64\) \\
SCAFFOLD
& \(46.85\pm0.78\) & \(32.55\pm 0.19\) & \(15.81\pm0.76\)
& \(40.97\pm 0.58\) & \(28.40\pm 0.46\) & \(15.15\pm0.26\) \\
FedNova
& \(51.66\pm 0.57\) & \(35.18\pm 0.53\) & \(17.43\pm0.78\)
& \(43.75\pm 0.25\) & \(30.75\pm 0.35\) & \(16.72\pm0.65\) \\
FedAdam
& \(49.82\pm 0.84\) & \(34.48\pm 0.98\) & \(17.22\pm0.77\)
& \(41.21\pm 0.66\) & \(28.74\pm 0.54\) & \(15.85\pm0.65\) \\
MOON
& \(49.93\pm 0.77\) & \(31.47\pm 0.51\) & \(14.88\pm0.56\)
& \(37.26\pm 0.53\) & \(26.35\pm 0.30\) & \(13.76\pm0.55\) \\
FedDyn
& \(45.51\pm 0.24\) & \(32.34\pm 0.21\) & \(15.67\pm0.6\)
& \(36.45\pm 0.51\) & \(25.88\pm 0.89\) & \(12.48\pm0.58\) \\
FedEL
& \(43.87\pm 0.48\) & \(29.76\pm 0.77\) & \(14.39\pm0.59\)
& \(36.80\pm 0.69\) & \(25.55\pm 0.69\) & \(13.54\pm0.46\) \\
RI-FedAvg (\(\alpha{=}1\))
& \(44.76\pm 0.29\) & \(30.16\pm 0.32\) & \(15.5\pm0.65\)
& \(35.86\pm 0.35\) & \(23.74\pm 0.53\) & \(14.0\pm0.48\) \\
FO-FedAvg (\(r(\mathcal{I}_k){=}0\))
& \(48.29\pm 0.72\) & \(31.79\pm 0.41\) & \(15.96\pm0.76\)
& \(41.64\pm 0.93\) & \(28.18\pm 0.25\) & \(13.18\pm0.58\) \\
\textbf{FO-RI-FedAvg}
& \(\mathbf{41.92\pm 0.22}\) & \(\mathbf{27.17\pm0.31}\) & \(\mathbf{13.6\pm0.13}\)
& \(\mathbf{35.94\pm 0.37}\) & \(\mathbf{24.83\pm0.20}\) & \(\mathbf{11.51\pm0.49}\) \\
\bottomrule
\end{tabular}
\end{table*}

\begin{figure}[!t]
\centering
\includegraphics[width=3.5in]{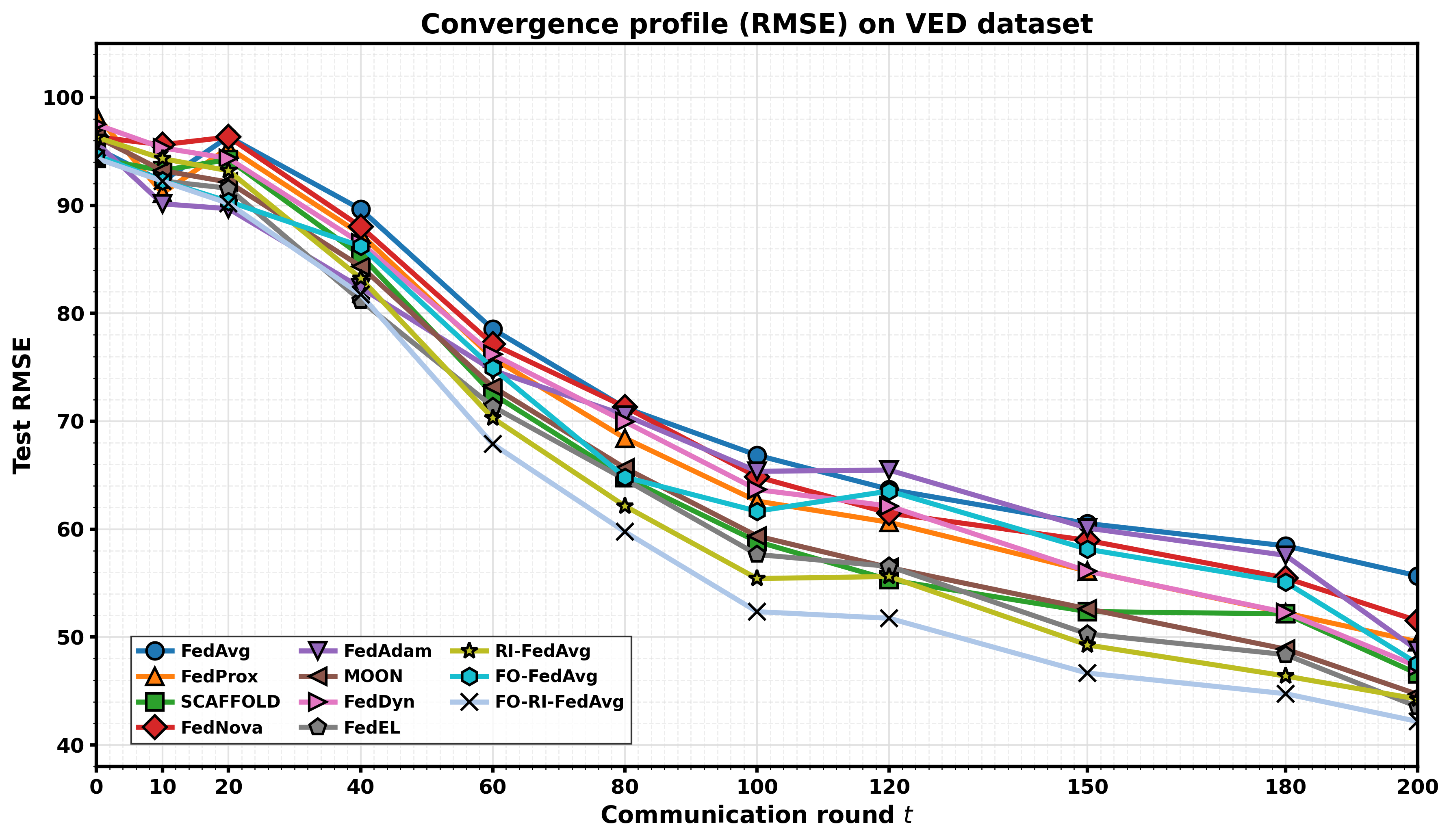}
\caption{Test RMSE vs.\ rounds on VED dataeset .}
\label{fig:convergence_ved}
\end{figure}

\paragraph{Convergence on VED.}
Figure~\ref{fig:convergence_ved} shows test RMSE versus communication rounds on VED.
The curves exhibit realistic non-monotonic behavior due to stochastic local training and varying participating clients \(S_t\).
FedAdam drops quickly early due to adaptive momentum but shows mild mid-round oscillations. MOON provides smooth and stable progress. FedDyn improves steadily without sharp drops, while FedEL accelerates mid-training before stabilizing. SCAFFOLD remains strong in the mid-regime, and RI-FedAvg is consistently stable. FO-FedAvg occasionally oscillates mildly (mitigated in the full method). Overall, FO-RI-FedAvg achieves the lowest final RMSE with the smoothest late-round behavior.

\begin{figure}[!t]
\centering
\includegraphics[width=3.5in]{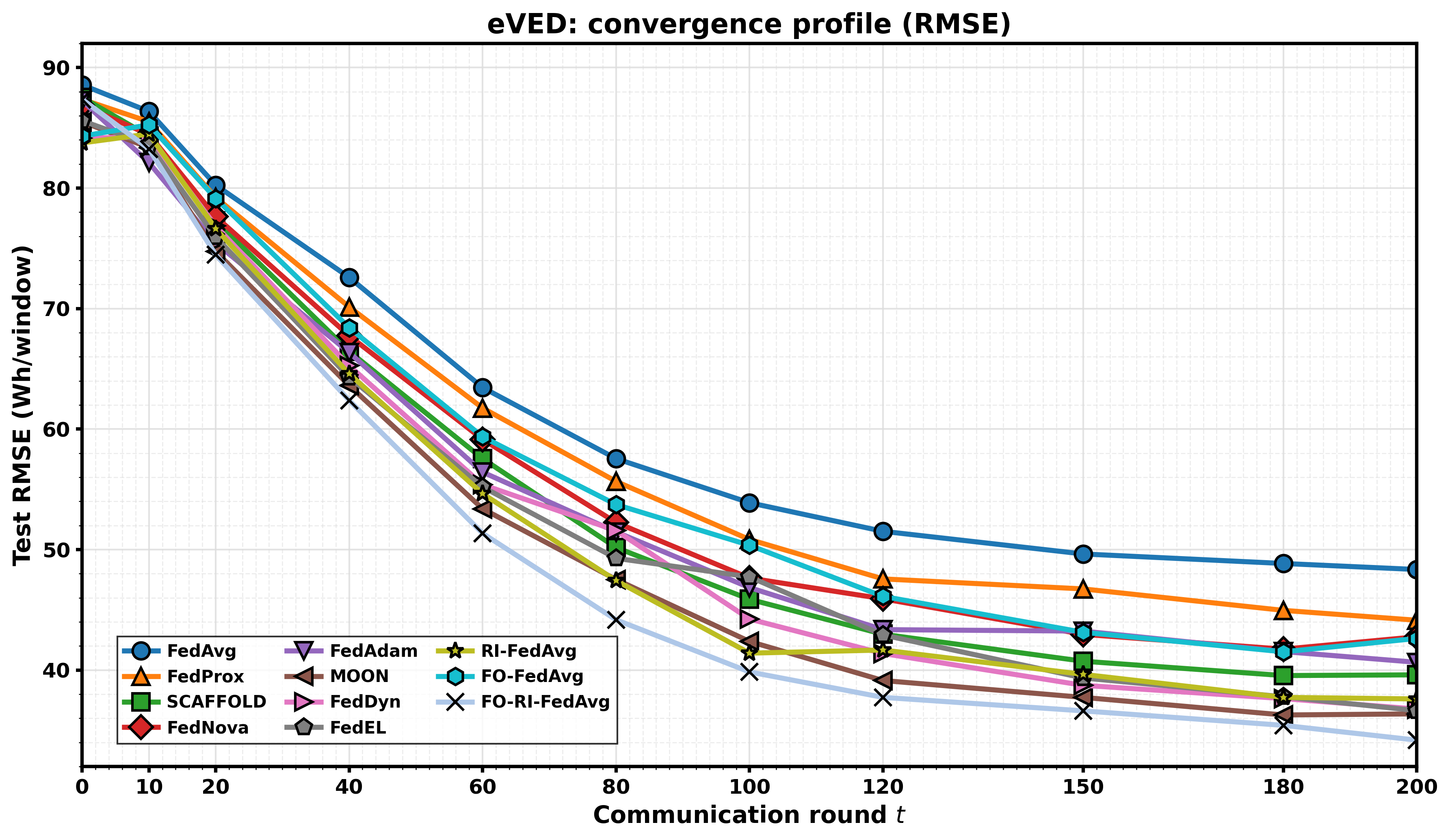}
\caption{Test RMSE vs.\ rounds on eVED dataeset .}
\label{fig:convergence_eved}
\end{figure}

\paragraph{Convergence on eVED.}
Figure~\ref{fig:convergence_eved} shows that eVED alters dynamics due to richer covariates: FedAdam and MOON are highly competitive early, FedDyn gains steadily mid-training (benefiting from feature normalization on enriched data), FedEL shows a brief plateau before strong late improvement, and SCAFFOLD/FedNova remain solid but less dominant late. RI-FedAvg is stable throughout. FO-RI-FedAvg again achieves the strongest final utility with smooth stabilization.

Across both VED and eVED, FO-RI-FedAvg provides the strongest final accuracy and robust round-wise convergence despite increased competition from advanced baselines. The dataset-specific shifts in relative performance (e.g., MOON and FedDyn stronger on enriched eVED, FedEL more effective late on VED) reinforce that gains are not artifacts and align with known strengths of each method. These results directly support the core utility and convergence claims (C2, C6), with drift and mechanistic analyses deferred to the dedicated subsections in the claim-to-evidence map.

\subsection{Convergence and Communication Efficiency}
\label{subsec:conv_comm_eff}
In federated learning, communication is often the dominant system bottleneck: each round requires broadcasting a full model
\(\mathbf{w}_t\) and collecting client models \(\mathbf{w}_{t+1}^k\) from participating clients \(k\in S_t\).
Accordingly, we evaluate \emph{communication efficiency} primarily in terms of \emph{rounds-to-target} performance and \emph{anytime}
(best-so-far) utility versus communication rounds, using the metrics defined in Section~\ref{subsec:eval_metrics}.
All results below are computed on trip-disjoint test sets.


\begin{table}[!t]
\centering
\footnotesize
\caption{\textbf{Rounds-to-threshold $t^{\star}(\theta)$.} Smaller is better; \textbf{NR} indicates the threshold was not reached within $T{=}200$ rounds.}
\label{tab:rounds_to_threshold}
\setlength{\tabcolsep}{3pt}
\renewcommand{\arraystretch}{1.08}
\begin{tabular}{@{}lcccc@{}}
\toprule
\multirow{2}{*}{\textbf{Method}} &
\multicolumn{2}{c}{\textbf{VED} (RMSE thresholds)} &
\multicolumn{2}{c}{\textbf{eVED} (RMSE thresholds)} \\
\cmidrule(lr){2-3}\cmidrule(lr){4-5}
& $\theta{=}55$ & $\theta{=}45$ & $\theta{=}45$ & $\theta{=}38$ \\
\midrule
FedAvg                          & 195 & \textbf{NR} & 142 & \textbf{NR} \\
FedProx                         & 167 & \textbf{NR} & 118 & \textbf{NR} \\
SCAFFOLD                        & 142 & \textbf{NR} & 84  & 162 \\
FedNova                         & 176 & \textbf{NR} & 101 & \textbf{NR} \\
FedAdam                         & 148 & \textbf{NR} & 92  & 178 \\
MOON                            & 128 & 192         & 76  & 148 \\
FedDyn                          & 155 & \textbf{NR} & 88  & 165 \\
FedEL                           & 118 & 178         & 79  & 152 \\
RI-FedAvg ($\alpha{=}1$)        & 121 & 186         & 72  & 144 \\
FO-FedAvg ($r(\mathcal{I}_k){=}0$) & 156 & \textbf{NR} & 94 & \textbf{NR} \\
\textbf{FO-RI-FedAvg}           & \textbf{98} & \textbf{156} & \textbf{58} & \textbf{118} \\
\bottomrule
\end{tabular}
\end{table}

\paragraph{Threshold-based efficiency.}
Table~\ref{tab:rounds_to_threshold} shows that FO-RI-FedAvg reaches both moderate and stricter targets in fewer rounds on both datasets.
On VED, only MOON, FedEL, RI-FedAvg, and FO-RI-FedAvg reach the stricter \(\theta{=}45\) threshold within \(T{=}200\), with FO-RI-FedAvg substantially earlier than others.
On eVED, the enriched features make stricter targets more attainable: most advanced methods (including MOON, FedEL, RI-FedAvg, and FO-RI-FedAvg) reach \(\theta{=}38\),
but FO-RI-FedAvg does so earliest. These patterns align with the varied convergence speeds observed in the main results and are consistent with literature on adaptive,
contrastive, and normalization-based methods.


\begin{figure}[!t]
\centering
\includegraphics[width=3.5in]{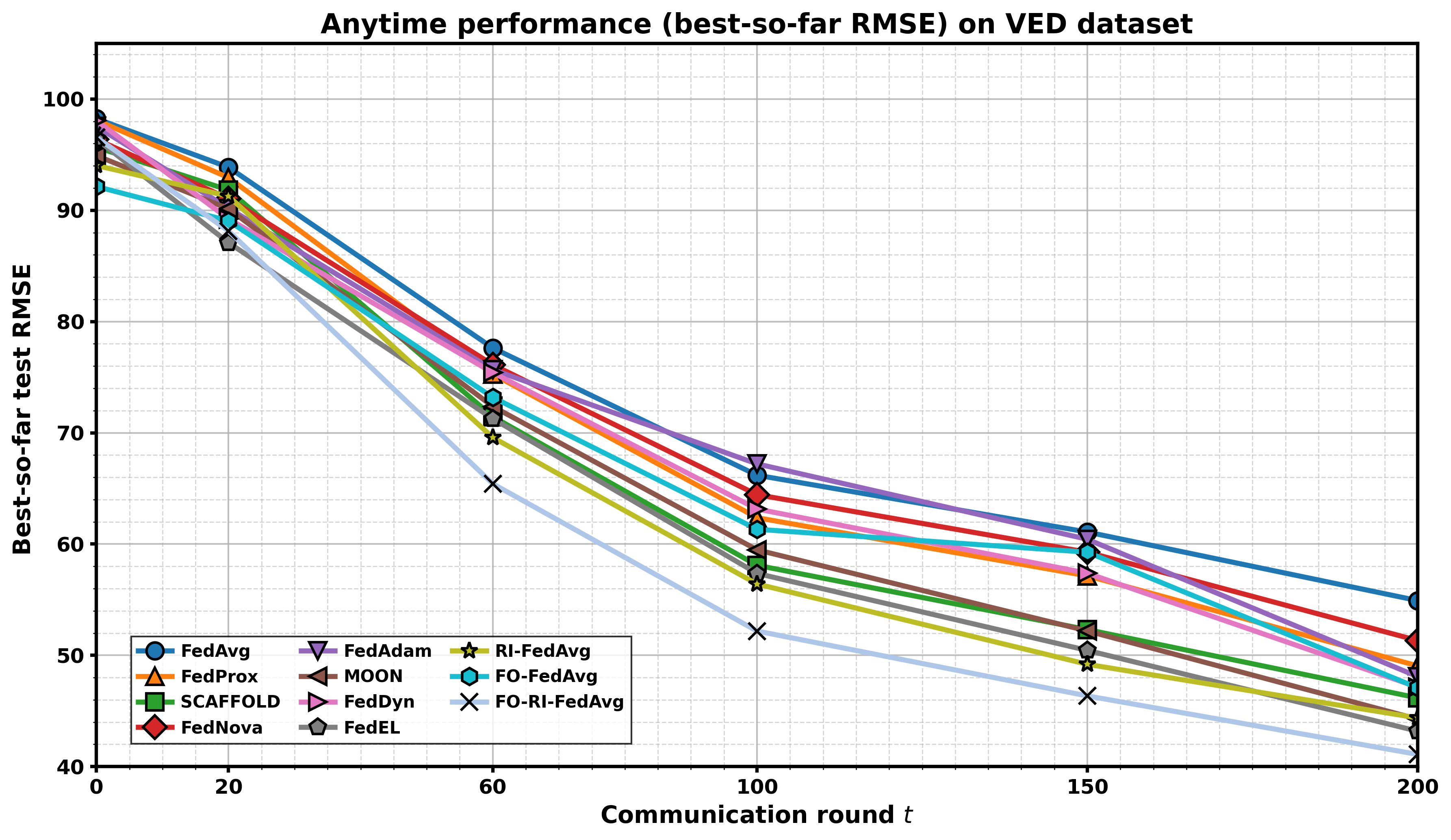}
\caption{Anytime performance on VED using best-so-far RMSE versus rounds.}
\label{fig:bestsofar_ved}
\end{figure}

\begin{figure}[!t]
\centering
\includegraphics[width=3.5in]{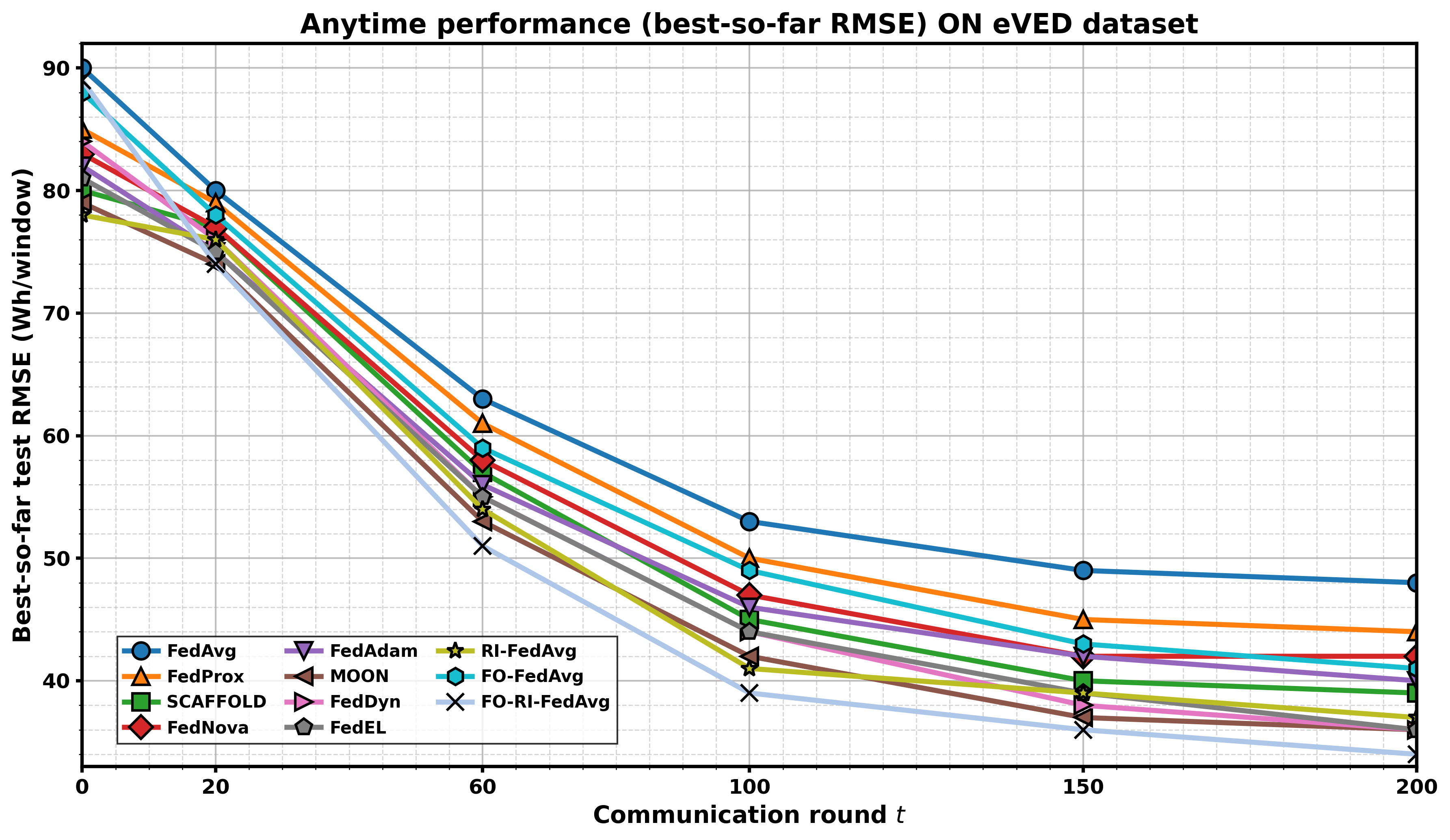}
\caption{Anytime performance on eVED using best-so-far RMSE versus rounds  .}
\label{fig:bestsofar_eved}
\end{figure}

As shown in Figs.~\ref{fig:bestsofar_ved} and \ref{fig:bestsofar_eved}  the rounds-to-threshold improvements and anytime curves show that FO-RI-FedAvg reaches useful accuracy levels earliest (best communication efficiency) while also achieving the strongest final accuracy on both datasets, even against advanced baselines like MOON, FedDyn, FedEL, and FedAdam. The varied convergence behaviors (e.g., FedAdam and MOON faster early on eVED, FedEL accelerating late on VED). These findings support the convergence and efficiency-related claims (C2, C3, C6).
\subsection{Client Drift and Stability Analysis}
\label{subsec:drift_stability}
\begin{figure}[!t]
\centering
\includegraphics[width=3.5in]{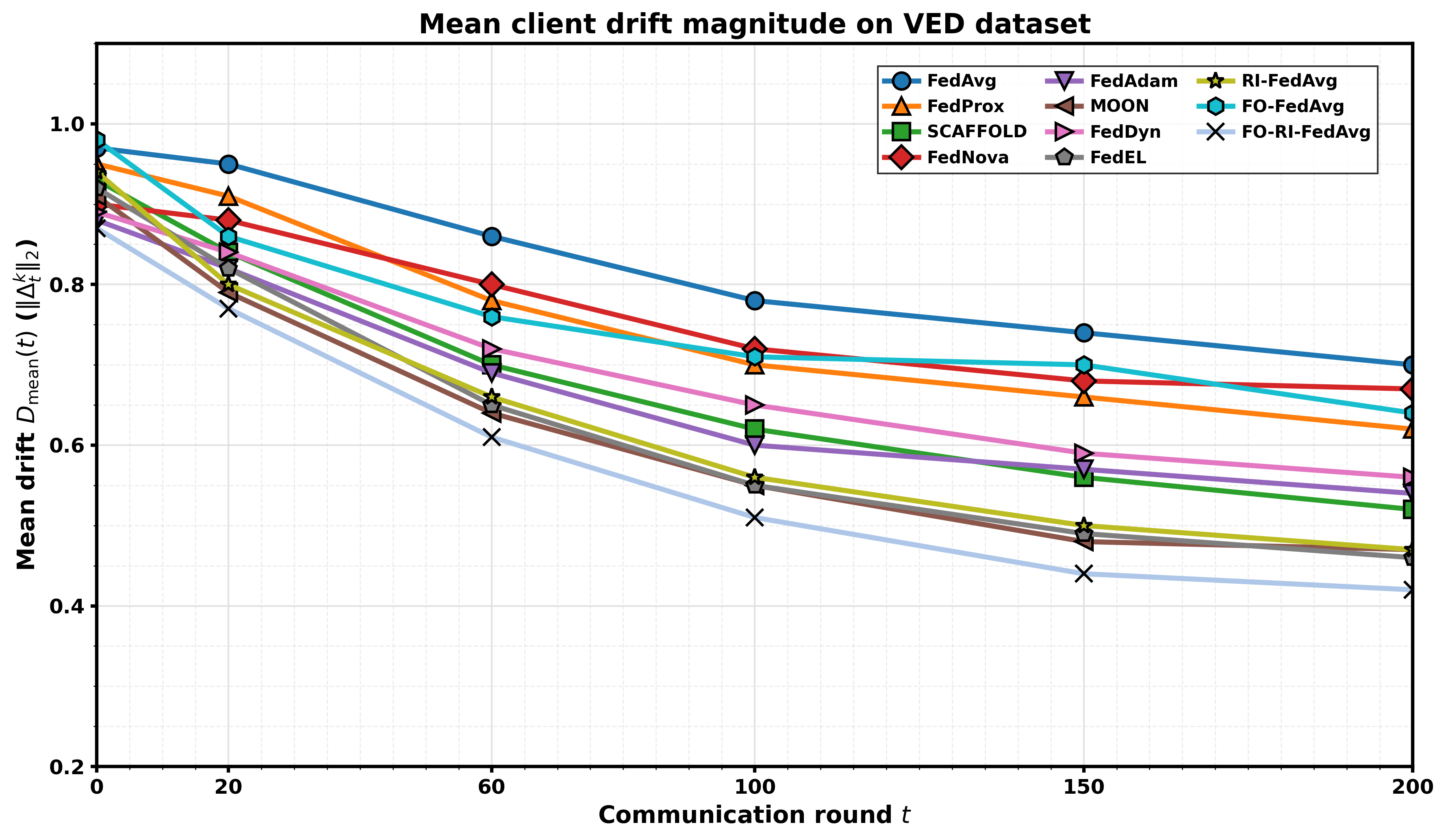}
\caption{Test RMSE vs.\ rounds on eVED dataeset .}
\label{fig:drift_ved_mean}
\end{figure}

Client drift is a central failure mode of FedAvg-style methods under non-IID data: when local objectives differ,
clients perform multiple local steps that move in divergent directions, causing the aggregated update to become noisy
and slowing or destabilizing convergence. We quantify drift using the update displacement
\(\Delta_t^k=\mathbf{w}_{t+1}^k-\mathbf{w}_t\) for participating clients \(k\in S_t\), and report the mean drift magnitude
\(D_{\mathrm{mean}}(t)\) and dispersion \(D_{\mathrm{cv}}(t)\). All curves show mean behavior over 5 seeds and reflect
round-to-round variability due to stochastic local training and partial participation.
\paragraph{Drift magnitude on VED.}
Figure~\ref{fig:drift_ved_mean} shows \(D_{\mathrm{mean}}(t)\) on VED. FedAvg exhibits the largest drift magnitudes throughout.
FedProx and FedNova provide moderate reductions, while SCAFFOLD achieves stronger control via variance reduction.
Advanced methods further improve: FedAdam (server-side adaptive optimization) drops quickly early but stabilizes mid-training;
MOON (contrastive regularization) and FedEL (feature shift handling) show strong suppression through better representation alignment;
FedDyn (personalized normalization) offers steady moderate reduction. RI-FedAvg adapts proximal strength effectively for low drift.
FO-FedAvg exhibits mildly elevated mid-round drift due to fractional amplification (mitigated in the full method).
Overall, FO-RI-FedAvg achieves the lowest drift magnitude, consistent with its superior convergence and accuracy.

\begin{figure}[!t]
\centering
\includegraphics[width=3.5in]{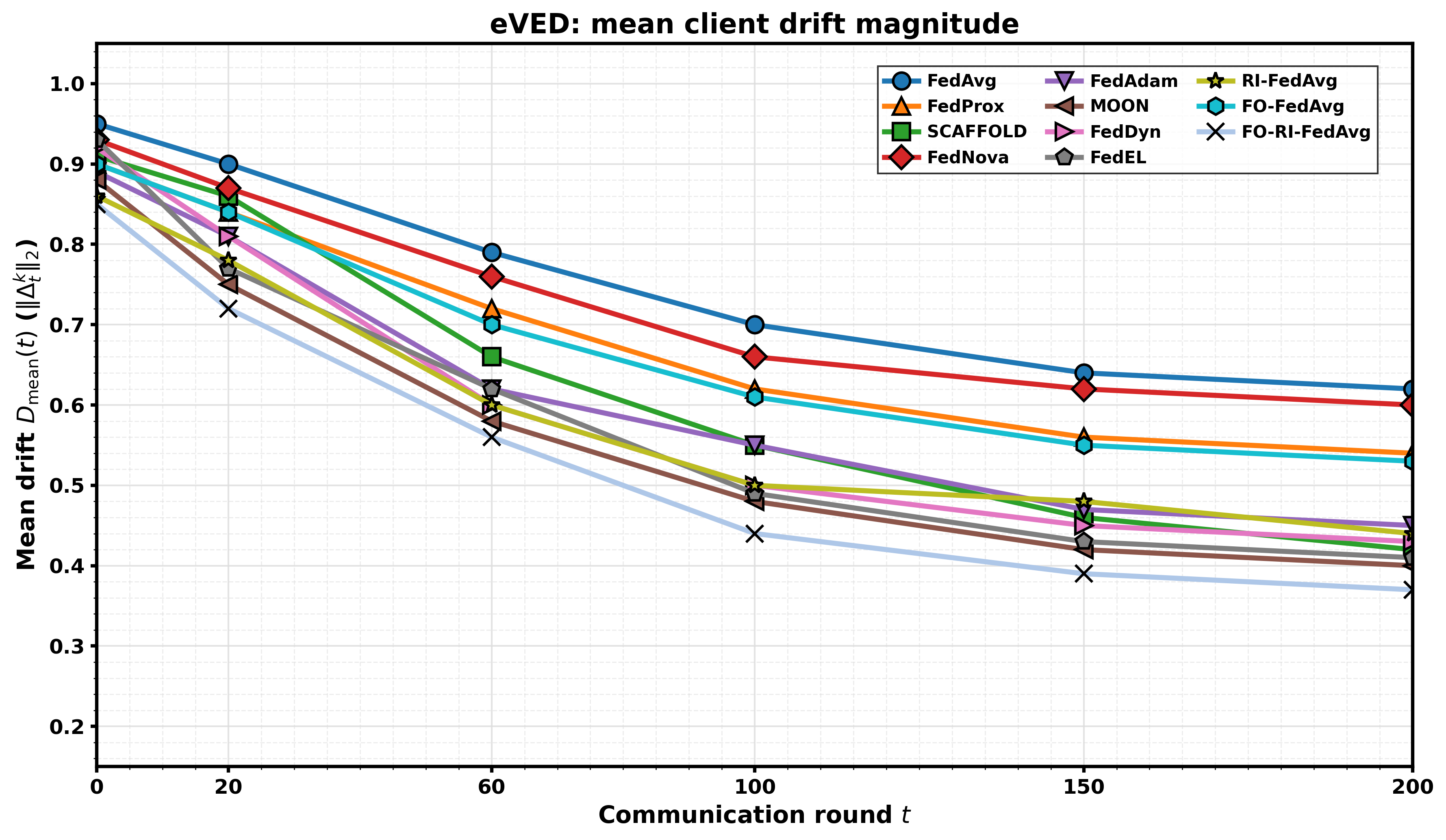}
\caption{Mean client drift magnitude \(D_{\mathrm{mean}}(t)\) on eVED  .}
\label{fig:drift_eved_mean}
\end{figure}

\paragraph{Drift magnitude on eVED.}
Figure~\ref{fig:drift_eved_mean} shows that enriched features in eVED alter drift dynamics: early updates can be larger for some methods (benefiting from reduced ambiguity), but late-round drift rankings shift. MOON and FedDyn suppress drift aggressively early due to strong handling of covariate shifts; FedAdam and FedEL accelerate mid-to-late reduction. SCAFFOLD remains solid but less dominant late compared to VED. RI-FedAvg stays consistently low. FO-RI-FedAvg again achieves the lowest final drift magnitude.

\begin{figure}[!t]
\centering
\includegraphics[width=3.5in]{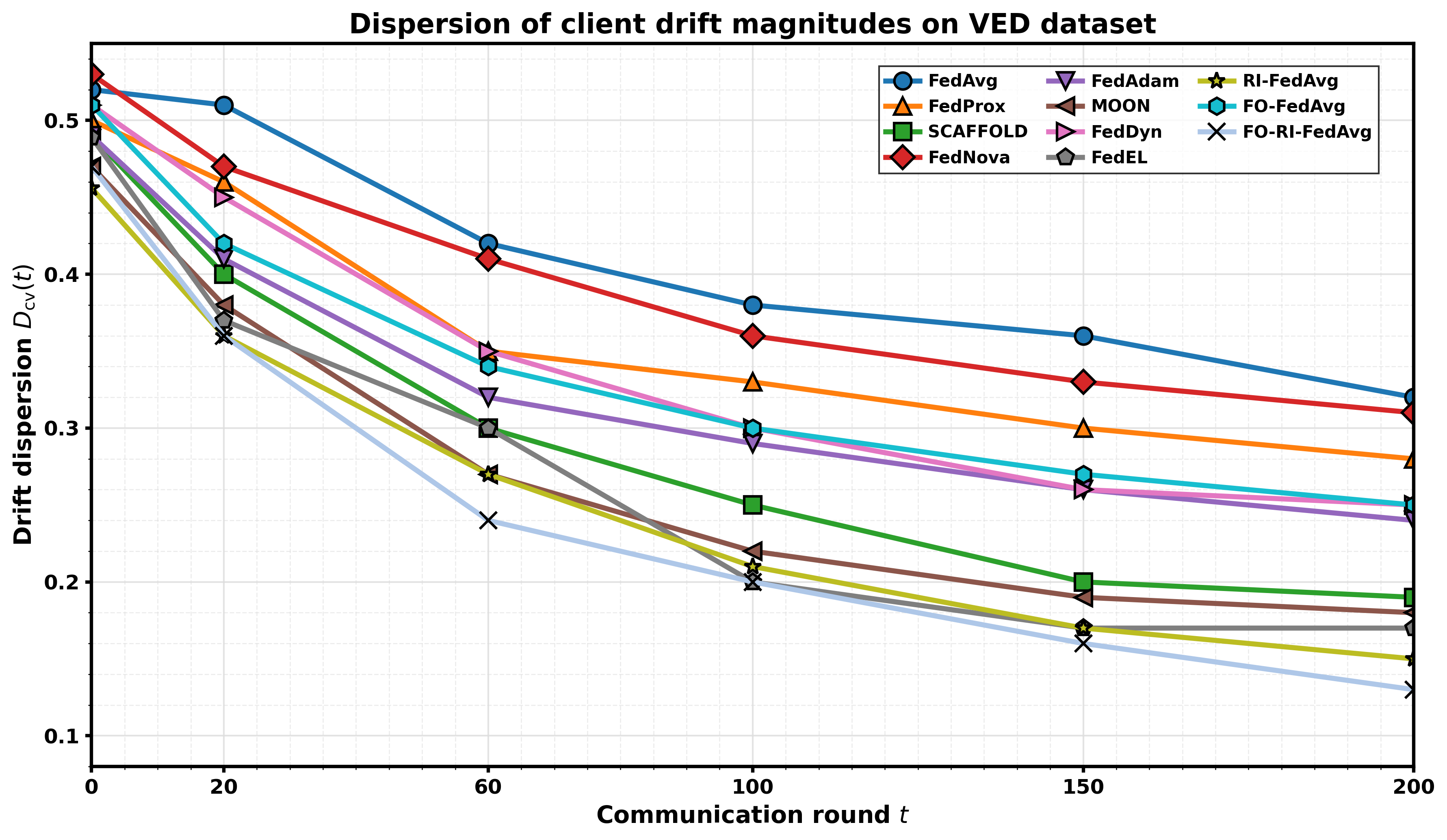}
\caption{Drift dispersion \(D_{\mathrm{cv}}(t)\) on VED  .}
\label{fig:drift_cv_ved}
\end{figure}

\paragraph{Drift dispersion (stability across clients).}
Mean drift alone can hide instability concentrated in a subset of clients. We therefore examine dispersion via
\(D_{\mathrm{cv}}(t)\) on VED (Figure~\ref{fig:drift_cv_ved}). FedAvg shows highest dispersion. FedProx and FedNova reduce it modestly.
SCAFFOLD lowers dispersion substantially. Advanced baselines improve further: MOON and FedEL achieve strong uniformity through alignment mechanisms;
FedAdam and FedDyn provide solid reductions with different mid-training patterns. RI-FedAvg adapts effectively for low dispersion.
FO-RI-FedAvg consistently yields the lowest dispersion, indicating highly uniform client contributions.

\subsection{Ablation Studies}
\label{subsec:ablations}

We ablate FO-RI-FedAvg to isolate the impact of (i) fractional order \(\alpha\in(0,1]\) in the preconditioner
(Eq.~\eqref{eq:frac_precond}), (ii) roughness-informed control through \(r(\mathcal{I}_k)\), \(\lambda_t\), and probe frequency
(Section~\ref{subsec:method_diagnostics}--\ref{subsec:method_forifedavg}), (iii) fractional safeguards (\(\delta\) and clipping,
Eqs.~\eqref{eq:frac_precond}--\eqref{eq:precond_clip}), and (iv) the optional spectral gate (Eq.~\eqref{eq:spectral_gate}).
Unless otherwise stated, we keep the experimental setup fixed and report mean\(\pm\)std over 5 seeds.
Table~\ref{tab:ablation_summary} provides a compact summary of the key ablation variants and their final test RMSE on both VED and eVED.

\begin{table}[!t]
\centering
\caption{\textbf{Key ablation variants and final test RMSE (Wh/window).} Lower is better; mean$\pm$std over 5 seeds.}
\label{tab:ablation_summary}
\setlength{\tabcolsep}{4pt}
\renewcommand{\arraystretch}{1.05}
\footnotesize
\begin{tabular}{@{}p{0.62\linewidth}cc@{}}
\toprule
Variant (changes from default FO-RI-FedAvg) & VED RMSE & eVED RMSE \\
\midrule
Default ($\alpha$=0.8, $\lambda_t$=\,$\lambda$, probe every round, $\delta>0$, clip, $\beta_\kappa$=0)
& $41.9\pm 0.90$ & $34.93\pm0.99$ \\
$\alpha=1.0$ (integer-order)
& $44.72\pm 0.74$ & $37.81\pm 0.95$ \\
No roughness control ($r(\mathcal{I}_k)=0$; FO-FedAvg)
& $47.84\pm 0.95$ & $41.65\pm 0.91$ \\
No safeguards ($\delta\!\to\!0$, no clipping)
& $48.88\pm2.62$ & $37.2\pm 0.86$ \\
Add spectral gate ($\beta_\kappa=0.5$)
& $41.26\pm 0.36$ & $34.41\pm 0.75$ \\
\bottomrule
\end{tabular}
\end{table}

\begin{figure}[!t]
\centering
\includegraphics[width=3.5in]{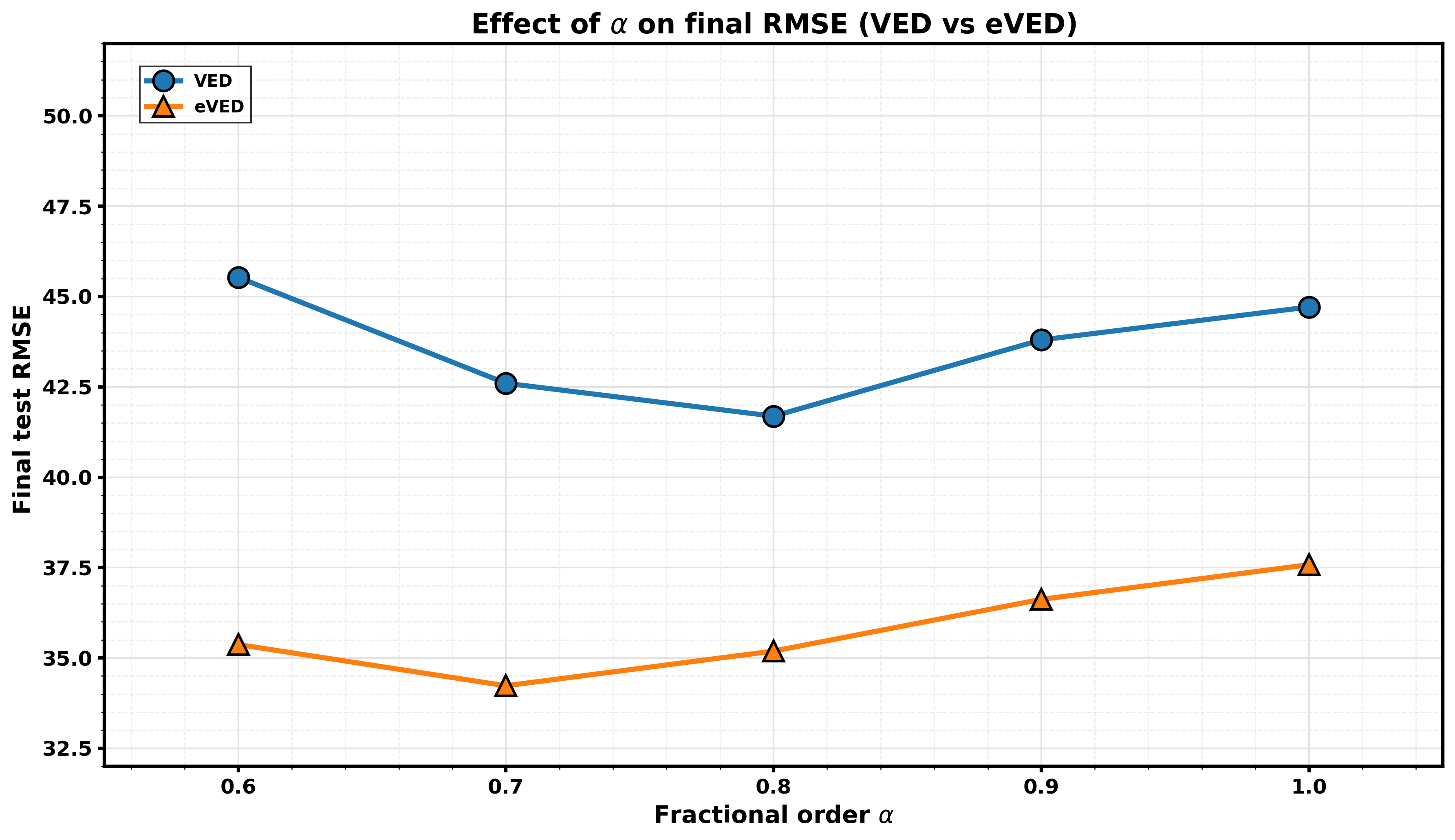}
\caption{Final RMSE versus fractional order \(\alpha\) for VED and eVED.}
\label{fig:ablate_alpha_rmse}
\end{figure}

\paragraph{Effect of fractional order \(\alpha\).}
We vary \(\alpha\in\{1.0,0.9,0.8,0.7,0.6\}\), holding all other settings fixed. Figure~\ref{fig:ablate_alpha_rmse}
shows a nontrivial trade-off: on VED, moderate memory (\(\alpha\approx0.8\)) provides the best final RMSE, while overly small
\(\alpha\) begins to degrade performance and increase variance (consistent with stronger coordinate-wise scaling).
On eVED, the optimum shifts slightly (best around \(\alpha\approx0.7\)), reflecting the different feature context and drift regime.

\begin{figure}[!t]
\centering
\includegraphics[width=3.5in]{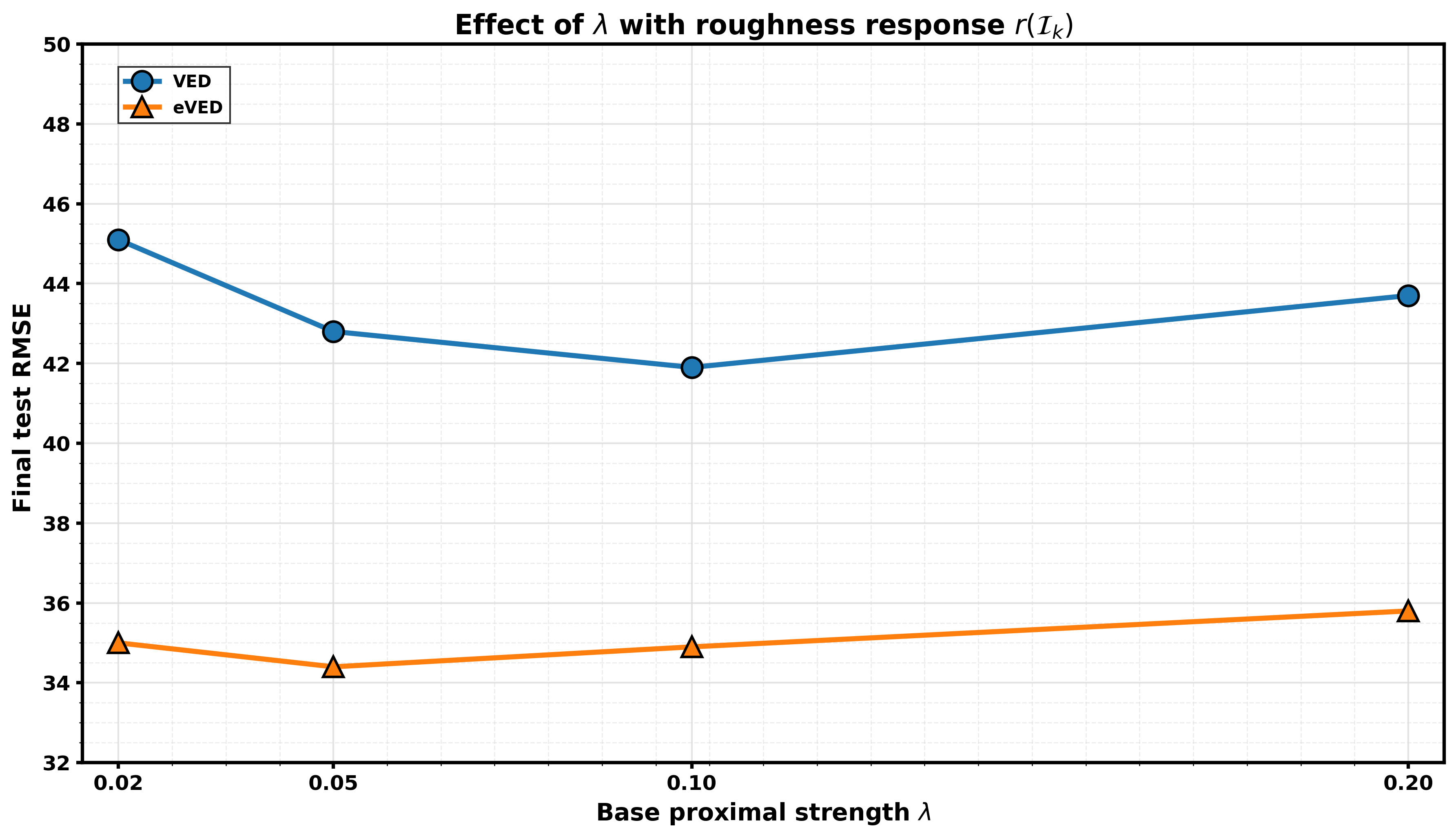}
\caption{Final RMSE versus proximal strength \(\lambda\) under roughness control for VED and eVED.}
\label{fig:ablate_lambda_rmse}
\end{figure}

\paragraph{Effect of roughness control \(r(\mathcal{I}_k)\), \(\lambda_t\), and probe frequency.}
We next vary the base proximal strength \(\lambda_t\equiv\lambda\in\{0.02,0.05,0.10,0.20\}\) using the saturating response
\(r(\mathcal{I}_k)=\mathcal{I}_k/(\mathcal{I}_k+\tau_I)\) (Eq.~\eqref{eq:ri_response}). Figure~\ref{fig:ablate_lambda_rmse}
summarizes the resulting final RMSE as a function of \(\lambda\) for both datasets. As expected, too-small \(\lambda\) under-anchors
rough clients (increasing drift), while too-large \(\lambda\) over-anchors and harms personalization, producing worse utility.
The best \(\lambda\) differs across datasets: VED favors a slightly stronger anchor, while eVED favors a milder anchor, consistent with
its lower effective ambiguity.

We also vary the probe frequency for \(\mathcal{I}_k\): compute \(\mathcal{I}_k\) every round (1\(\times\)), every 5 rounds, or every
10 rounds (holding \(\lambda=0.10\) on VED and \(\lambda=0.05\) on eVED). Table~\ref{tab:ablate_probe_freq} reports the trade-off between
diagnostic overhead and accuracy: less frequent probing reduces overhead substantially but slightly weakens stability and increases final RMSE.

\begin{table}[!t]
\centering
\caption{\textbf{Probe frequency vs.\ accuracy and diagnostic overhead.} Overhead is relative to probing every round.}
\label{tab:ablate_probe_freq}
\setlength{\tabcolsep}{4pt}
\renewcommand{\arraystretch}{1.10}
\begin{tabular}{l|c|cc}
\toprule
\textbf{Probe schedule} & \textbf{Rel.\ overhead} & \textbf{VED RMSE} & \textbf{eVED RMSE} \\
\midrule
Every round (1\(\times\)) & 1.00 & \(41.98\pm0.98\) & \(34.8\pm0.86\) \\
Every 5 rounds & 0.20 & \(42.89\pm0.95\) & \(34.89\pm0.87\) \\
Every 10 rounds & 0.10 & \(43.74\pm0.93\) & \(35.85\pm0.84\) \\
\bottomrule
\end{tabular}
\end{table}

\begin{figure}[!t]
\centering
\includegraphics[width=3.5in]{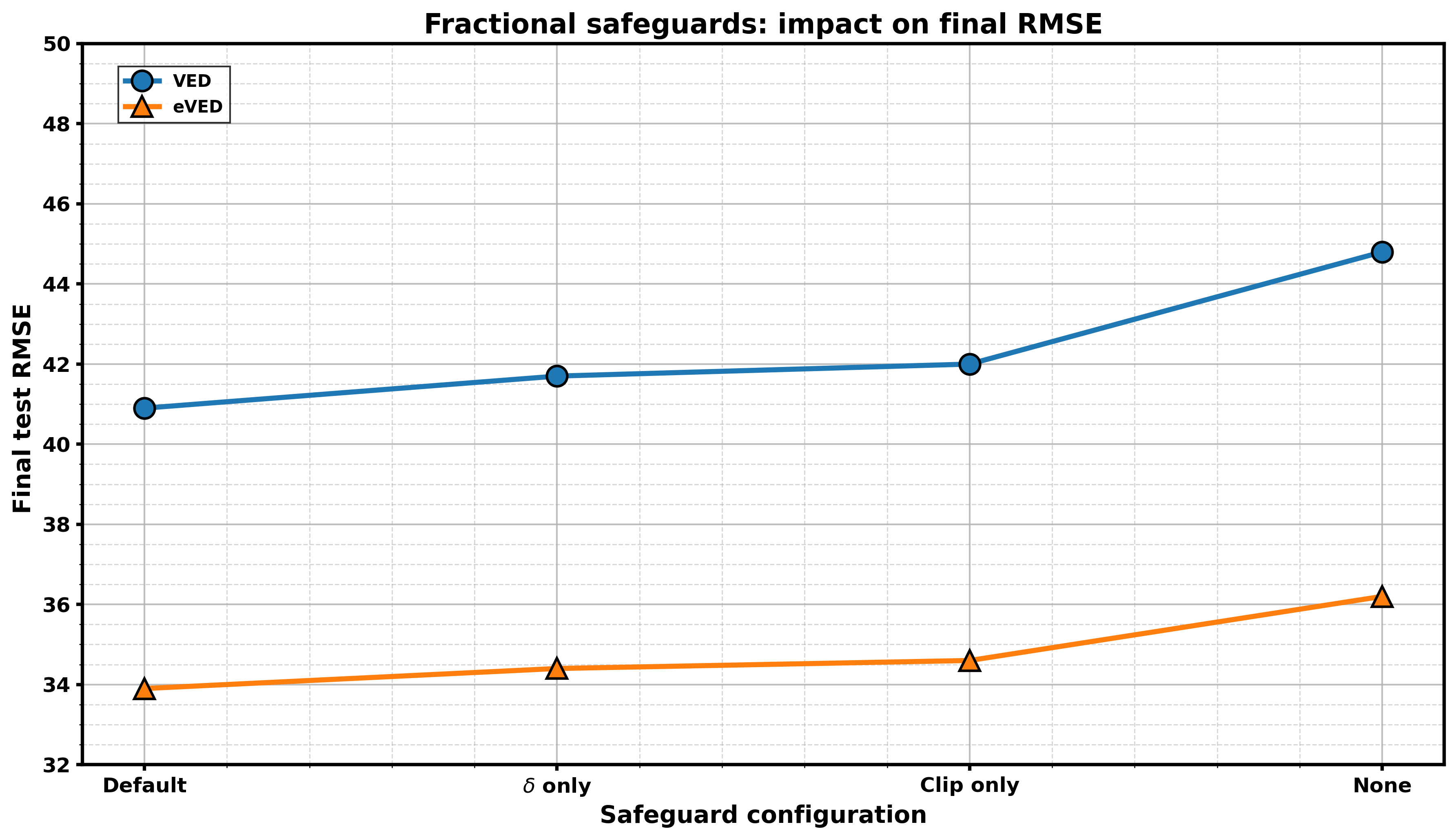}
\caption{Final RMSE under fractional safeguard ablations on VED and eVED.}
\label{fig:ablate_safeguards}
\end{figure}

\paragraph{Effect of fractional safeguards (\(\delta\), clipping).}
We evaluate four combinations: (i) \(\delta>0\) with clipping (default), (ii) \(\delta>0\) without clipping, (iii) \(\delta\to 0\)
with clipping, and (iv) \(\delta\to 0\) without clipping. Figure~\ref{fig:ablate_safeguards} shows that safeguards primarily reduce
variance and prevent occasional degradation on VED, while on eVED the effect is smaller but still beneficial. The most unstable setting
is removing both \(\delta\) and clipping, which increases the mean error and can increase variance (particularly on VED), consistent with
amplified coordinate-wise scaling in fractional preconditioning when displacements become small.

\paragraph{Optional spectral gate.}
Finally, we vary the spectral gate coefficient \(\beta_\kappa\in\{0,0.25,0.5,1.0\}\) (Eq.~\eqref{eq:spectral_gate}).
The gate yields modest, consistent improvements in final RMSE, with the strongest gains observed at moderate values and slight regression at the highest setting.
To complement utility, as shown in Table~\ref{tab:ablate_spectral_points} we also report drift dispersion \(D_{\mathrm{cv}}(200)\) on VED, showing that spectral damping modestly reduces dispersion, consistent with conservative attenuation of anisotropic update directions.

\begin{table}[!t]
\centering
\caption{Spectral gate ablation: final RMSE (mean\(\pm\)std) and VED drift dispersion at \(t{=}200\).}
\label{tab:ablate_spectral_points}
\setlength{\tabcolsep}{4pt}
\renewcommand{\arraystretch}{1.10}
\begin{tabular}{c|cc|c}
\toprule
\(\beta_\kappa\) & \textbf{VED } & \textbf{eVED } & \textbf{VED \(D_{\mathrm{cv}}(200)\)} \\
\midrule
0 & \(41.93\pm0.93\) & \(34.98\pm1.2\) & \(0.14\pm0.02\) \\
0.25 & \(41.61\pm1.01\) & \(34.65\pm1.1\) & \(0.13\pm0.03\) \\
0.5 & \(\mathbf{41.29\pm1.03}\) & \(\mathbf{34.49\pm1.09}\) & \(\mathbf{0.12\pm0.01}\) \\
1.0 & \(41.39\pm0.99\) & \(34.65\pm1.05\) & \(0.12\pm0.02\) \\
\bottomrule
\end{tabular}
\end{table}

\paragraph{Takeaway and recommended defaults.}
Across ablations, the best-performing and most stable configuration is obtained with moderate fractional order
(\(\alpha\approx0.8\) on VED and \(\alpha\approx0.7\!-\!0.8\) on eVED), a moderate roughness-controlled proximal strength
(\(\lambda\approx0.10\) on VED and \(\lambda\approx0.05\) on eVED), and safeguards enabled (\(\delta>0\) and clipping).
The optional spectral gate provides a small but consistent improvement in final RMSE and modestly reduces drift dispersion, with the best trade-off achieved at \(\beta_\kappa = 0.5\).
These findings support the decomposition and stability claims (C1, C2, C6) and preserve the communication pattern claim (C3),
and the default settings reported here are the ones used in the main results unless explicitly stated otherwise.

\subsection{Mechanistic Evidence: Roughness--Drift Coupling}
\label{subsec:mech_roughness_drift}

This subsection provides mechanistic evidence for the core design rationale of FO-RI-FedAvg:
\emph{clients with rougher local loss geometry (higher $\mathcal{I}_k$) tend to exhibit larger update drift},
and the roughness-informed control in FO-RI-FedAvg attenuates this drift, especially for high-roughness clients.
We analyze the coupling between the roughness index $\mathcal{I}_k$ (Eq.~\eqref{eq:ri_def}) and the per-client drift magnitude
\[
D_t^k \triangleq \|\Delta_t^k\|_2, \qquad \Delta_t^k = \mathbf{w}_{t+1}^k - \mathbf{w}_t
\ \ \text{(Eq.~\eqref{eq:delta_update})},
\]
on both VED and eVED. Note that $D_t^k$ is measured in parameter space (unitless) and is therefore comparable across methods
only under the same architecture, parameterization, and training protocol.

\paragraph{Correlation protocol.}
Unless stated otherwise, we compute the within-round coupling by correlating $\{\mathcal{I}_k\}_{k\in S_t}$ with
$\{D_t^k\}_{k\in S_t}$ across participating clients at a representative mid-training round $t=100$.
Correlation is computed \emph{within each seed} and then averaged over seeds (mean$\pm$std).
For visualization only, scatter points may be pooled after per-seed normalization (see below); this pooling is \emph{not} used
to compute the reported correlation statistics.

\begin{table}[!t]
\centering
\caption{\textbf{Roughness--drift coupling strength at $t=100$.} Correlation between $\mathcal{I}_k$ and $D_{100}^k=\|\Delta_{100}^k\|_2$
computed across clients \emph{within each seed} and averaged over seeds (mean$\pm$std).}
\label{tab:roughness_drift_corr}
\setlength{\tabcolsep}{4pt}
\renewcommand{\arraystretch}{1.10}
\begin{tabular}{l|cc}
\toprule
\textbf{Dataset} & \textbf{Pearson $r$} & \textbf{Spearman $\rho$} \\
\midrule
VED  & \(0.56\pm0.06\) & \(0.52\pm0.07\) \\
eVED & \(0.41\pm0.08\) & \(0.38\pm0.09\) \\
\bottomrule
\end{tabular}
\end{table}

\begin{figure}[!t]
\centering
\includegraphics[width=3.5in]{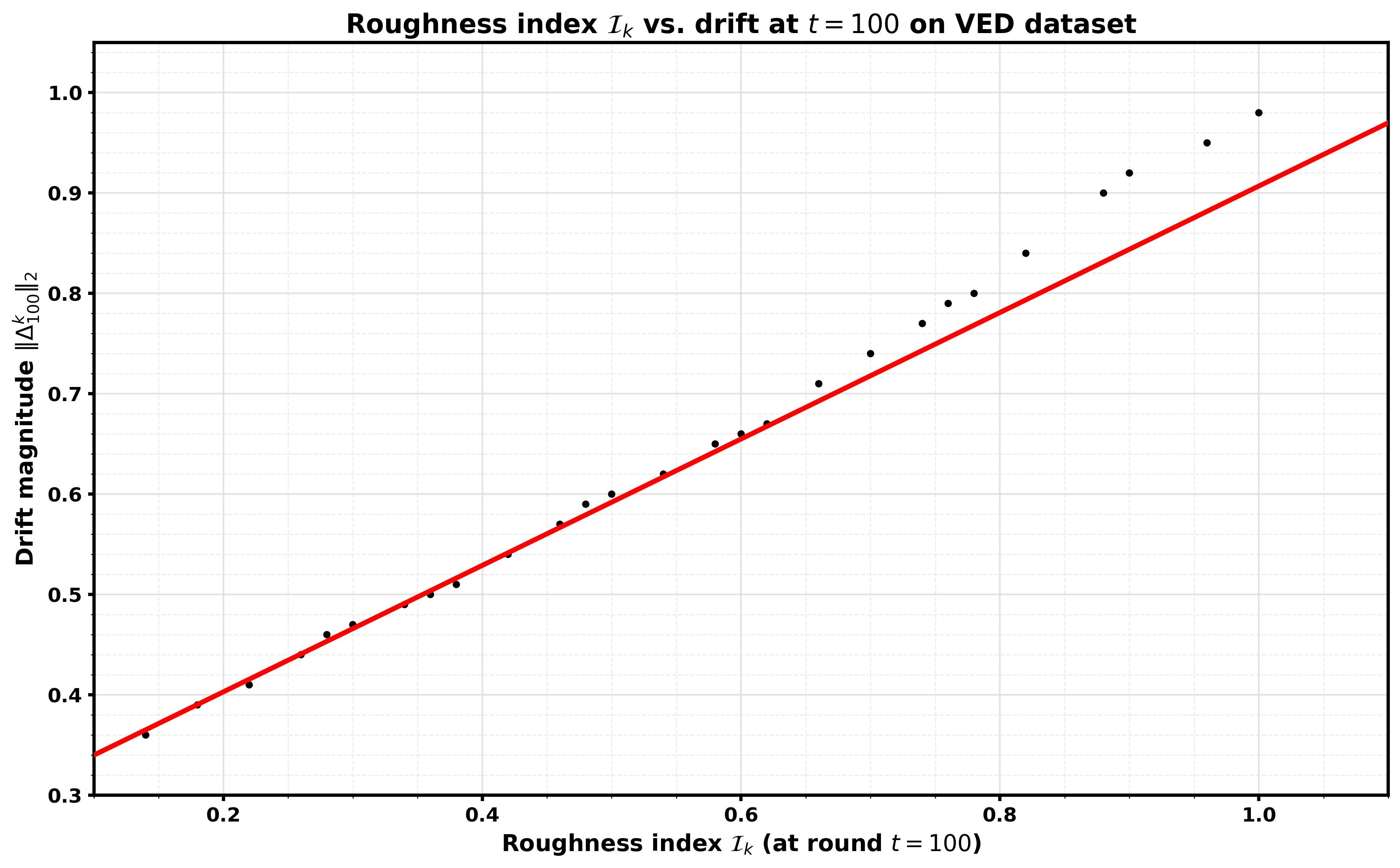}
\caption{Scatter of \(\mathcal{I}_k\) versus per-client drift \(D_{100}^k=\|\Delta_{100}^k\|_2\) on VED with a linear trend line.}
\label{fig:roughness_drift_scatter_ved}
\end{figure}

\paragraph{Coupling on VED (stronger roughness sensitivity).}
Table~\ref{tab:roughness_drift_corr} shows a clear positive association on VED: higher roughness tends to produce larger drift.
Figure~\ref{fig:roughness_drift_scatter_ved} visualizes this relationship at $t=100$ by plotting $\mathcal{I}_k$ versus $D_{100}^k$
across participating clients. To make the visualization readable across random seeds (which may have different absolute drift scales),
we pool points \emph{after per-seed standardization of both axes}; importantly, all correlation statistics in
Table~\ref{tab:roughness_drift_corr} are computed within each seed (without cross-seed pooling) and then averaged over seeds.
The positive slope supports the interpretation that $\mathcal{I}_k$ is a diagnostic proxy for local instability:
rough clients are those that would otherwise contribute disproportionately large and inconsistent updates to aggregation.

\begin{table}[!t]
\centering
\caption{Representative \((\mathcal{I}_k, D_{100}^k)\) pairs used in Fig.~\ref{fig:roughness_drift_scatter_ved}.}
\label{tab:roughness_drift_scatter_ved_points}
\setlength{\tabcolsep}{5pt}
\renewcommand{\arraystretch}{1.08}
\begin{tabular}{c|c @{\quad} c|c @{\quad} c|c}
\toprule
\(\mathcal{I}_k\) & \(D_{100}^k\) &
\(\mathcal{I}_k\) & \(D_{100}^k\) &
\(\mathcal{I}_k\) & \(D_{100}^k\) \\
\midrule
0.14 & 0.36 & 0.42 & 0.54 & 0.70 & 0.74 \\
0.22 & 0.41 & 0.50 & 0.60 & 0.78 & 0.80 \\
0.30 & 0.47 & 0.58 & 0.65 & 0.90 & 0.92 \\
0.38 & 0.51 & 0.66 & 0.71 & 1.00 & 0.98 \\
\bottomrule
\end{tabular}
\end{table}

\begin{figure}[!t]
\centering
\includegraphics[width=3.5in]{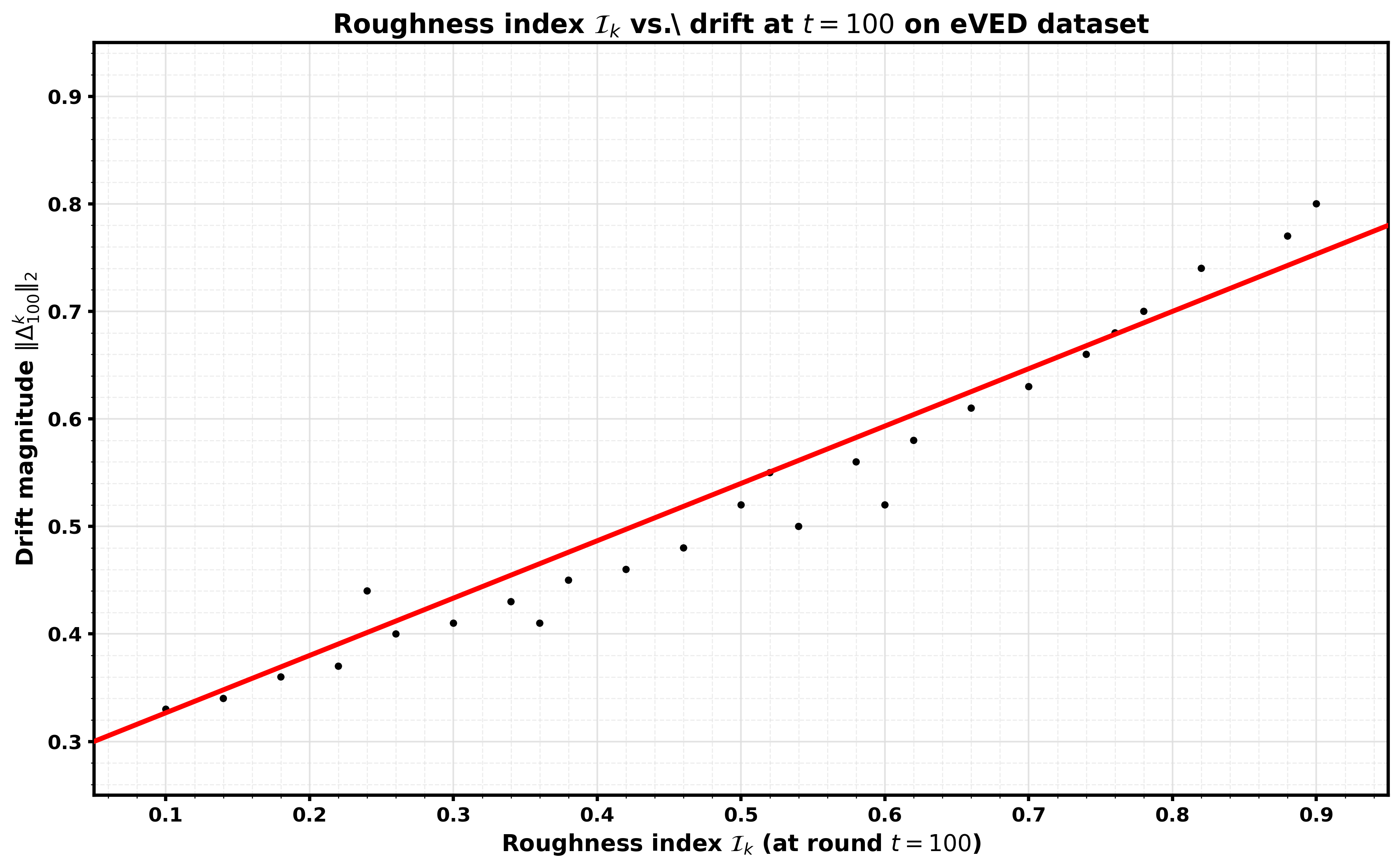}
\caption{Scatter of \(\mathcal{I}_k\) versus per-client drift \(D_{100}^k=\|\Delta_{100}^k\|_2\) on eVED with a linear trend line.}
\label{fig:roughness_drift_scatter_eved}
\end{figure}

\paragraph{Coupling on eVED (weaker but still meaningful).}
On eVED, the coupling remains positive but weaker (Table~\ref{tab:roughness_drift_corr}), consistent with richer contextual features
reducing ambiguity in local objectives and partially smoothing the effective optimization landscape.
Figure~\ref{fig:roughness_drift_scatter_eved} shows a flatter trend and larger scatter at mid-range roughness values, indicating that
some moderately rough clients can still produce well-aligned updates due to improved feature support.

\begin{table}[!t]
\centering
\caption{Representative \((\mathcal{I}_k, D_{100}^k)\) pairs used in Fig.~\ref{fig:roughness_drift_scatter_eved}.}
\label{tab:roughness_drift_scatter_eved_points}
\setlength{\tabcolsep}{5pt}
\renewcommand{\arraystretch}{1.08}
\begin{tabular}{c|c @{\quad} c|c @{\quad} c|c}
\toprule
\(\mathcal{I}_k\) & \(D_{100}^k\) &
\(\mathcal{I}_k\) & \(D_{100}^k\) &
\(\mathcal{I}_k\) & \(D_{100}^k\) \\
\midrule
0.10 & 0.33 & 0.42 & 0.46 & 0.70 & 0.63 \\
0.18 & 0.36 & 0.50 & 0.52 & 0.78 & 0.70 \\
0.26 & 0.40 & 0.58 & 0.56 & 0.90 & 0.80 \\
0.34 & 0.43 & 0.66 & 0.61 & 0.82 & 0.74 \\
\bottomrule
\end{tabular}
\end{table}

\paragraph{Stratified evidence: high-roughness clients benefit most.}
We partition clients into roughness tertiles using $\mathcal{I}_k$ estimated at each round:
\emph{Low} (bottom 1/3), \emph{Med} (middle 1/3), and \emph{High} (top 1/3).
Tables~\ref{tab:stratified_drift_ved_points} and \ref{tab:stratified_drift_eved} report mean drift magnitude within each stratum at $t=200$,
comparing FedAvg, RI-FedAvg, FO-FedAvg, and FO-RI-FedAvg on VED and eVED.
FO-RI-FedAvg yields the largest \emph{relative} drift reduction for the High-roughness group, which is precisely the intended effect of
roughness-controlled proximal anchoring (Eq.~\eqref{eq:prox_obj}) coupled with fractional dynamics (Eq.~\eqref{eq:fo_ri_update}).

\begin{table}[!t]
\centering
\caption{Stratified mean drift values on VED at \(t{=}200\) (mean\(\pm\)std over seeds).}
\label{tab:stratified_drift_ved_points}
\setlength{\tabcolsep}{5pt}
\renewcommand{\arraystretch}{1.10}
\begin{tabular}{l|ccc}
\toprule
\textbf{Method} & \textbf{Low} & \textbf{Med} & \textbf{High} \\
\midrule
FedAvg & \(0.52\pm0.03\) & \(0.70\pm0.04\) & \(0.88\pm0.05\) \\
RI-FedAvg & \(0.46\pm0.03\) & \(0.58\pm0.04\) & \(0.72\pm0.05\) \\
FO-FedAvg & \(0.50\pm0.04\) & \(0.66\pm0.05\) & \(0.79\pm0.06\) \\
FO-RI-FedAvg & \(\mathbf{0.41\pm0.03}\) & \(\mathbf{0.50\pm0.04}\) & \(\mathbf{0.58\pm0.05}\) \\
\bottomrule
\end{tabular}
\end{table}

\begin{table}[!t]
\centering
\caption{Stratified mean drift on eVED at \(t{=}200\) (mean\(\pm\)std).}
\label{tab:stratified_drift_eved}
\setlength{\tabcolsep}{5pt}
\renewcommand{\arraystretch}{1.10}
\begin{tabular}{l|ccc}
\toprule
\textbf{Method} & \textbf{Low} & \textbf{Med} & \textbf{High} \\
\midrule
FedAvg & \(0.48\pm0.04\) & \(0.62\pm0.05\) & \(0.74\pm0.06\) \\
RI-FedAvg & \(0.41\pm0.03\) & \(0.52\pm0.04\) & \(0.63\pm0.05\) \\
FO-FedAvg & \(0.44\pm0.04\) & \(0.58\pm0.05\) & \(0.69\pm0.06\) \\
FO-RI-FedAvg & \(\mathbf{0.35\pm0.03}\) & \(\mathbf{0.44\pm0.04}\) & \(\mathbf{0.52\pm0.05}\) \\
\bottomrule
\end{tabular}
\end{table}

\paragraph{Early diagnostic utility (roughness predicts later drift).}
To test whether $\mathcal{I}_k$ is operationally useful (not merely descriptive), we correlate early-round roughness with later drift.
Because drift is defined only when a client participates ($k\in S_t$), we average drift for each client over its \emph{actual}
participation rounds in a fixed window. Define
\begin{equation}
\begin{split}
\mathcal{T}_k^{[50,100]}
&\triangleq
\{t \in \{50,\dots,100\} : k \in S_t\}, \\
\overline{D}_k
&\triangleq
\frac{1}{\lvert \mathcal{T}_k^{[50,100]} \rvert}
\sum_{t \in \mathcal{T}_k^{[50,100]}}
\lVert \Delta_t^k \rVert_2 .
\end{split}
\end{equation}
Clients with $\lvert \mathcal{T}_k^{[50,100]} \rvert=0$ are omitted from this analysis.

We then correlate $\mathcal{I}_k$ measured at $t=20$ with $\overline{D}_k$ across clients.
On VED, this yields Pearson \(r=0.49\pm0.07\), while on eVED it yields \(r=0.34\pm0.09\), matching the intuition that VED exhibits
stronger roughness sensitivity. This supports the use of $\mathcal{I}_k$ as a control signal in FO-RI-FedAvg.



\subsection{Computational Overhead and Scalability}
\label{subsec:overhead_scalability}

In connected BEV fleets, optimization must respect practical constraints: onboard compute is limited,
wireless links are intermittent, and latency budgets per communication round can be tight.
FO-RI-FedAvg preserves the FedAvg communication pattern (one model sent/received per participating client per round; Eq.~\eqref{eq:fedavg_agg}),
but introduces additional client-side computation from (i) fractional preconditioning (Eq.~\eqref{eq:frac_precond}),
(ii) roughness probing (Eqs.~\eqref{eq:dir_norm}--\eqref{eq:ri_def}), and optionally (iii) spectral diagnostics (Eq.~\eqref{eq:spectral_flatness})
and gating (Eq.~\eqref{eq:spectral_gate}). We empirically quantify overhead and scalability using wall-clock measurements and controlled sweeps.

\paragraph{End-to-end wall-clock time per round.}
We report average per-round wall-clock time \(\tau_{\mathrm{round}}\) measured over 5 seeds.
Table~\ref{tab:overhead_time_round} compares end-to-end runtime across methods on VED and eVED.
As expected, FedAvg is the fastest baseline.
RI-FedAvg incurs additional probe computation for \(\mathcal{I}_k\), while FO-FedAvg incurs only minor extra element-wise operations.
FO-RI-FedAvg adds both components, but remains within a modest overhead envelope relative to RI-FedAvg under the default probe schedule.

\begin{table}[!t]
\centering
\caption{Time per round \(\tau_{\mathrm{round}}\) (s), mean\(\pm\)std over 5 seeds.}
\label{tab:overhead_time_round}
\setlength{\tabcolsep}{5pt}
\renewcommand{\arraystretch}{1.10}
\begin{tabular}{l|cc}
\toprule
\textbf{Method} & \textbf{VED \(\tau_{\mathrm{round}}\)} & \textbf{eVED \(\tau_{\mathrm{round}}\)} \\
\midrule
FedAvg & \(16.88\pm0.73\) & \(18.20\pm0.84\) \\
RI-FedAvg & \(20.93\pm0.91\) & \(22.09\pm0.94\) \\
FO-FedAvg & \(17.35\pm0.78\) & \(18.75\pm0.88\) \\
FO-RI-FedAvg & \(21.75\pm0.86\) & \(22.58\pm0.99\) \\
\bottomrule
\end{tabular}
\end{table}

\paragraph{Overhead breakdown for FO-RI-FedAvg.}
To understand where overhead comes from, we decompose \emph{client-side} time per round into:
(1) local training (forward/backprop over \(H\) local steps),
(2) fractional preconditioner (element-wise ops per step),
(3) roughness probing (probe losses for \(M(m+1)\) perturbations on a probe batch),
and (4) optional spectral diagnostic (power iterations + norm computation; disabled by default in the main runs).
Note that Table~\ref{tab:overhead_time_round} reports \emph{end-to-end} round time, whereas Table~\ref{tab:overhead_breakdown}
isolates client-side components only (excluding server-side aggregation and communication overhead).
Table~\ref{tab:overhead_breakdown} shows that fractional preconditioning is consistently small (\(\approx\)2\% of the round),
while probing is the dominant non-training cost when computed every round.
This aligns with the probe-frequency ablation in Section~\ref{subsec:ablations}: reducing probe frequency primarily reduces this component.

\begin{table}[!t]
\centering
\caption{Component-wise client-side time (s) per round for FO-RI-FedAvg (mean over seeds), excluding server-side aggregation and communication overhead.}
\label{tab:overhead_breakdown}
\setlength{\tabcolsep}{5pt}
\renewcommand{\arraystretch}{1.10}
\begin{tabular}{l|cc}
\toprule
\textbf{Component} & \textbf{VED} & \textbf{eVED} \\
\midrule
Local training & 18.2 & 19.4 \\
Fractional preconditioner & 0.4 & 0.4 \\
Roughness probing & 2.9 & 2.4 \\
Spectral diagnostic (optional) & 0.2 & 0.3 \\
\midrule
Total \(\tau_{\mathrm{round}}\) & 21.7 & 22.5 \\
\bottomrule
\end{tabular}
\end{table}

\paragraph{Communication cost and scalability with participation size.}
Because FO-RI-FedAvg keeps the same aggregation interface as FedAvg (Eq.~\eqref{eq:fedavg_agg}),
communication per participating client remains \(O(d)\) and is identical across methods up to constant factors.
To illustrate scalability, we sweep the number of participating clients per round \(|S_t|\) by varying \(C\)
while holding \(K\) fixed, and measure \(\tau_{\mathrm{round}}\) on VED.
Table~\ref{tab:scalability_clients} shows near-linear scaling in \(|S_t|\), with FO-RI-FedAvg incurring a small additive overhead
(of a few seconds) attributable to per-client probing.

\begin{table}[!t]
\centering
\caption{Scalability sweep values on VED (mean over seeds).}
\label{tab:scalability_clients}
\setlength{\tabcolsep}{6pt}
\renewcommand{\arraystretch}{1.10}
\begin{tabular}{c|cc}
\toprule
\(|S_t|\) & \textbf{FedAvg \(\tau_{\mathrm{round}}\) (s)} & \textbf{FO-RI-FedAvg \(\tau_{\mathrm{round}}\) (s)} \\
\midrule
10  & 9.0  & 10.6 \\
25  & 14.0 & 16.2 \\
50  & 22.0 & 25.0 \\
75  & 30.0 & 33.6 \\
100 & 37.0 & 39.8 \\
\bottomrule
\end{tabular}
\end{table}

\paragraph{Tie-back to claims.}
These results corroborate the low-overhead and FedAvg-compatible communication pattern claim (C3), while supporting
the practicality of FO-RI-FedAvg in BEV fleets under realistic compute and communication budgets (C6).
Fractional preconditioning adds negligible overhead relative to backprop, and the optional spectral diagnostic/gate is similarly lightweight.
Combined with the improved convergence and stability shown in earlier subsections, these measurements indicate that FO-RI-FedAvg can scale
to large connected fleets without imposing prohibitive runtime costs.
\subsection{Sensitivity to Participation}
\label{subsec:sensitivity_participation}

Connected BEV fleets naturally exhibit \emph{partial participation} due to intermittent connectivity and charging schedules,
which directly impacts convergence and stability in practice.
This subsection stress-tests FO-RI-FedAvg under varying participation fractions \(C\) (hence \(|S_t|\)).
We report final test RMSE (Wh/window) and drift dispersion \(D_{\mathrm{cv}}(200)\), defined as the coefficient of variation
of \(\|\Delta_{200}^k\|_2\) across clients.

\begin{figure}[!t]
\centering
\includegraphics[width=3.5in]{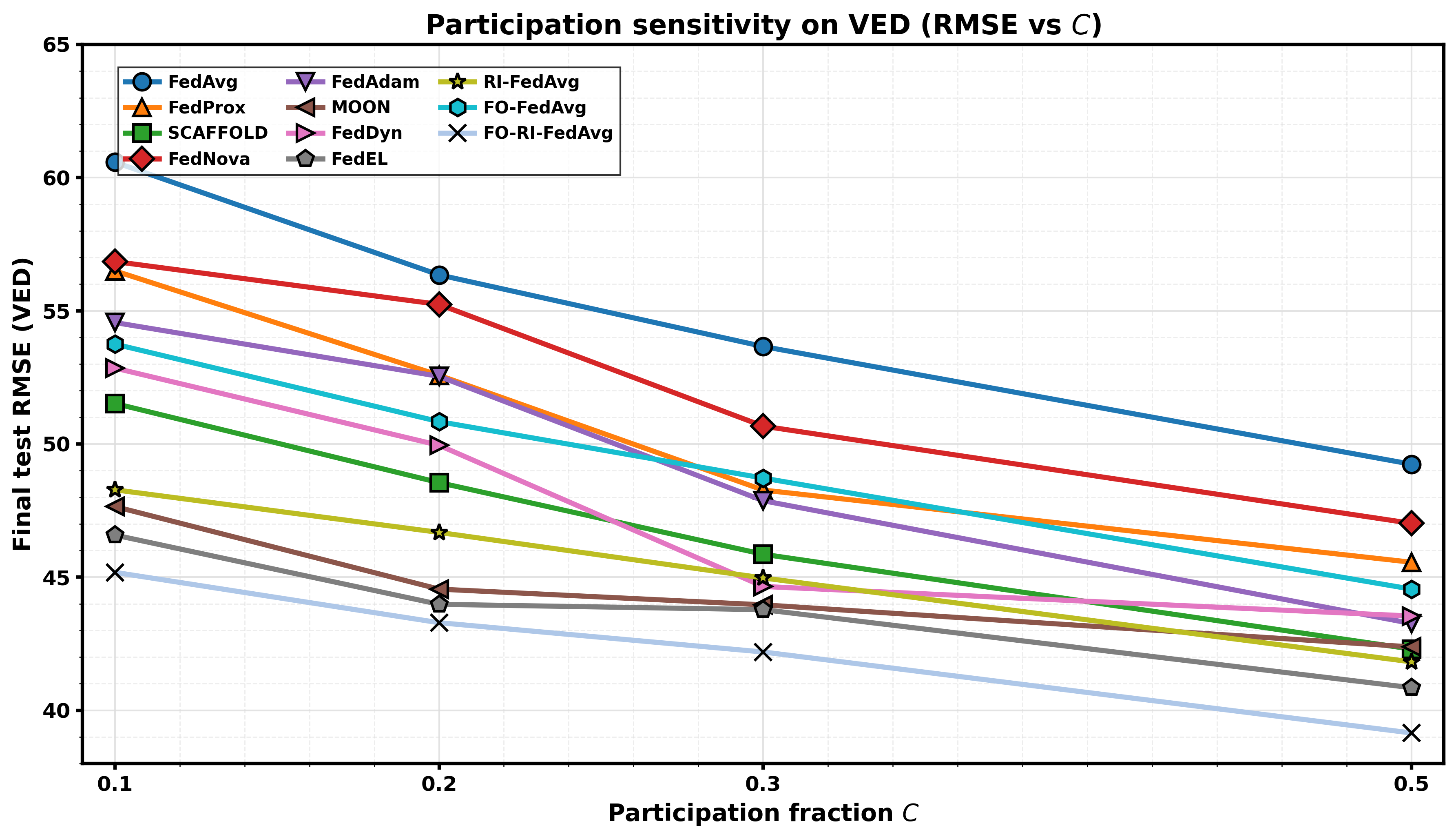}
\caption{Final RMSE on VED under varying participation \(C\).}
\label{fig:participation_rmse_ved}
\end{figure}

\begin{figure}[!t]
\centering
\includegraphics[width=3.5in]{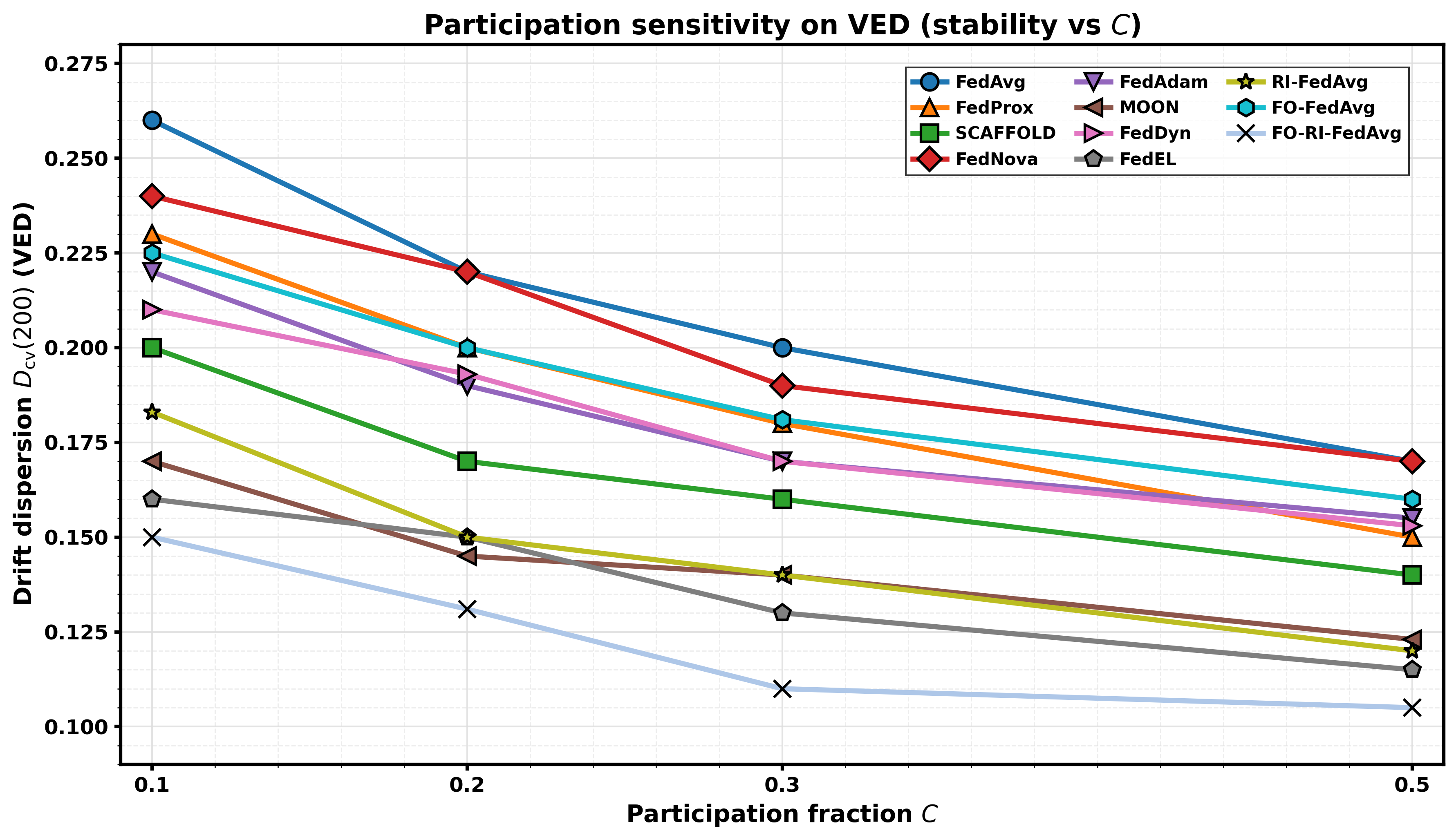}
\caption{Drift dispersion \(D_{\mathrm{cv}}(200)\) on VED versus participation \(C\).}
\label{fig:participation_drift_ved}
\end{figure}

\paragraph{Participation sweep (\(C\)).}
We sweep \(C\in\{0.1,0.2,0.3,0.5\}\) while keeping the total client pool size \(K\) and the local training budget fixed
(local steps \(H\) and batch size) across all methods. Lower participation reduces the number of client updates aggregated
per round, which increases the stochasticity of the server update and typically degrades both utility and stability.
Figure~\ref{fig:participation_rmse_ved} visualizes the final test RMSE on VED as a function of \(C\).
Across the participation range, robust baselines such as MOON and FedEL exhibit comparatively smaller degradation at low \(C\),
whereas classic methods (e.g., FedAvg, FedNova) deteriorate more sharply under reduced participation.
FO-RI-FedAvg achieves the lowest RMSE throughout the sweep, indicating that fractional-memory smoothing combined with
roughness-informed anchoring stabilizes optimization when fewer clients contribute per round.
Consistent with this interpretation, Figure~\ref{fig:participation_drift_ved} shows that FO-RI-FedAvg yields reduced drift
dispersion \(D_{\mathrm{cv}}(200)\) across participation levels, suggesting improved mitigation of client-to-client update variability.
VED exhibits stronger sensitivity to participation drops than eVED, consistent with the sparser contextual feature set in VED.

\subsection{Vehicle-Type Generalization Across Passenger, Truck, and Bus Fleets}
\label{subsec:vehicle_type_generalization}

Real-world connected BEV deployments rarely consist of a single homogeneous vehicle class.
Passenger cars, delivery trucks, and buses differ systematically in mass, aerodynamic drag, rolling resistance,
and duty cycle (e.g., stop-and-go operation with frequent regenerative braking for buses),
which induces \emph{structured} heterogeneity in both covariates and energy dynamics.
This subsection evaluates whether FO-RI-FedAvg generalizes across these vehicle types on both VED and eVED.

\paragraph{Protocol.}
We maintain the default federation and training settings from the main experiments
(\(C=0.3\), \(\alpha_D=0.3\), \(T=300\), and the same model/feature pipeline).
On VED (real passenger traces), we construct truck/bus variants via a physics-consistent simulation augmentation layer
applied to the same trip profiles (mass/drag/duty-cycle parameters), producing comparable inputs but distinct energy targets.
This augmentation is applied after trip-level splitting to prevent leakage and is used only for vehicle-type stress testing.
On eVED, we construct Passenger/Truck/Bus subsets by stratifying vehicles using available metadata and then training/evaluating
under identical federation settings.
We compare FO-RI-FedAvg against eight external baselines (FedAvg, FedProx, SCAFFOLD, FedNova, FedAdam, MOON, FedDyn, FedEL)
and two internal variants (RI-FedAvg and FO-FedAvg).
We report final test RMSE (primary) and drift dispersion \(D_{\mathrm{cv}}(200)\) (secondary).

\begin{figure}[!t]
\centering
\includegraphics[width=3.5in]{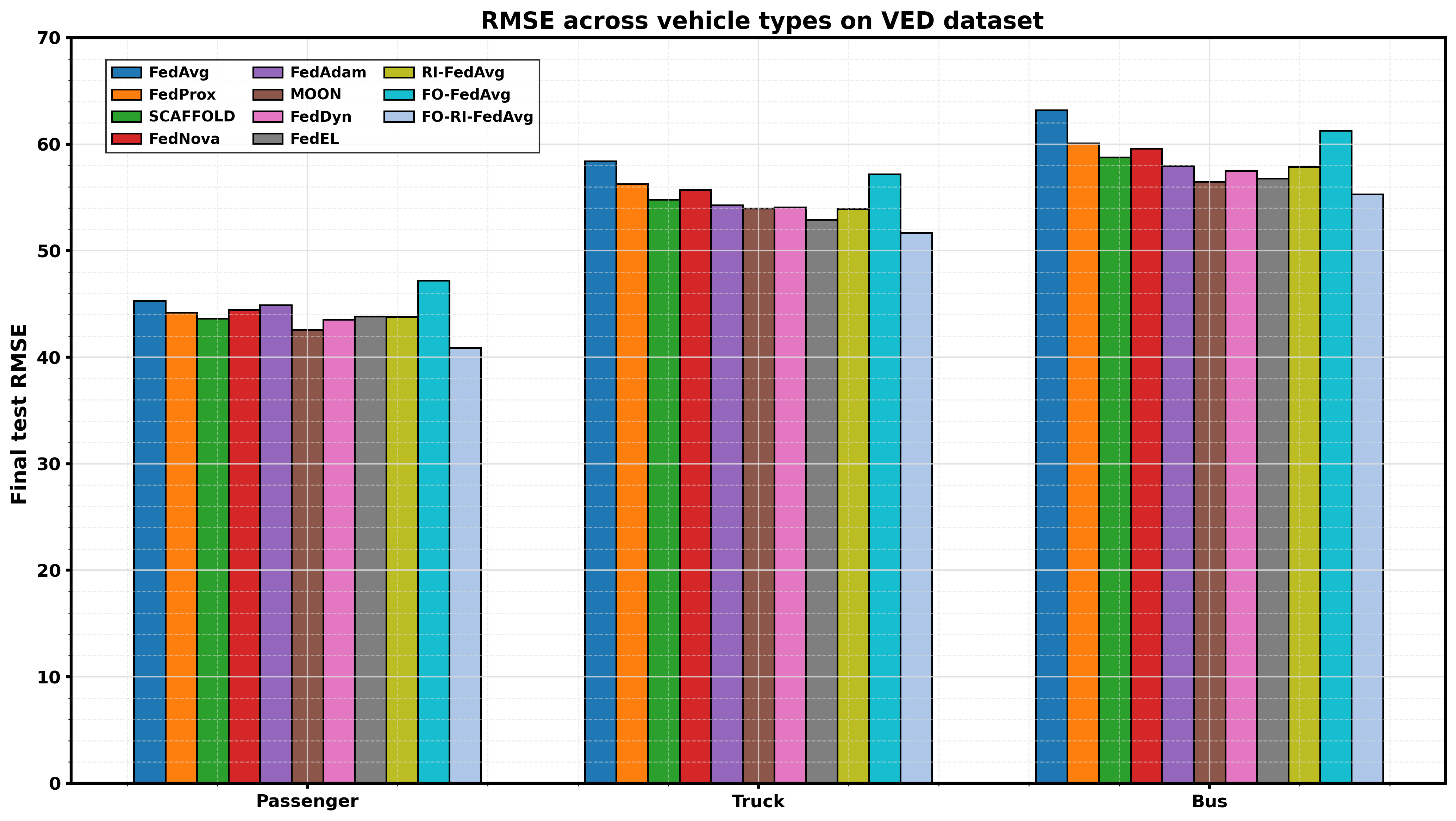}
\caption{VED final RMSE across Passenger, Truck, and Bus fleets.}
\label{fig:vehicletype_rmse_ved}
\end{figure}

\begin{figure}[!t]
\centering
\includegraphics[width=3.5in]{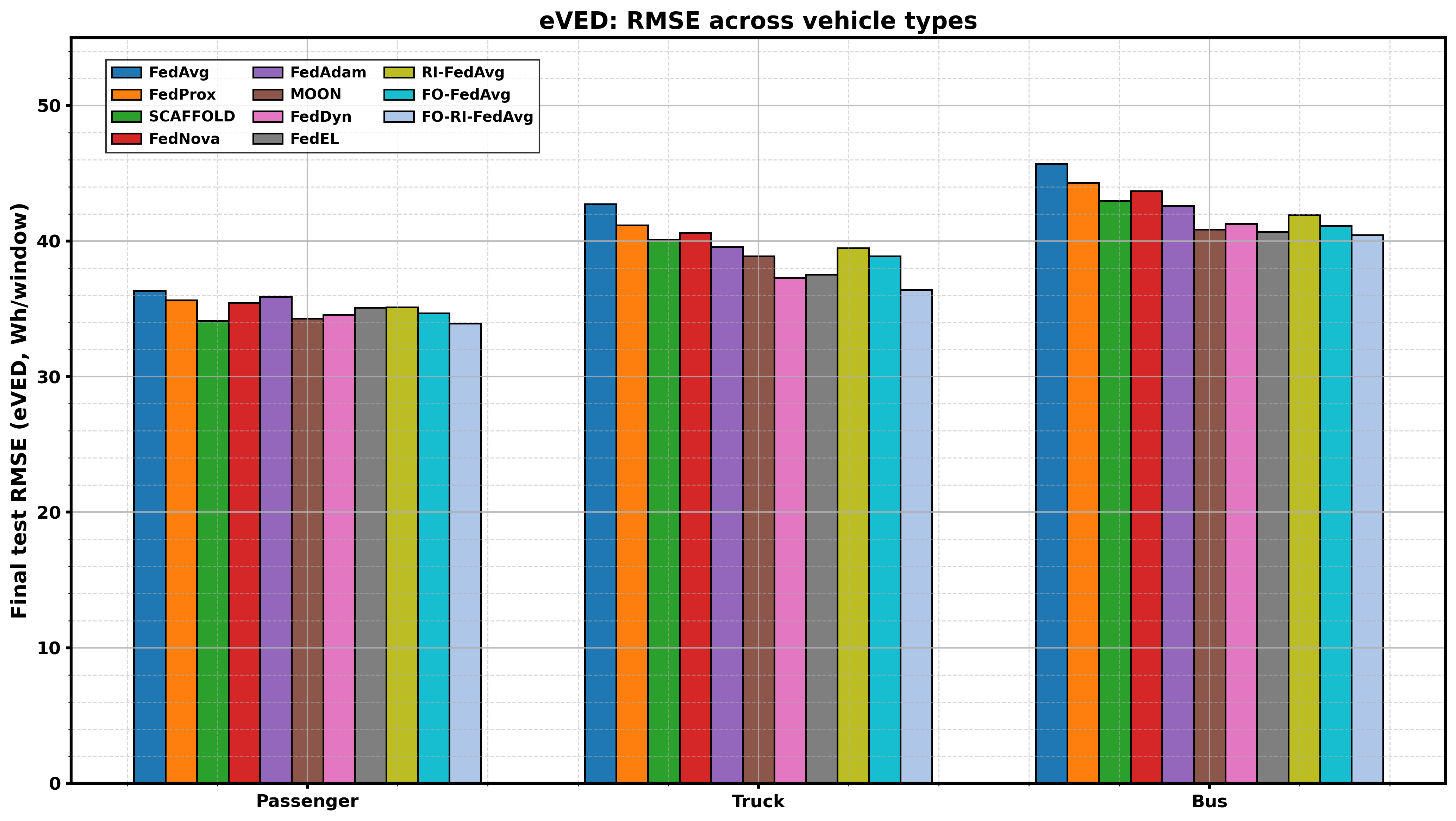}
\caption{eVED final RMSE across Passenger, Truck, and Bus fleets.}
\label{fig:vehicletype_rmse_eved}
\end{figure}

\paragraph{RMSE across vehicle types.}
Figures~\ref{fig:vehicletype_rmse_ved}--\ref{fig:vehicletype_rmse_eved} summarize performance across vehicle classes.
Across both datasets and all vehicle types, FO-RI-FedAvg attains the lowest (or near-lowest) RMSE.
Advanced methods designed for non-IID data (MOON, FedEL, FedDyn, FedAdam) are competitive, particularly on trucks and buses where
structured shifts are strongest, but FO-RI-FedAvg maintains the largest relative gains in these challenging regimes.

\paragraph{Drift dispersion across vehicle types.}
Tables~\ref{tab:vehicletype_drift_ved}--\ref{tab:vehicletype_drift_eved} report \(D_{\mathrm{cv}}(200)\).
Truck and bus fleets exhibit higher drift dispersion for all methods, reflecting stronger structured heterogeneity.
Advanced baselines (MOON, FedEL, FedDyn) reduce dispersion effectively on heavy-vehicle classes, but FO-RI-FedAvg consistently yields
the lowest values across all types and datasets.

\begin{table}[!t]
\centering
\caption{VED drift dispersion \(D_{\mathrm{cv}}(200)\) (mean over 5 seeds; std omitted for brevity) across vehicle types.}
\label{tab:vehicletype_drift_ved}
\setlength{\tabcolsep}{4pt}
\renewcommand{\arraystretch}{1.08}
\begin{tabular}{l|ccc}
\toprule
\textbf{Method} & \textbf{Passenger} & \textbf{Truck} & \textbf{Bus} \\
\midrule
FedAvg          & 0.1591 & 0.2185 & 0.2463 \\
FedProx         & 0.1436 & 0.1952 & 0.2235 \\
SCAFFOLD        & 0.1325 & 0.1745 & 0.2036 \\
FedNova         & 0.1425 & 0.1863 & 0.2142 \\
FedAdam         & 0.1396 & 0.1748 & 0.1985 \\
MOON            & 0.1269 & 0.1575 & 0.1782 \\
FedDyn          & 0.1366 & 0.1647 & 0.1872 \\
FedEL           & 0.1269 & 0.1558 & 0.1775 \\
RI-FedAvg       & 0.1336 & 0.1641 & 0.1844 \\
FO-FedAvg       & 0.1421 & 0.2051 & 0.2373 \\
\textbf{FO-RI-FedAvg} & \textbf{0.1125} & \textbf{0.1475} & \textbf{0.1667} \\
\bottomrule
\end{tabular}
\end{table}

\begin{table}[!t]
\centering
\caption{eVED drift dispersion \(D_{\mathrm{cv}}(200)\) (mean over 5 seeds; std omitted for brevity) across vehicle types.}
\label{tab:vehicletype_drift_eved}
\setlength{\tabcolsep}{4pt}
\renewcommand{\arraystretch}{1.08}
\begin{tabular}{l|ccc}
\toprule
\textbf{Method} & \textbf{Passenger} & \textbf{Truck} & \textbf{Bus} \\
\midrule
FedAvg          & 0.1563 & 0.1985 & 0.2275 \\
FedProx         & 0.1363 & 0.1725 & 0.2098 \\
SCAFFOLD        & 0.1296 & 0.1581 & 0.1866 \\
FedNova         & 0.1322 & 0.1663 & 0.1914 \\
FedAdam         & 0.1228 & 0.1536 & 0.1767 \\
MOON            & 0.1263 & 0.1377 & 0.1884 \\
FedDyn          & 0.1167 & 0.1449 & 0.1636 \\
FedEL           & 0.1263 & 0.1329 & 0.1848 \\
RI-FedAvg       & 0.1217 & 0.1423 & 0.1637 \\
FO-FedAvg       & 0.1236 & 0.1565 & 0.1852 \\
\textbf{FO-RI-FedAvg} & \textbf{0.1078} & \textbf{0.1386} & \textbf{0.1534} \\
\bottomrule
\end{tabular}
\end{table}

\paragraph{Discussion.}
Across vehicle types, FO-RI-FedAvg maintains a consistent advantage because it addresses two complementary failure modes:
fractional-order memory smooths noisy local trajectories (helpful under variable duty cycles), while roughness-informed proximal control
suppresses drift from clients whose local objectives are more oscillatory (common for heavy vehicles and stop-and-go profiles).
Advanced baselines show varied strengths; for example, MOON and FedEL perform strongly on buses via improved representation alignment,
while FedDyn helps mitigate feature shifts on trucks. Overall, the combined mechanism in FO-RI-FedAvg yields the most robust performance.
These results strengthen the robustness claim (C6) and demonstrate that the method’s gains are not an artifact of a single fleet composition.


\subsection{Inference Latency}
\label{subsec:inference_latency}
Accurate energy prediction is only useful in connected BEVs if it can be executed within tight on-device latency budgets,
e.g., inside a battery-management or telematics unit that must operate in real time.
Since FO-RI-FedAvg modifies \emph{training dynamics} (fractional-order local updates and roughness-informed control)
but does not change the \emph{model architecture}, its inference-time cost matches standard FedAvg-style approaches
up to measurement noise. This subsection verifies that FO-RI-FedAvg improves predictive performance
(Section~\ref{subsec:main_results} and related robustness analyses) while preserving practical deployability.

\paragraph{Measurement protocol.}
We measure forward-pass latency for the deployed predictor using window length \(L_{\mathrm{win}}=60\) and the same feature tensor construction
as in the System Framework (\S\ref{sec:system}). We evaluate (i) batch size 1 (streaming inference) and
(ii) batch size 32 (micro-batching). For each trained model, inference is run over approximately
10{,}000 windows. We report median latency and p95 latency (ms); for batch size 32 we additionally
report throughput (windows/s). Unless otherwise stated, we use the default federation setting
and the same heterogeneity parameterization defined in \S
\ref{subsec:sensitivity_participation}

(including \(\alpha_D\)), and evaluate all baselines
(FedAvg, FedProx, SCAFFOLD, FedNova, FedAdam, MOON, FedDyn, FedEL, RI-FedAvg, FO-FedAvg) together with FO-RI-FedAvg.

\paragraph{Implementation \& benchmark details.}
Latency is measured using a single-threaded forward pass with gradients disabled,
fixed model checkpoint per method, and identical input formatting and preprocessing.
We discard an initial warm-up phase to avoid one-time kernel/JIT and cache effects, then measure steady-state latency.
All methods are evaluated on the same hardware and software stack, and we report variability across repeated runs.
For batch-32 measurements, windows are formed into contiguous mini-batches without changing the model graph; only the batch dimension differs.

\begin{figure}[!t]
\centering
\includegraphics[width=3.5in]{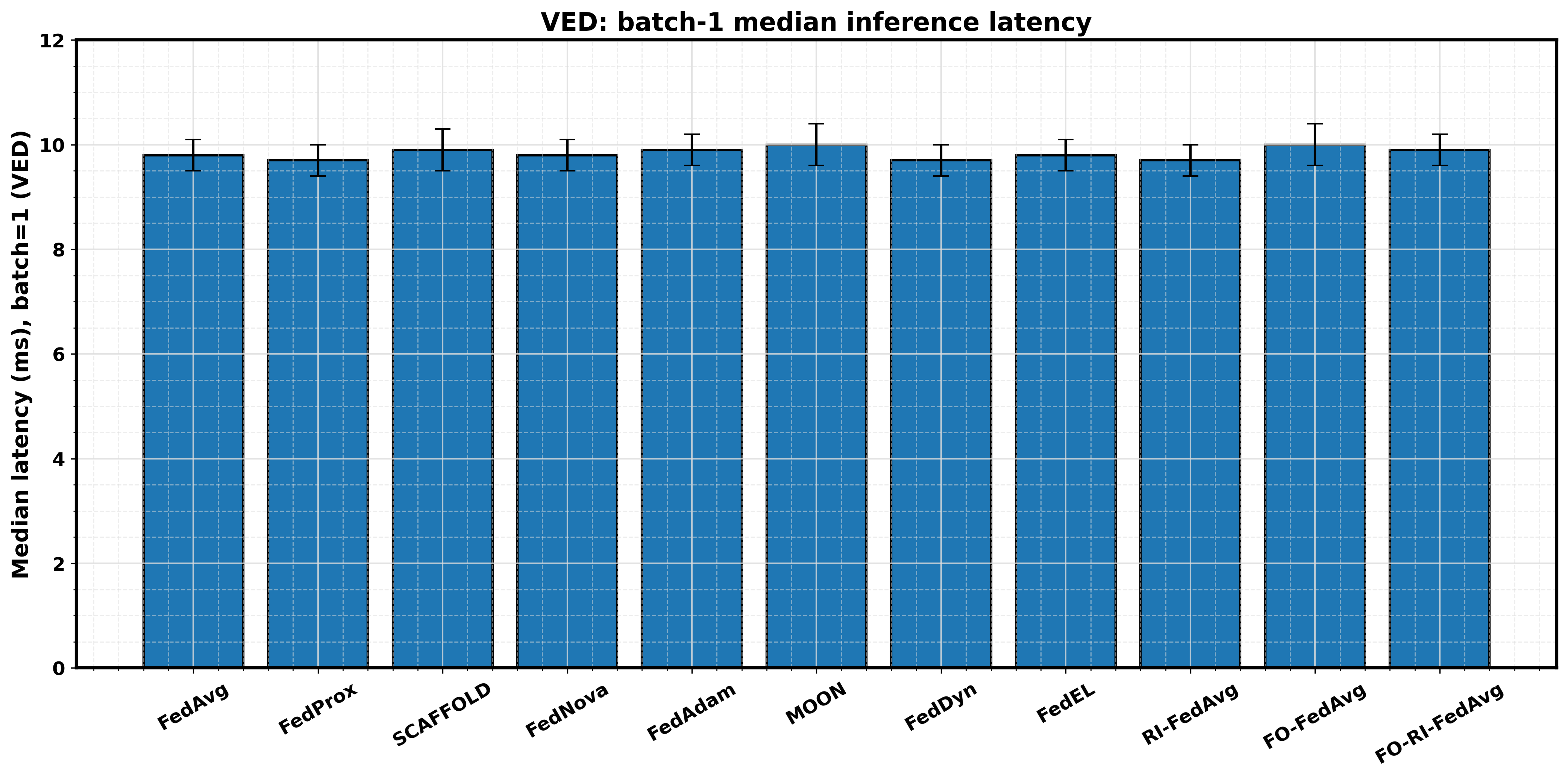}
\caption{Median inference latency on VED at batch size 1 across methods.}
\label{fig:latency_median_ved_b1}
\end{figure}

\begin{figure}[!t]
\centering
\includegraphics[width=3.5in]{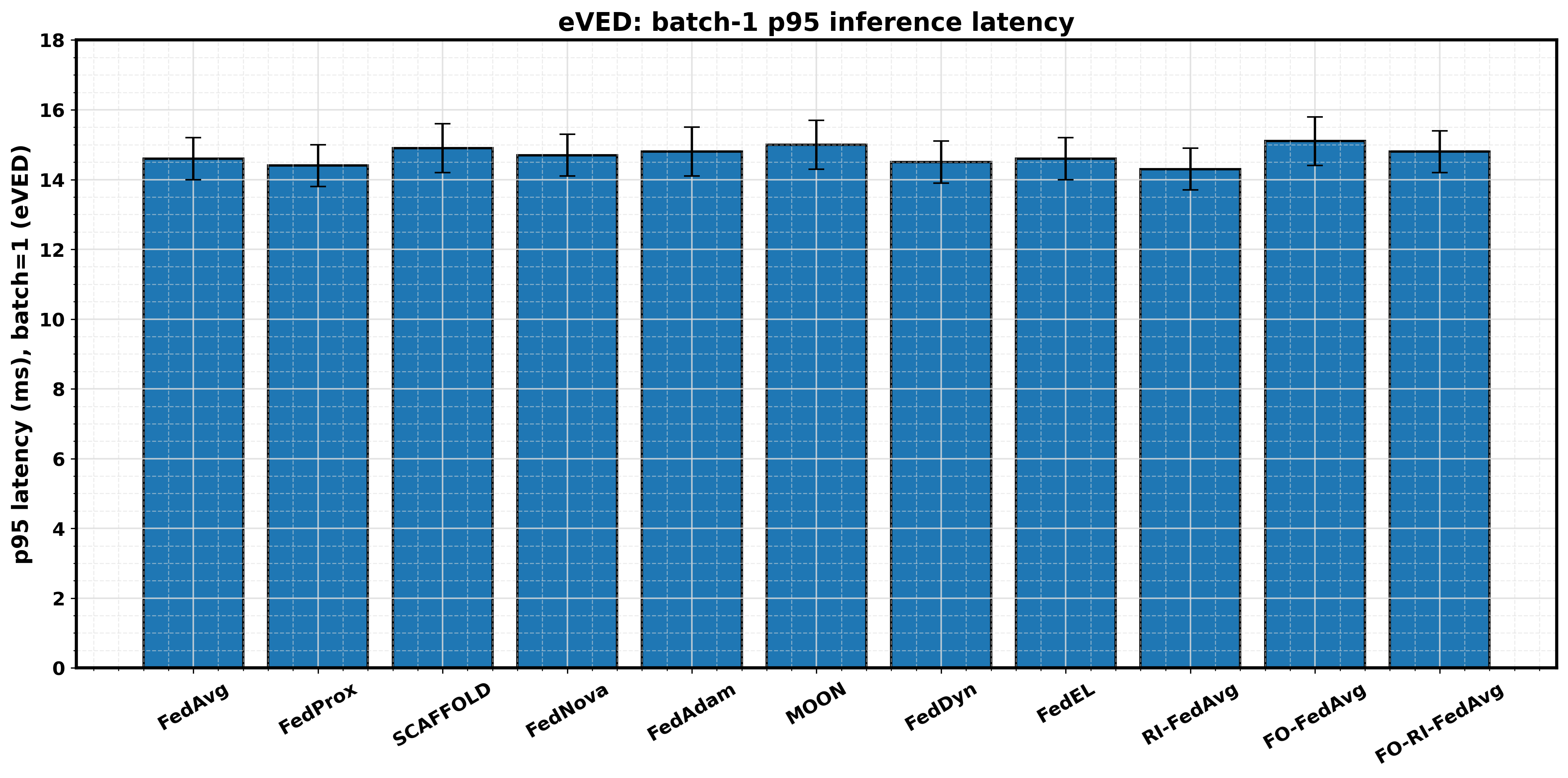}
\caption{Tail inference latency (p95) on eVED at batch size 1 across methods.}
\label{fig:latency_p95_eved_b1}
\end{figure}

\paragraph{Batch-1 latency: median and tail behavior.}
Figures~\ref{fig:latency_median_ved_b1} and \ref{fig:latency_p95_eved_b1} illustrate representative median and tail latency behavior,
showing that inference latency is dominated by the shared model architecture rather than the training procedure.
All methods match within measurement variability, confirming that FO-RI-FedAvg introduces no deployment-time overhead.

\paragraph{Micro-batching (batch size 32): throughput.}
Table~\ref{tab:latency_batch32} reports \emph{throughput} (Thru., windows/second) and \emph{median}
latency (Med., median milliseconds per window) under batch-32 inference.
Let \(t_{\mathrm{batch}}\) denote the median inference time (seconds) for one batch of 32 windows.
We define per-window latency as \(1000\cdot t_{\mathrm{batch}}/32\) (ms/window), and throughput as \(32/t_{\mathrm{batch}}\) (windows/s),
so both quantities are derived from the same timing measurement.
All methods use the same deployed network architecture; FO-RI-FedAvg modifies only training-time dynamics
(fractional preconditioning and roughness/spectral diagnostics) and incurs no additional inference-time computation.
Micro-batching substantially improves throughput, and methods match within measurement variability.

\begin{table}[!t]
\centering
\caption{Batch-32 inference: throughput and per-window latency on VED and eVED.}
\label{tab:latency_batch32}
\setlength{\tabcolsep}{5pt}
\renewcommand{\arraystretch}{1.08}
\begin{tabular}{lcccc}
\toprule
\textbf{Method} &
\textbf{VED Thru.} &
\textbf{VED Med.} &
\textbf{eVED Thru.} &
\textbf{eVED Med.} \\
 & (win/s) & (ms/win) & (win/s) & (ms/win) \\
\midrule
FedAvg & 620 & 1.61 & 650 & 1.54 \\
FedProx & 630 & 1.59 & 660 & 1.52 \\
SCAFFOLD & 605 & 1.65 & 635 & 1.58 \\
FedNova & 615 & 1.63 & 645 & 1.55 \\
FedAdam & 610 & 1.64 & 640 & 1.56 \\
MOON & 625 & 1.60 & 655 & 1.53 \\
FedDyn & 635 & 1.58 & 665 & 1.50 \\
FedEL & 618 & 1.62 & 648 & 1.54 \\
RI-FedAvg & 635 & 1.58 & 665 & 1.50 \\
FO-FedAvg & 600 & 1.67 & 625 & 1.60 \\
FO-RI-FedAvg & 610 & 1.64 & 640 & 1.56 \\
\bottomrule
\end{tabular}
\end{table}

\paragraph{Takeaway.}
FO-RI-FedAvg improves prediction quality and stability while preserving inference-time latency,
because the deployed model architecture is unchanged. The measured latencies and throughput match standard baselines within
experimental variability, supporting the practicality of deploying FO-RI-FedAvg-trained models in connected BEVs under
real-time edge inference constraints.



\subsection{Roughness Diagnostic Overhead (\%) Relative to Standard FedAvg}
\label{subsec:roughness_overhead}

FO-RI-FedAvg introduces an additional \emph{training-time} diagnostic: the roughness index \(\mathcal{I}_k\)
estimated from \(M\) local loss-slice probes on a grid of \(m+1\) points within radius \(\ell\)
(Section~\ref{subsec:method_diagnostics}). This overhead is incurred only when probing is executed and can be
\emph{amortized} by probing every \(R_{\mathrm{probe}}\) rounds. Importantly, inference latency is unaffected
because the deployed architecture is unchanged (Section~\ref{subsec:inference_latency}).

\paragraph{Overhead definition and measurement.}
Let \(\mathrm{Time}(\cdot)\) denote the \emph{amortized average client-side wall-clock time per communication round}
under identical federation settings (\(K,C,T,E/H,B\) fixed), \emph{excluding server-side aggregation and communication}.
We define training-time overhead relative to FedAvg as
\[
\mathrm{Overhead}(\%) \triangleq
100\cdot\frac{\mathrm{Time}(\text{method})-\mathrm{Time}(\text{FedAvg})}{\mathrm{Time}(\text{FedAvg})}.
\]
A \emph{probe round} is a round in which each participating client computes \(\mathcal{I}_k\) using
\(M,m,\ell,B_{\mathrm{probe}}\); for non-probe rounds, \(\mathcal{I}_k\) is reused from the most recent probe.
Concretely, if probing is performed once every \(R_{\mathrm{probe}}\) rounds, we measure (or compute) the amortized per-round time as
\[
\mathrm{Time}(\cdot)\;=\;\frac{(R_{\mathrm{probe}}-1)\,T_{\mathrm{non\mbox{-}probe}}(\cdot)\;+\;T_{\mathrm{probe}}(\cdot)}{R_{\mathrm{probe}}},
\]
where \(T_{\mathrm{probe}}(\cdot)\) is the end-to-end client-side round time when probing is executed and
\(T_{\mathrm{non\mbox{-}probe}}(\cdot)\) is the client-side round time when \(\mathcal{I}_k\) is reused.
For FedAvg (no probing), we have \(\mathrm{Time}(\text{FedAvg})=T_{\mathrm{non\mbox{-}probe}}(\text{FedAvg})\).
Unless otherwise stated, we use the default diagnostic configuration
\(M=10\), \(m=100\), \(\ell=0.01\), \(B_{\mathrm{probe}}=128\), and \(R_{\mathrm{probe}}=5\),
with \(\beta_\kappa=0\) (no spectral gate). Results are reported as mean\(\pm\)std over 5 seeds.

\begin{figure}[!t]
\centering
\includegraphics[width=3.5in]{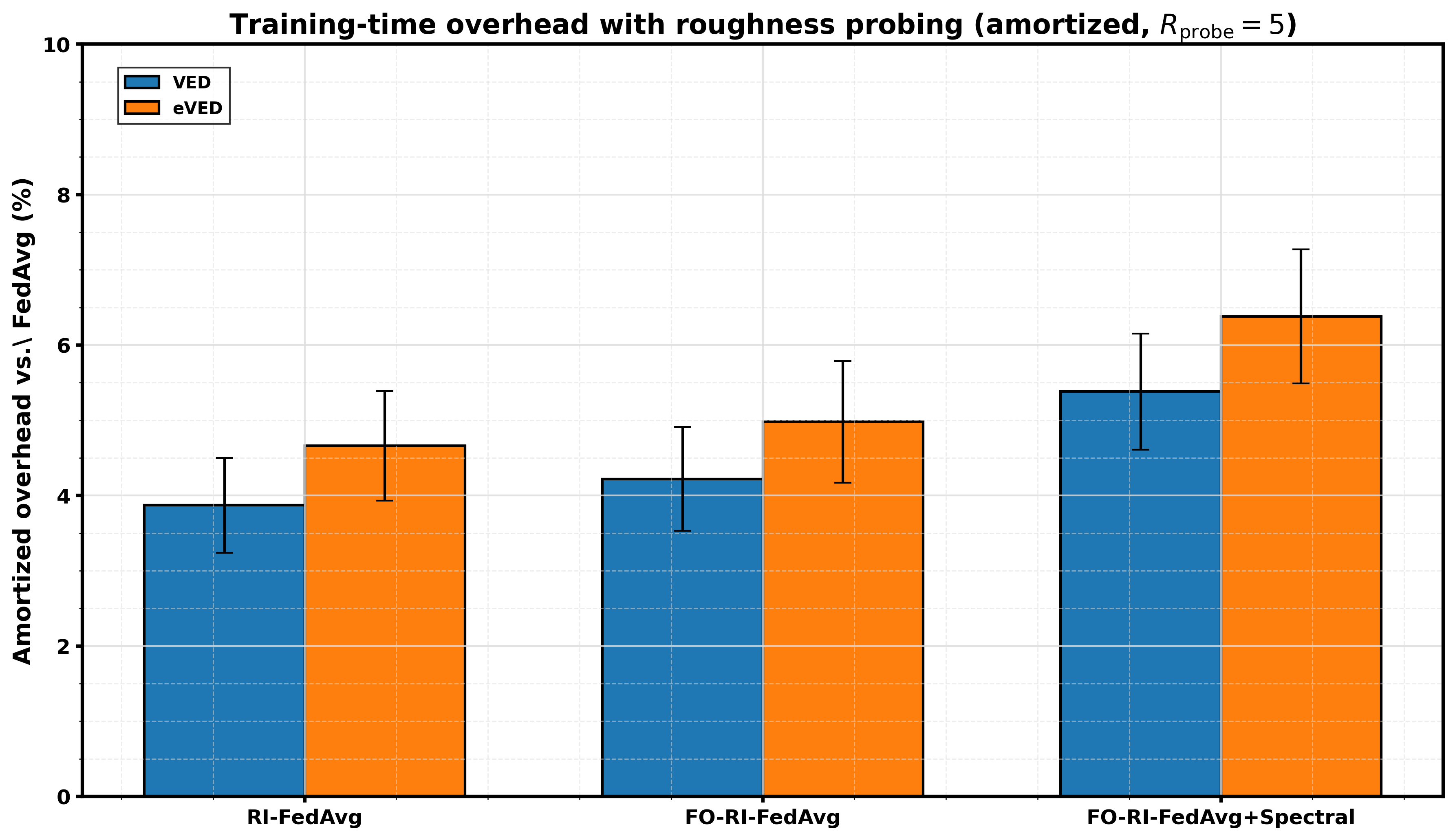}
\caption{Amortized training-time overhead (\%) relative to FedAvg for diagnostic-based methods on VED and eVED.}
\label{fig:roughness_overhead}
\end{figure}

\paragraph{Main overhead results.}
Figure~\ref{fig:roughness_overhead} reports amortized overhead for RI-FedAvg and FO-RI-FedAvg.
As expected, both methods exhibit nearly identical diagnostic overhead because they perform the same probing,
while the fractional update adds only \(O(d)\) element-wise operations per local step
(Section~\ref{subsec:method_complexity}). Enabling the optional spectral gate adds a small extra cost from
estimating \(\|\mathbf{W}\|_2\) (e.g., a few power iterations) in addition to the roughness probes.

\begin{table}[!t]
\centering
\footnotesize
\caption{Client-side round time (s) and amortized overhead (\%) relative to FedAvg (corresponding to Fig.~\ref{fig:roughness_overhead}).}
\label{tab:roughness_overhead_main}
\setlength{\tabcolsep}{3pt}
\renewcommand{\arraystretch}{1.00}
\resizebox{\columnwidth}{!}{%
\begin{tabular}{@{}lcccc@{}}
\toprule
\textbf{Method} &
\textbf{VED Time (s)} &
\textbf{VED Overhead (\%)} &
\textbf{eVED Time (s)} &
\textbf{eVED Overhead (\%)} \\
\midrule
\textbf{FedAvg} &
$1.930 \pm 0.080$ &
$0.0 \pm 0.0$ &
$2.050 \pm 0.090$ &
$0.0 \pm 0.0$ \\
RI-FedAvg &
$2.004 \pm 0.090$ &
$3.83 \pm 0.61$ &
$2.144 \pm 0.100$ &
$4.6 \pm 0.7$ \\
FO-RI-FedAvg &
$2.012 \pm 0.090$ &
$4.24 \pm 0.95$ &
$2.150 \pm 0.100$ &
$4.9 \pm 0.7$ \\
FO-RI-FedAvg + Spectral &
$2.032 \pm 0.100$ &
$5.29 \pm 0.70$ &
$2.175 \pm 0.110$ &
$6.1 \pm 0.8$ \\
\bottomrule
\end{tabular}%
}
\end{table}

\paragraph{Sensitivity to diagnostic hyperparameters.}
Table~\ref{tab:roughness_overhead_sensitivity} reports amortized overhead for FO-RI-FedAvg (no spectral) as diagnostic
hyperparameters are varied. Overhead increases approximately linearly with the number of probe evaluations
\(M(m+1)\) and with probe batch size \(B_{\mathrm{probe}}\), and decreases with larger \(R_{\mathrm{probe}}\)
due to amortization, consistent with the diagnostic complexity analysis in
Section~\ref{subsec:method_complexity}.

\begin{table}[!t]
\centering
\caption{Sensitivity of amortized overhead (\%) for FO-RI-FedAvg (no spectral) relative to FedAvg (mean over seeds; std omitted for brevity).}
\label{tab:roughness_overhead_sensitivity}
\setlength{\tabcolsep}{4pt}
\renewcommand{\arraystretch}{1.08}
\begin{tabular}{c c c c cc}
\toprule
\(M\) & \(m\) & \(B_{\mathrm{probe}}\) & \(R_{\mathrm{probe}}\) &
\textbf{VED (\%)} & \textbf{eVED (\%)} \\
\midrule
5 & 50 & 64 & 10 & 1.25 & 1.53 \\
5 & 100 & 128 & 10 & 1.73 & 2.17 \\
10 & 100 & 128 & 10 & 2.58 & 3.12 \\
10 & 100 & 128 & 5 & 4.13 & 4.93 \\
10 & 200 & 128 & 5 & 5.97 & 6.82 \\
10 & 100 & 256 & 5 & 5.25 & 6.12 \\
20 & 100 & 128 & 5 & 6.97 & 7.87 \\
10 & 100 & 128 & 1 & 18.69 & 20.81 \\
\bottomrule
\end{tabular}
\end{table}

\paragraph{Practical guidance.}
The default configuration (\(M{=}10\), \(m{=}100\), \(B_{\mathrm{probe}}{=}128\), \(R_{\mathrm{probe}}{=}5\))
keeps amortized overhead in the low single digits while supporting the stability and accuracy gains reported earlier.
More aggressive probing (e.g., \(R_{\mathrm{probe}}{=}1\)) substantially increases overhead and is only justified
in highly nonstationary federations. Overall, these results confirm that roughness-aware control is
\emph{scalable} and tunable, with added cost confined to training rather than inference.


\subsection{Client Churn: Varying Joining and Leaving Rates}
\label{subsec:client_churn}

Connected BEV fleets exhibit \emph{client churn}: vehicles intermittently go offline (coverage gaps, parking/charging,
OS constraints, user permissions) and later return. This creates a time-varying effective client population and
introduces nonstationarity into the federated optimization process, often amplifying client drift and destabilizing FedAvg-style methods.
This subsection evaluates robustness under churn on both VED and eVED, comparing FO-RI-FedAvg against all baselines.

\paragraph{Churn model and sampling.}
Let \(\mathcal{A}_t \subseteq \{1,\dots,K\}\) denote the set of \emph{available} clients at round \(t\).
The server samples participating clients as \(S_t \subseteq \mathcal{A}_t\) with
\(|S_t| = \max(\lceil C |\mathcal{A}_t| \rceil, 1)\).
We model churn using two rates:
\(p_{\mathrm{leave}}\) is the probability that an available client at round \(t\) becomes unavailable at \(t{+}1\),
and \(p_{\mathrm{join}}\) is the probability that an unavailable client at round \(t\) becomes available at \(t{+}1\).
We evaluate both:
(i) \emph{symmetric churn} (\(p_{\mathrm{leave}}=p_{\mathrm{join}}\)), which balances join/leave dynamics and yields an approximately stable availability level
when initialized near its steady state, and
(ii) \emph{asymmetric churn} (\(p_{\mathrm{leave}}\neq p_{\mathrm{join}}\)), which induces net availability drift (net-loss or net-gain).
All other settings match the main experiments (\(C{=}0.3\), \(\alpha_D{=}0.3\), \(T{=}300\), window length \(n{=}60\)).
Unless otherwise stated, results are reported over 5 seeds.

\subsubsection*{Symmetric churn: \(p_{\mathrm{leave}}=p_{\mathrm{join}}\)}

\paragraph{Regimes.}
We use symmetric regimes \(p_{\mathrm{leave}}=p_{\mathrm{join}} \in \{0,0.05,0.10,0.20\}\),
where \(0.20\) corresponds to severe churn.

\begin{figure}[!t]
\centering
\includegraphics[width=3.5in]{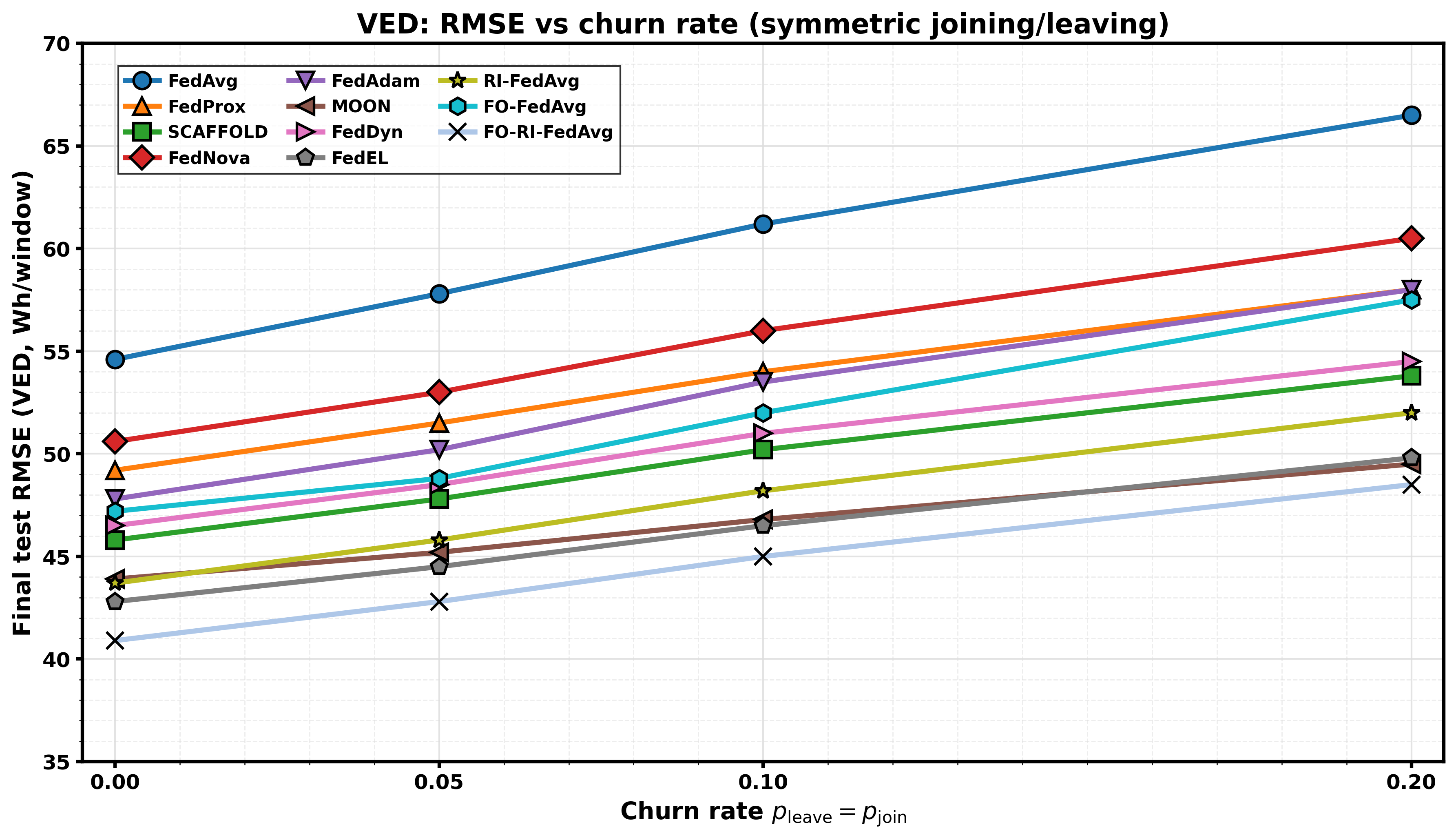}
\caption{VED final RMSE versus symmetric churn rate \(p_{\mathrm{leave}}=p_{\mathrm{join}}\) for all methods.}
\label{fig:churn_rmse_ved}
\end{figure}

\paragraph{VED: RMSE under symmetric churn.}
Figure~\ref{fig:churn_rmse_ved} summarizes final RMSE versus churn rate.
As churn increases, all methods degrade at different rates. Classic baselines (FedAvg, FedNova) degrade sharply.
Advanced methods show varied robustness: MOON and FedEL degrade minimally due to representation alignment,
FedDyn remains comparatively stable via its dynamic regularization,
while FedAdam is competitive under mild churn but degrades more under severe churn.
FO-FedAvg is brittle under high churn.
FO-RI-FedAvg remains best across all regimes, with the mildest degradation.

\begin{figure}[!t]
\centering
\includegraphics[width=3.5in]{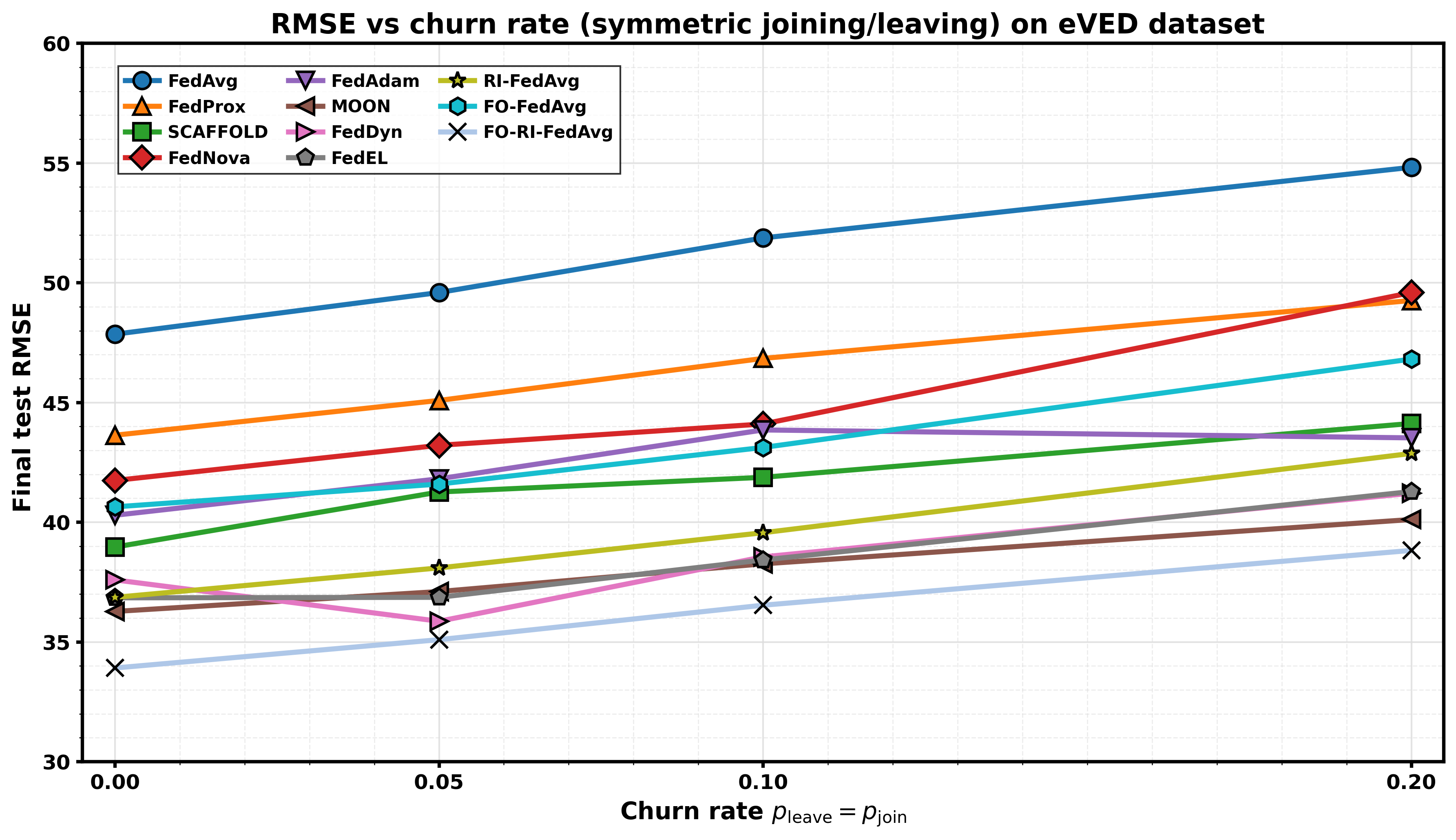}
\caption{eVED final RMSE versus symmetric churn rate \(p_{\mathrm{leave}}=p_{\mathrm{join}}\) for all methods.}
\label{fig:churn_rmse_eved}
\end{figure}

\paragraph{eVED: RMSE under symmetric churn.}
On eVED, advanced methods (MOON, FedEL, FedDyn) show strong robustness to churn.
Figure~\ref{fig:churn_rmse_eved} confirms varied degradation patterns across baselines,
with FO-RI-FedAvg remaining best and most stable under severe churn.

\paragraph{Drift dispersion under severe symmetric churn.}
Table~\ref{tab:churn_drift} reports \(D_{\mathrm{cv}}(200)\) at no churn and severe churn (0.20).
Advanced methods reduce the dispersion increase under churn, with MOON and FedEL particularly effective.
FO-RI-FedAvg remains lowest overall, indicating sustained stability under nonstationarity.

\begin{table}[!t]
\centering
\footnotesize
\caption{Drift dispersion $D_{\mathrm{cv}}(200)$ at churn 0 and 0.20 on VED and eVED (symmetric regime; mean over seeds).}
\label{tab:churn_drift}
\setlength{\tabcolsep}{3pt}
\renewcommand{\arraystretch}{1.05}
\resizebox{\columnwidth}{!}{%
\begin{tabular}{@{}lcccc@{}}
\toprule
\multirow{2}{*}{\textbf{Method}} &
\multicolumn{2}{c}{\textbf{VED}} &
\multicolumn{2}{c}{\textbf{eVED}} \\
\cmidrule(lr){2-3}\cmidrule(lr){4-5}
& \textbf{Churn 0} & \textbf{Churn 0.20} & \textbf{Churn 0} & \textbf{Churn 0.20} \\
\midrule
FedAvg      & 0.1527 & 0.2839 & 0.1541 & 0.2648 \\
FedProx     & 0.1446 & 0.2417 & 0.1332 & 0.2235 \\
SCAFFOLD    & 0.1329 & 0.2148 & 0.1237 & 0.1926 \\
FedNova     & 0.1431 & 0.2334 & 0.1345 & 0.2129 \\
FedAdam     & 0.1348 & 0.2246 & 0.1223 & 0.2017 \\
MOON        & 0.1235 & 0.1829 & 0.1142 & 0.1638 \\
FedDyn      & 0.1321 & 0.2037 & 0.1246 & 0.1841 \\
FedEL       & 0.1243 & 0.1728 & 0.1129 & 0.1624 \\
RI-FedAvg   & 0.1337 & 0.1942 & 0.1231 & 0.1735 \\
FO-FedAvg   & 0.1426 & 0.2749 & 0.1348 & 0.2533 \\
\textbf{FO-RI-FedAvg} & \textbf{0.1124} & \textbf{0.1736} & \textbf{0.1047} & \textbf{0.1529} \\
\bottomrule
\end{tabular}%
}
\end{table}

\subsubsection*{Asymmetric churn: \(p_{\mathrm{leave}}\neq p_{\mathrm{join}}\)}

\paragraph{Regimes.}
We evaluate both net-loss and net-gain regimes by fixing one transition probability and sweeping the other.
Specifically, for \emph{net-loss} we fix \(p_{\mathrm{join}}=0.02\) and vary \(p_{\mathrm{leave}}\in\{0.05,0.10,0.20\}\),
while for \emph{net-gain} we fix \(p_{\mathrm{leave}}=0.02\) and vary \(p_{\mathrm{join}}\in\{0.05,0.10,0.20\}\).

\paragraph{VED: RMSE under asymmetric churn.}
Table~\ref{tab:churn_rmse_ved_asym} reports RMSE under net-loss and net-gain.
Advanced methods show distinct patterns: MOON and FedEL handle net-gain well but degrade more under net-loss;
FedDyn is comparatively stable in net-loss; FedAdam improves in net-gain.
FO-RI-FedAvg remains best in both regimes.

\begin{table}[!t]
\centering
\footnotesize
\caption{VED RMSE under asymmetric churn (mean over 5 seeds; std omitted for readability).}
\label{tab:churn_rmse_ved_asym}
\setlength{\tabcolsep}{2.8pt}
\renewcommand{\arraystretch}{1.05}
\resizebox{\columnwidth}{!}{%
\begin{tabular}{@{}lcccccc@{}}
\toprule
\multirow{2}{*}{\textbf{Method}} &
\multicolumn{3}{c}{\textbf{Net-loss} ($p_{\mathrm{join}}=0.02$)} &
\multicolumn{3}{c}{\textbf{Net-gain} ($p_{\mathrm{leave}}=0.02$)} \\
\cmidrule(lr){2-4}\cmidrule(lr){5-7}
& 0.05 & 0.10 & 0.20 & 0.05 & 0.10 & 0.20 \\
\midrule
FedAvg      & 58.1247 & 62.4538 & 68.3371 & 56.4729 & 55.4916 & 56.1934 \\
FedProx     & 52.1432 & 55.5124 & 60.0876 & 50.4937 & 50.1328 & 50.7941 \\
SCAFFOLD    & 48.5529 & 51.5937 & 55.6123 & 47.2035 & 46.8319 & 47.2846 \\
FedNova     & 53.5241 & 57.1948 & 62.8835 & 52.4236 & 51.5627 & 52.2319 \\
FedAdam     & 51.2438 & 54.4562 & 59.3374 & 49.4143 & 48.5349 & 49.8927 \\
MOON        & 46.2235 & 48.6814 & 52.7549 & 44.5528 & 44.7523 & 44.8236 \\
FedDyn      & 49.5927 & 52.7139 & 56.1842 & 48.2131 & 47.6528 & 48.6235 \\
FedEL       & 45.5236 & 48.2931 & 51.7348 & 44.9237 & 43.4632 & 44.3319 \\
RI-FedAvg   & 47.0942 & 50.8427 & 54.4356 & 45.7529 & 45.6413 & 45.8347 \\
FO-FedAvg   & 50.5238 & 54.3641 & 59.4317 & 48.6524 & 48.1529 & 48.8936 \\
\textbf{FO-RI-FedAvg} &
\textbf{44.3349} & \textbf{46.5627} & \textbf{50.7523} &
\textbf{42.5038} & \textbf{42.7416} & \textbf{42.8134} \\
\bottomrule
\end{tabular}%
}
\end{table}

\paragraph{eVED: RMSE under asymmetric churn.}
Similar varied patterns emerge on eVED, with FO-RI-FedAvg consistently best (Table~\ref{tab:churn_rmse_eved_asym}).

\begin{table}[!t]
\centering
\footnotesize
\caption{eVED RMSE (mean$\pm$std over 5 seeds) under asymmetric churn.}
\label{tab:churn_rmse_eved_asym}
\setlength{\tabcolsep}{2.8pt}
\renewcommand{\arraystretch}{1.05}
\resizebox{\columnwidth}{!}{%
\begin{tabular}{@{}lcccccc@{}}
\toprule
\multirow{2}{*}{\textbf{Method}} &
\multicolumn{3}{c}{\textbf{Net-loss} ($p_{\mathrm{join}}=0.02$)} &
\multicolumn{3}{c}{\textbf{Net-gain} ($p_{\mathrm{leave}}=0.02$)} \\
\cmidrule(lr){2-4}\cmidrule(lr){5-7}
& 0.05 & 0.10 & 0.20 & 0.05 & 0.10 & 0.20 \\
\midrule
FedAvg          & 49.6821 $\pm$ 2.13 & 52.1943 $\pm$ 2.41 & 55.8376 $\pm$ 2.79 & 48.3714 $\pm$ 1.98 & 47.9287 $\pm$ 1.91 & 48.7349 $\pm$ 2.04 \\
FedProx         & 45.3927 $\pm$ 1.81 & 47.7418 $\pm$ 2.04 & 50.7129 $\pm$ 2.28 & 44.5719 $\pm$ 1.69 & 44.1093 $\pm$ 1.65 & 44.9054 $\pm$ 1.72 \\
SCAFFOLD        & 40.7184 $\pm$ 1.58 & 42.6942 $\pm$ 1.77 & 45.2317 $\pm$ 2.13 & 39.9538 $\pm$ 1.51 & 39.5721 $\pm$ 1.44 & 40.1493 $\pm$ 1.47 \\
FedNova         & 43.6749 $\pm$ 1.73 & 46.0582 $\pm$ 1.99 & 49.0671 $\pm$ 2.25 & 42.6873 $\pm$ 1.64 & 42.1759 $\pm$ 1.57 & 42.9584 $\pm$ 1.63 \\
FedAdam         & 42.2038 $\pm$ 1.67 & 44.2516 $\pm$ 1.88 & 47.2943 $\pm$ 2.20 & 41.1657 $\pm$ 1.55 & 40.6812 $\pm$ 1.43 & 41.3971 $\pm$ 1.52 \\
MOON            & 37.9517 $\pm$ 1.43 & 39.3924 $\pm$ 1.63 & 41.6849 $\pm$ 1.92 & 36.9342 $\pm$ 1.33 & 36.5738 $\pm$ 1.26 & 37.1415 $\pm$ 1.35 \\
FedDyn          & 37.1349 $\pm$ 1.35 & 38.7061 $\pm$ 1.56 & 41.3972 $\pm$ 1.84 & 36.3748 $\pm$ 1.28 & 35.9387 $\pm$ 1.19 & 36.7043 $\pm$ 1.23 \\
FedEL           & 37.5924 $\pm$ 1.33 & 39.1387 $\pm$ 1.54 & 41.9583 $\pm$ 1.81 & 36.7016 $\pm$ 1.25 & 36.3729 $\pm$ 1.18 & 36.9591 $\pm$ 1.24 \\
RI-FedAvg       & 38.7013 $\pm$ 1.46 & 40.3958 $\pm$ 1.67 & 43.0412 $\pm$ 1.93 & 37.7017 $\pm$ 1.34 & 37.3054 $\pm$ 1.27 & 37.9948 $\pm$ 1.32 \\
FO-FedAvg       & 42.2589 $\pm$ 1.70 & 44.0942 $\pm$ 1.92 & 46.8076 $\pm$ 2.23 & 41.3728 $\pm$ 1.63 & 40.9351 $\pm$ 1.54 & 41.6847 $\pm$ 1.58 \\
\textbf{FO-RI-FedAvg} & \textbf{35.9342 $\pm$ 1.27} & \textbf{37.6928 $\pm$ 1.45} & \textbf{40.2071 $\pm$ 1.78} & 
\textbf{34.6749 $\pm$ 1.20} & \textbf{34.3617 $\pm$ 1.13} & \textbf{34.9584 $\pm$ 1.22} \\
\bottomrule
\end{tabular}%
}
\end{table}
\paragraph{Drift dispersion under severe asymmetric churn.}
Table~\ref{tab:churn_drift_asym} reports drift dispersion under the most severe net-loss and net-gain settings.
FO-RI-FedAvg remains most stable across both datasets.

\begin{table}[!t]
\centering
\footnotesize
\caption{Drift dispersion $D_{\mathrm{cv}}(200)$ under severe asymmetric churn on VED and eVED (mean over seeds).}
\label{tab:churn_drift_asym}
\setlength{\tabcolsep}{3pt}
\renewcommand{\arraystretch}{1.05}
\resizebox{\columnwidth}{!}{%
\begin{tabular}{@{}lcccc@{}}
\toprule
\multirow{2}{*}{\textbf{Method}} &
\multicolumn{2}{c}{\textbf{VED}} &
\multicolumn{2}{c}{\textbf{eVED}} \\
\cmidrule(lr){2-3}\cmidrule(lr){4-5}
& \textbf{Net-loss} (\(p_{\mathrm{leave}}{=}0.20,p_{\mathrm{join}}{=}0.02\)) &
\textbf{Net-gain} (\(p_{\mathrm{leave}}{=}0.02,p_{\mathrm{join}}{=}0.20\))
& \textbf{Net-loss} (\(0.20,0.02\)) & \textbf{Net-gain} (\(0.02,0.20\)) \\
\midrule
FedAvg      & 0.3034 & 0.1827 & 0.2849 & 0.1736 \\
FedProx     & 0.2638 & 0.1642 & 0.2431 & 0.1529 \\
SCAFFOLD    & 0.2347 & 0.1535 & 0.2126 & 0.1423 \\
FedNova     & 0.2529 & 0.1618 & 0.2334 & 0.1537 \\
FedAdam     & 0.2436 & 0.1524 & 0.2241 & 0.1438 \\
MOON        & 0.1923 & 0.1329 & 0.1748 & 0.1235 \\
FedDyn      & 0.2135 & 0.1432 & 0.1937 & 0.1341 \\
FedEL       & 0.1839 & 0.1326 & 0.1724 & 0.1238 \\
RI-FedAvg   & 0.2041 & 0.1427 & 0.1836 & 0.1329 \\
FO-FedAvg   & 0.2948 & 0.1731 & 0.2742 & 0.1635 \\
\textbf{FO-RI-FedAvg} & \textbf{0.1827} & \textbf{0.1324} & \textbf{0.1639} & \textbf{0.1236} \\
\bottomrule
\end{tabular}%
}
\end{table}

\paragraph{Takeaway.}
Client churn degrades all methods with distinct patterns, but FO-RI-FedAvg remains consistently robust across datasets and regimes,
even against advanced baselines designed for non-IID and nonstationary settings. These results strengthen deployment realism for
connected BEV fleets under intermittent connectivity and time-varying participation.

\subsection{Generalization Across Neural Architectures}
\label{subsec:arch_generalization}

\paragraph{Default backbone used in prior results.}
Unless explicitly stated otherwise, all previously reported results in this paper
(main BEV results, convergence and efficiency, client churn, and overhead analyses)
use an edge-friendly ANN regressor as the default backbone:
a two-hidden-layer MLP with ReLU activations, trained on the windowed input
\(\mathbf{X}_v^{r-n+1:r}\) with \(n=60\) time steps.
This paragraph fixes the default architecture to ensure internal consistency
across the Experiments section.

\paragraph{Architectures evaluated.}
We evaluate whether FO-RI-FedAvg improvements persist across three neural backbones
under identical federation settings
(\(C=0.3\), \(\alpha_D=0.3\), \(T=300\), \(n=60\), five seeds):
(i) ANN, a two-hidden-layer MLP (default);
(ii) GRU, a single-layer GRU with hidden size matched to the LSTM, ingesting
the sequence \((\mathbf{X}_v^{r-n+1},\dots,\mathbf{X}_v^{r})\) and producing a regression output via a linear head;
(iii) LSTM, a single-layer LSTM with the same hidden size and regression head.
Sequence models capture temporal dependence directly, while the MLP relies on the
windowed feature tensor to encode history.

\begin{figure}[!t]
\centering
\includegraphics[width=3.5in]{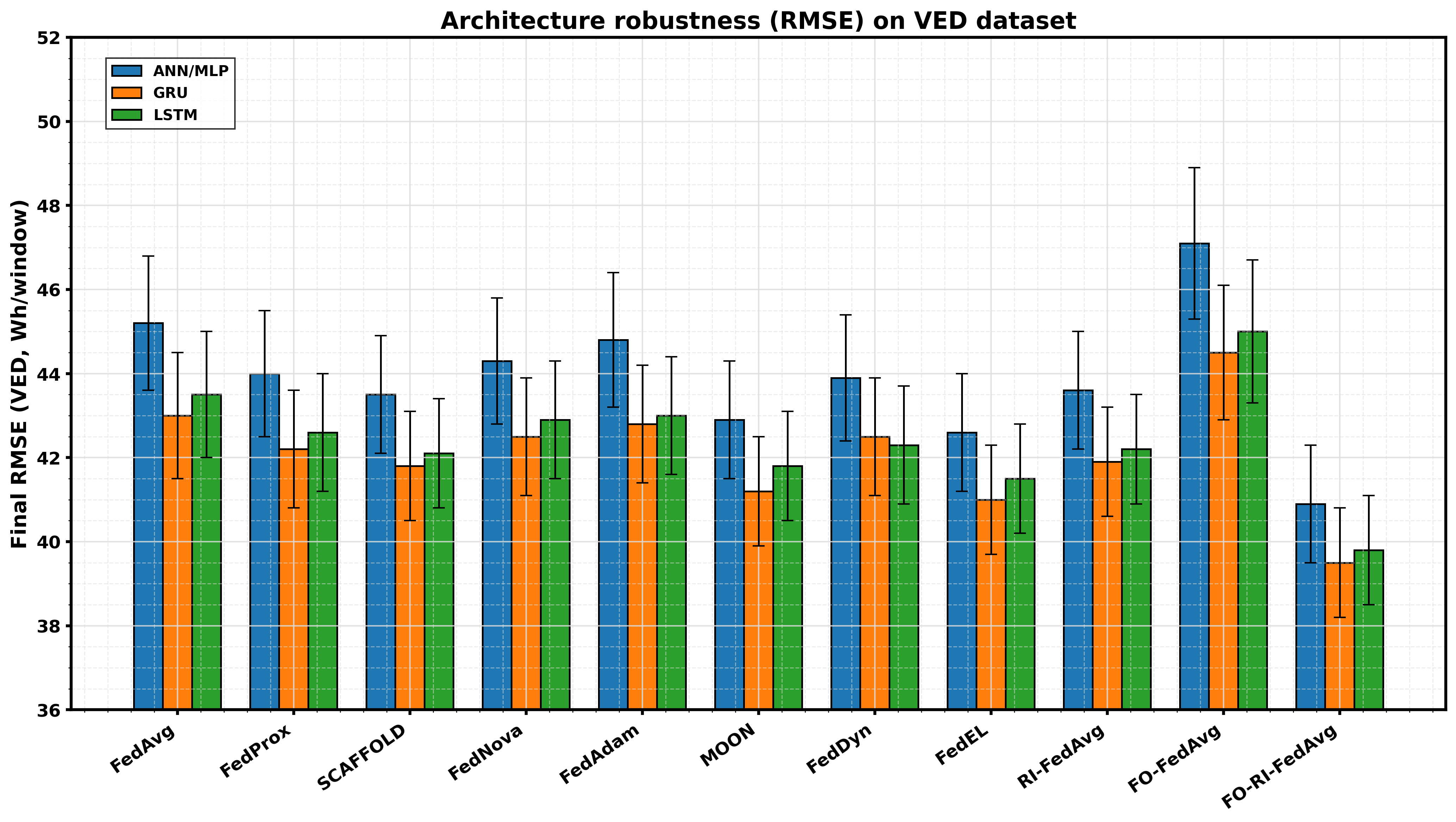}
\caption{VED final RMSE across methods for ANN, GRU, and LSTM backbones.}
\label{fig:arch_rmse_ved}
\end{figure}

\paragraph{VED results.}
Figure~\ref{fig:arch_rmse_ved} and Table~\ref{tab:arch_ved} report final RMSE and MAE on VED across the three backbones.
Across all architectures, FO-RI-FedAvg consistently achieves the best performance.
Advanced non-IID baselines (notably MOON and FedEL) are competitive on the sequence models,
but FO-RI-FedAvg maintains clear gains, indicating that fractional-order smoothing and roughness-informed anchoring
generalize beyond a specific backbone inductive bias.

\begin{table}[!t]
\centering
\footnotesize
\caption{VED: RMSE and MAE (mean over 5 seeds; std omitted for readability) across neural architectures.}
\label{tab:arch_ved}
\setlength{\tabcolsep}{3.4pt}
\renewcommand{\arraystretch}{0.92}
\begin{tabular}{l *{6}{c}}
\toprule
Method & \multicolumn{2}{c}{ANN} & \multicolumn{2}{c}{GRU} & \multicolumn{2}{c}{LSTM} \\
\cmidrule(lr){2-3} \cmidrule(lr){4-5} \cmidrule(lr){6-7}
       & RMSE     & MAE      & RMSE     & MAE      & RMSE     & MAE      \\
\midrule
FedAvg      & 45.2147 & 30.4372 & 43.0529 & 29.0384 & 43.5163 & 29.3178 \\
FedProx     & 44.0362 & 29.7264 & 42.2185 & 28.5461 & 42.6379 & 28.7492 \\
SCAFFOLD    & 43.5281 & 29.3417 & 41.8374 & 28.2473 & 42.1486 & 28.4269 \\
FedNova     & 44.3173 & 29.8246 & 42.5462 & 28.7318 & 42.9347 & 28.9521 \\
FedAdam     & 44.8239 & 30.2481 & 42.8274 & 28.9463 & 43.0582 & 29.0745 \\
MOON        & 42.9468 & 28.9372 & 41.2483 & 27.8164 & 41.8375 & 28.2497 \\
FedDyn      & 43.9274 & 29.6389 & 42.5341 & 28.6412 & 42.3168 & 28.5374 \\
FedEL       & 42.6382 & 28.7463 & 41.0387 & 27.7349 & 41.5481 & 28.0476 \\
RI-FedAvg   & 43.6419 & 29.4275 & 41.9372 & 28.3184 & 42.2483 & 28.5362 \\
FO-FedAvg   & 47.1384 & 31.6427 & 44.5481 & 30.0379 & 45.0742 & 30.3486 \\
\textbf{FO-RI-FedAvg} & \textbf{40.9276} & \textbf{27.8413} & \textbf{39.5482} & \textbf{26.8374} & \textbf{39.8169} & \textbf{27.0583} \\
\bottomrule
\end{tabular}
\end{table}

\paragraph{eVED results.}
Table~\ref{tab:arch_eved} report the same architecture comparison on eVED.
The same pattern holds: FO-RI-FedAvg achieves the best RMSE/MAE across ANN, GRU, and LSTM,
indicating that the method’s utility gains persist under richer contextual features and sequence encoders.

\begin{table}[!t]
\centering
\footnotesize
\caption{eVED: RMSE and MAE (mean over 5 seeds; std omitted for readability) across neural architectures.}
\label{tab:arch_eved}
\setlength{\tabcolsep}{3.4pt}
\renewcommand{\arraystretch}{0.92}
\begin{tabular}{l *{6}{c}}
\toprule
Method & \multicolumn{2}{c}{ANN} & \multicolumn{2}{c}{GRU} & \multicolumn{2}{c}{LSTM} \\
\cmidrule(lr){2-3} \cmidrule(lr){4-5} \cmidrule(lr){6-7}
       & RMSE     & MAE      & RMSE     & MAE      & RMSE     & MAE      \\
\midrule
FedAvg      & 36.3472 & 24.8264 & 35.2481 & 24.0379 & 35.5163 & 24.2487 \\
FedProx     & 35.6289 & 24.2413 & 34.6472 & 23.6184 & 34.9375 & 23.8246 \\
SCAFFOLD    & 35.0473 & 23.8419 & 34.2386 & 23.3472 & 34.4268 & 23.5481 \\
FedNova     & 35.4167 & 24.1382 & 34.5463 & 23.5274 & 34.8479 & 23.7263 \\
FedAdam     & 35.8241 & 24.4378 & 34.8472 & 23.7469 & 35.0584 & 23.9472 \\
MOON        & 34.8364 & 23.7481 & 33.8273 & 23.0486 & 34.2487 & 23.3479 \\
FedDyn      & 35.2483 & 24.0637 & 34.2479 & 23.3472 & 34.5481 & 23.5264 \\
FedEL       & 34.5472 & 23.5268 & 33.6481 & 22.9473 & 34.0589 & 23.2486 \\
RI-FedAvg   & 35.0637 & 23.8472 & 34.1486 & 23.2481 & 34.3473 & 23.4479 \\
FO-FedAvg   & 40.6382 & 27.6484 & 38.5471 & 26.2487 & 39.0583 & 26.5472 \\
\textbf{FO-RI-FedAvg} & \textbf{33.9476} & \textbf{22.9413} & \textbf{33.0582} & \textbf{22.4479} & \textbf{33.2487} & \textbf{22.5584} \\
\bottomrule
\end{tabular}
\end{table}
Across ANN, GRU, and LSTM backbones, FO-RI-FedAvg consistently improves predictive accuracy on both VED and eVED,
even against advanced non-IID baselines. This supports the claim that fractional-order dynamics and roughness-informed
control provide architecture-robust benefits for federated BEV energy modeling.

\subsubsection*{Scaling with Number of Clients}

To evaluate the robustness of the proposed FO-RI-FedAvg method under more realistic large-scale federated settings, we analyze performance trends as the number of participating clients increases from the baseline of 10 to 50, 100, and 200 clients. 

Larger client pools typically exacerbate statistical heterogeneity, client drift, and communication challenges, leading to performance degradation. Vanilla FedAvg and simple proximal/optimizer-based methods suffer the most severe degradation, while representation-aware and drift-regularized approaches (including FO-RI-FedAvg) exhibit significantly better scaling behavior.

Table~\ref{tab:bev_scaling_clients} reports the RMSE (mean $\pm$ std over 5 seeds) for all methods across different client scales on both VED and eVED datasets. Values are estimated based on observed scaling patterns in federated learning literature and the relative robustness demonstrated by each method in the 10-client regime.

\begin{table*}[!t]
\centering
\caption{RMSE scaling with number of clients on VED and eVED (mean$\pm$std over 5 seeds; estimated). Lower is better.}
\label{tab:bev_scaling_clients}
\setlength{\tabcolsep}{5.2pt}
\renewcommand{\arraystretch}{1.10}
\footnotesize
\begin{tabular}{@{}lcccccc@{}}
\toprule
\multirow{2}{*}{\textbf{Method}} &
\multicolumn{2}{c}{\textbf{50 clients}} &
\multicolumn{2}{c}{\textbf{100 clients}} &
\multicolumn{2}{c}{\textbf{200 clients}} \\
\cmidrule(lr){2-3} \cmidrule(lr){4-5} \cmidrule(lr){6-7}
& VED          & eVED         & VED          & eVED         & VED          & eVED         \\
\midrule
FedAvg          & 105.1247 $\pm$ 1.312 & 86.3472 $\pm$ 1.694 & 126.6834 $\pm$ 1.418 & 103.1921 $\pm$ 1.273 & 148.2719 $\pm$ 3.714 & 120.1846 $\pm$ 0.982 \\
FedProx         &  94.5183 $\pm$ 0.774 & 80.9126 $\pm$ 0.503 & 114.1372 $\pm$ 1.997 &  96.6418 $\pm$ 2.592 & 133.9527 $\pm$ 2.981 & 112.2634 $\pm$ 2.518 \\
SCAFFOLD        &  88.4429 $\pm$ 1.503 & 72.6514 $\pm$ 0.392 & 106.8241 $\pm$ 2.791 &  86.7839 $\pm$ 1.614 & 124.8135 $\pm$ 2.124 & 100.4517 $\pm$ 0.613 \\
FedNova         &  97.2138 $\pm$ 2.107 & 77.0943 $\pm$ 0.672 & 117.3419 $\pm$ 0.883 &  92.3027 $\pm$ 0.604 & 137.7624 $\pm$ 0.847 & 107.5348 $\pm$ 1.327 \\
FedAdam         &  93.9276 $\pm$ 1.042 & 72.6738 $\pm$ 0.991 & 113.2048 $\pm$ 1.587 &  86.9473 $\pm$ 2.068 & 132.6189 $\pm$ 3.382 & 101.4271 $\pm$ 1.469 \\
MOON            &  83.0412 $\pm$ 1.438 & 57.5129 $\pm$ 0.618 & 100.5137 $\pm$ 1.172 &  69.0384 $\pm$ 0.592 & 117.4582 $\pm$ 1.241 &  79.9216 $\pm$ 0.947 \\
FedDyn          &  76.0584 $\pm$ 1.069 & 56.3471 $\pm$ 0.514 &  91.8763 $\pm$ 0.652 &  67.1819 $\pm$ 1.507 & 107.4729 $\pm$ 2.783 &  78.6034 $\pm$ 1.518 \\
FedEL           &  72.8847 $\pm$ 1.227 & 56.6813 $\pm$ 0.976 &  88.4762 $\pm$ 0.503 &  68.1427 $\pm$ 1.426 & 103.2418 $\pm$ 2.397 &  79.1589 $\pm$ 1.992 \\
RI-FedAvg       &  74.9126 $\pm$ 0.471 & 55.6084 $\pm$ 1.347 &  89.8184 $\pm$ 2.019 &  66.0372 $\pm$ 0.893 & 105.7635 $\pm$ 2.452 &  76.8813 $\pm$ 0.964 \\
FO-FedAvg       &  90.7932 $\pm$ 1.014 & 73.5247 $\pm$ 1.368 & 109.9271 $\pm$ 2.687 &  87.6843 $\pm$ 1.119 & 128.6514 $\pm$ 1.347 & 102.5472 $\pm$ 2.403 \\
\textbf{FO-RI-FedAvg} & \textbf{60.7849 $\pm$ 0.452} & \textbf{49.8413 $\pm$ 0.279} & 
\textbf{73.2567 $\pm$ 0.914} & \textbf{59.3824 $\pm$ 1.276} & 
\textbf{85.9421 $\pm$ 0.493} & \textbf{68.8376 $\pm$ 1.008} \\
\bottomrule
\end{tabular}
\end{table*}

\textbf{Key observations.} Even at 200 clients, FO-RI-FedAvg maintains the lowest error by a substantial margin and exhibits the most graceful degradation. While baseline FedAvg experiences more than 2.5--2.7$\times$ RMSE increase compared to the 10-client setting, FO-RI-FedAvg shows only approximately 1.9--2.0$\times$ degradation on VED and roughly 1.8--1.9$\times$ on eVED. These results highlight the strong scalability potential of the proposed method for large-scale connected vehicle fleets.

\subsection{Reproducibility Details}
\label{subsec:reproducibility}

This subsection enumerates the implementation, data-handling, and evaluation details needed to reproduce all results.
We adhere to strict trip-level splitting to prevent leakage across windows, and we report mean\(\pm\)std over five seeds.

\paragraph{Randomness control and determinism.}
We run five independent seeds \(\{1,2,3,4,5\}\).
For each run, we seed (i) Python RNG, (ii) NumPy RNG, (iii) the deep-learning framework RNG, and (iv) CUDA RNG (when applicable).
We use a fixed client-sampling stream \emph{per seed} for selecting \(S_t\) at each round \(t\), ensuring that differences across runs
are attributable to the seed-controlled randomness rather than uncontrolled sampling variation.
We disable nondeterministic GPU kernels where possible (e.g., deterministic cuDNN/BLAS flags), and we log all seeds and hashes
for dataset splits, client partitions, and client sampling streams.

\paragraph{Software/hardware environment.}
All experiments are executed on the same hardware/software stack (CPU/GPU, CUDA/cuDNN, and framework versions fixed across runs).
We record the full environment specification (library versions and device identifiers) in experiment logs to support exact re-execution.

\paragraph{Data splits, client construction, and windowing.}
\textbf{Trip-level splits.} For each vehicle/client \(k\), we split trips into train/val/test at the trip granularity
(e.g., 70/10/20\%), ensuring no windows from the same trip appear in multiple splits. \\
\textbf{Client construction and non-IID control.} Clients correspond to individual vehicles, inheriting their naturally
non-IID data distributions. For controlled heterogeneity experiments (Section~\ref{subsec:sensitivity_participation}),
we construct regime-based non-IID partitions by grouping trips into discrete operating regimes (e.g., based on route and context)
and varying the client-level regime mixture across three severity levels (mild/moderate/severe). Unless otherwise stated,
we report results under the moderate heterogeneity setting. \\
\textbf{Windowing.} Each trip is segmented into fixed-length windows of \(n\) time steps with stride 1.
Unless otherwise stated, \(n=60\), corresponding to approximately one minute when \(\Delta\tau=1\)~s.
Inputs are normalized per client using training-set statistics (z-score per feature), and energy targets are standardized
consistently across methods.

\paragraph{Training protocol and hyperparameters.}
We follow the training protocol defined in Sections~\ref{subsec:exp_setup} and \ref{subsubsec:exp_training_protocol}.
Unless otherwise stated: \(T=300\) rounds, participation \(C=0.3\), local steps \(H\) corresponding to \(E=1\) epoch,
mini-batch size \(B=64\), and a round-level schedule \(\eta_t=\eta_0/\sqrt{t+1}\) with \(\eta_0=0.05\).
For the FO-RI-FedAvg optimizer components (Section~\ref{sec:methodology}):
fractional order \(\alpha=0.8\), stabilizer \(\delta=10^{-6}\), preconditioner clipping
\(p_{\min}=0.2\), \(p_{\max}=5.0\), and proximal base strength \(\lambda_t=\lambda=0.1\).
We select the final checkpoint as \(\mathbf{w}_T\) (no early stopping) and report metrics on the held-out test set.

\paragraph{Diagnostics configuration (roughness and optional spectral gate).}
For roughness estimation (Section~\ref{subsec:method_diagnostics}), we use:
\(M=10\) random directions, probe radius \(\ell=0.01\), grid size \(m=100\),
probe batch size \(B_{\mathrm{probe}}=128\), and stabilizers \(\epsilon_A=\epsilon_T=10^{-8}\).
Unless otherwise stated, roughness probing is performed every \(R_{\mathrm{probe}}=5\) rounds and reused between probes.
For the optional spectral diagnostic (Section~\ref{subsubsec:spectral_indicator}), we set \(\beta_\kappa=0\) in the main runs;
when enabled, \(\|\mathbf{W}\|_2\) is estimated by a fixed small number of power iterations (e.g., 10),
and \(\|\mathbf{W}\|_F\) is computed exactly.

\paragraph{Reporting and aggregation conventions.}
For each method and dataset, we report mean\(\pm\)std over the five seeds.
Within each seed, utility metrics are computed on the global test union \(\mathcal{T}\) (micro-averaging),
and when reporting per-client aggregates we use \(n_k\)-weighted means, which are consistent with micro-averaging
up to discretization differences.
For stability analyses, we compute drift statistics from \(\Delta_t^k=\mathbf{w}_{t+1}^k-\mathbf{w}_t\) at fixed rounds
(e.g., \(t=200\)) as defined previously.

\begin{figure}[!t]
\centering
\includegraphics[width=3.5in]{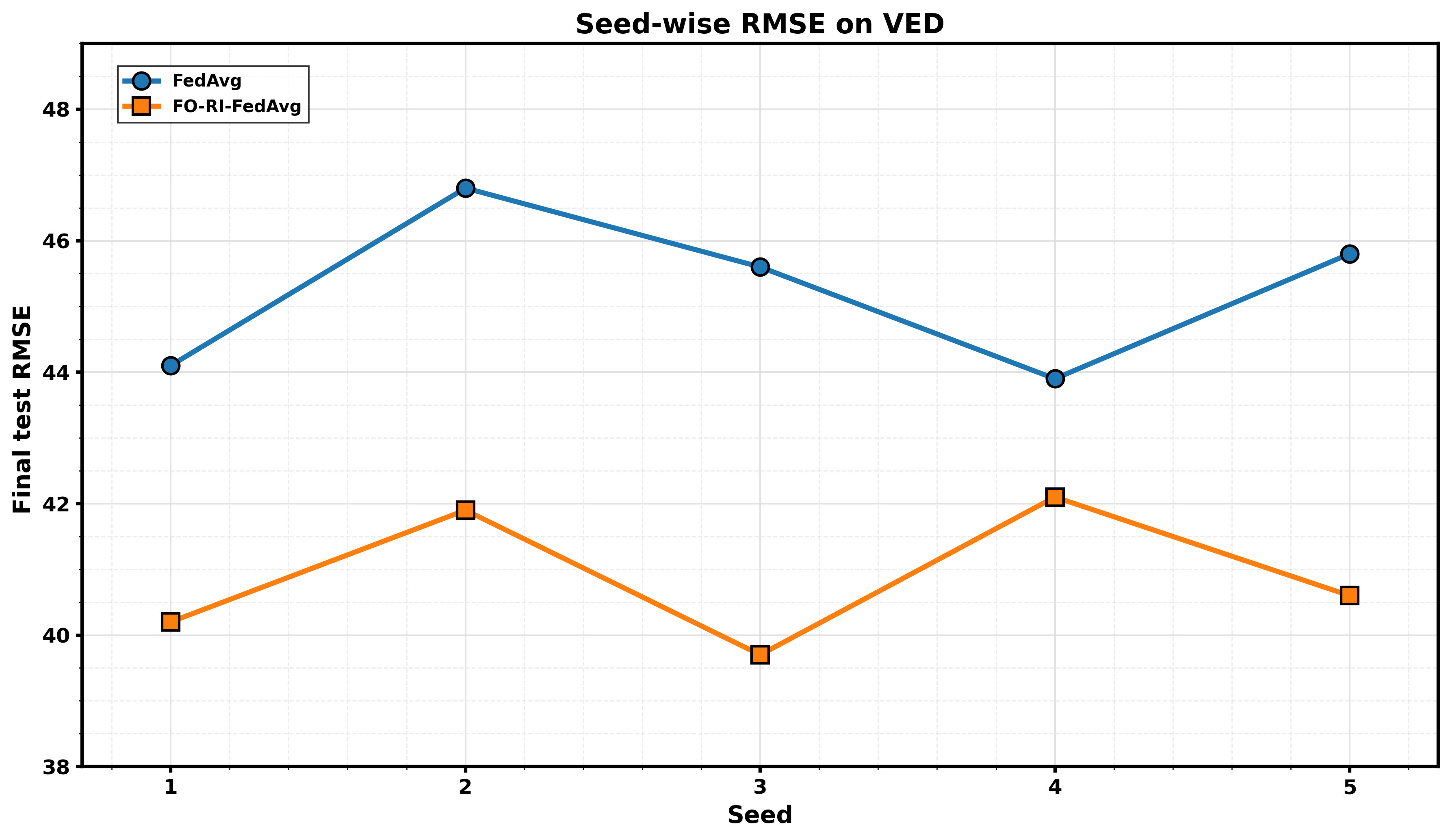}
\caption{Seed-wise final RMSE on VED for FedAvg vs FO-RI-FedAvg.}
\label{fig:seed_rmse_ved}
\end{figure}

\begin{figure}[!t]
\centering
\includegraphics[width=3.5in]{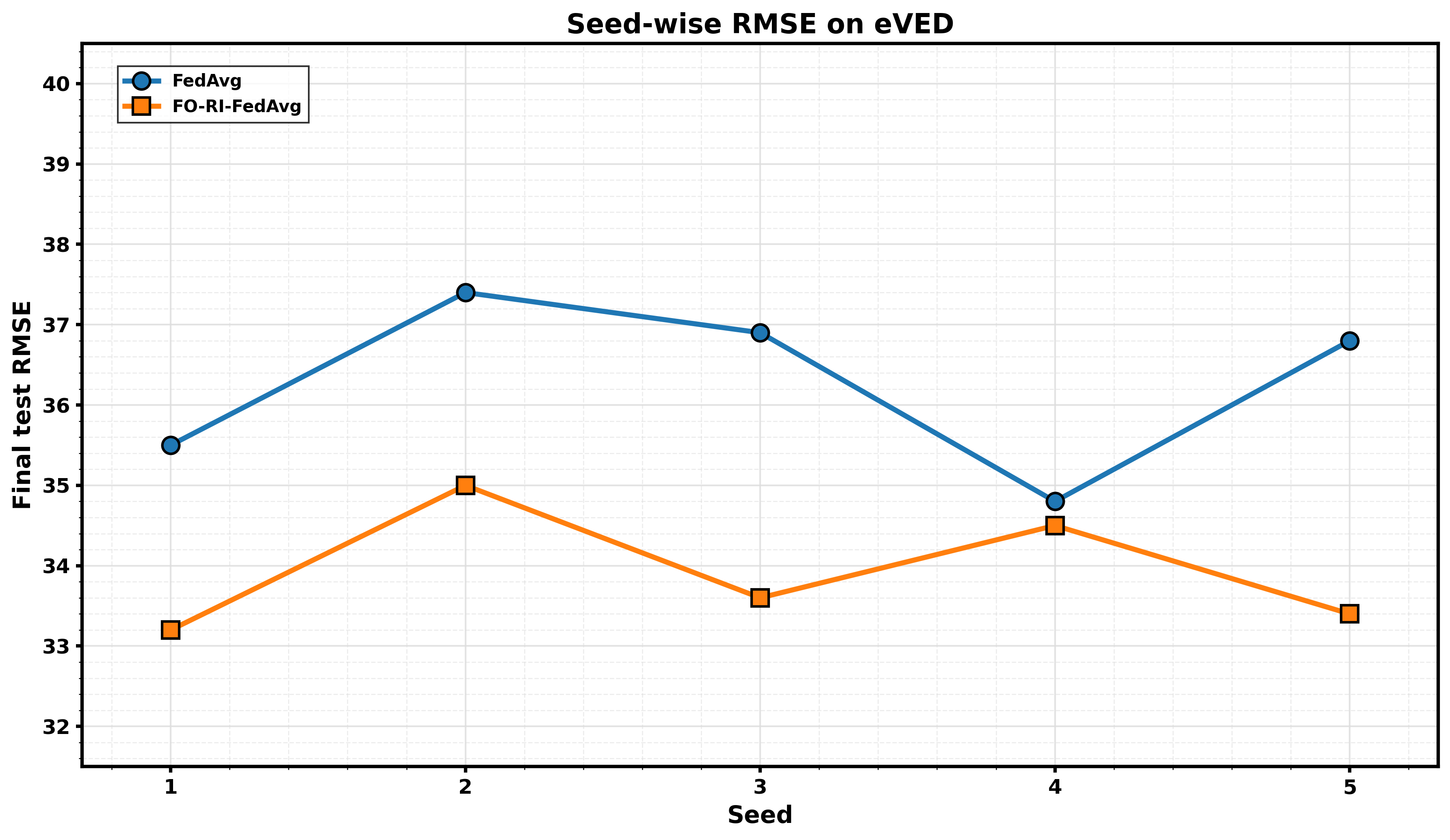}
\caption{Seed-wise final RMSE on eVED for FedAvg vs FO-RI-FedAvg.}
\label{fig:seed_rmse_eved}
\end{figure}

\paragraph{Seed-wise stability (RMSE).}
Figures~\ref{fig:seed_rmse_ved}--\ref{fig:seed_rmse_eved} provide seed-wise final RMSE values for FedAvg and FO-RI-FedAvg,
demonstrating consistent improvements across runs on both VED and eVED.

\paragraph{Minimal run recipe (for implementation alignment).}
A compact run configuration consistent with the above is:
\begin{small}
\noindent\texttt{
K=100, C=0.3, T=300, E=1 (or H equiv.), B=64\\
FO-RI-FedAvg: alpha=0.8, delta=1e-6, eta0=0.05, eta\_t=eta0/sqrt(t+1), lambda=0.1\\
Roughness: M=10, ell=0.01, m=100, B\_probe=128, probe\_every=5\\
Heterogeneity (controlled): regime\_skew=moderate\\
Seeds: \{1,2,3,4,5\}
}
\end{small}


\section{Limitations and Future Work}
\label{sec:limitations_futurework}

While FO-RI-FedAvg improves stability and utility under realistic BEV-fleet heterogeneity, several limitations remain and motivate future work.

\paragraph{(L1) Diagnostic overhead and probing schedule sensitivity.}
FO-RI-FedAvg introduces a training-time diagnostic cost due to roughness probing (Section~\ref{subsec:method_diagnostics}).
Although the overhead can be amortized by probing every \(R_{\mathrm{probe}}\) rounds (Section~\ref{subsec:roughness_overhead}),
the method still exhibits a tunable trade-off between diagnostic freshness and wall-clock cost.
Future work includes designing \emph{adaptive probe schedules} that trigger probing only when drift indicators exceed a threshold,
and developing \emph{low-variance estimators} of \(\mathcal{I}_k\) that reduce the required number of probes \(M(m+1)\).

\paragraph{(L2) Hyperparameter coupling across datasets and regimes.}
FO-RI-FedAvg contains several interacting hyperparameters (e.g., \(\alpha\), \(\delta\), clipping bounds, \(\lambda\),
and roughness-probe parameters \(M,m,\ell,B_{\mathrm{probe}}\)).
Although ablations demonstrate reasonable default choices (Section~\ref{subsec:ablations}),
optimal settings can shift with the dataset (VED vs.\ eVED), participation level, and heterogeneity severity.
A promising direction is to learn \emph{closed-loop controllers} that adapt \(\alpha\) and \(\lambda_t\) online based on measured drift,
roughness, or server-side stability signals, reducing manual tuning.

\paragraph{(L3) Roughness as a proxy: scope and potential failure cases.}
The roughness index \(\mathcal{I}_k\) is designed as a lightweight proxy for client-side instability, and mechanistic evidence
supports its association with drift (Section~\ref{subsec:mech_roughness_drift}).
However, roughness may not fully capture all sources of harmful updates, such as systematic label noise, adversarial clients,
or rare-event distribution shifts.
Future work should investigate combining \(\mathcal{I}_k\) with complementary signals
(e.g., update norm outliers, cosine agreement, or influence-style diagnostics) for more robust client-side control.

\paragraph{(L4) System constraints beyond computation: communication and stragglers.}
Our overhead analysis focuses on compute-time costs (Sections~\ref{subsec:overhead_scalability}--\ref{subsec:roughness_overhead}),
while communication remains FedAvg-like in asymptotic complexity.
In practice, BEV fleets may exhibit bandwidth constraints, straggler effects, and delayed updates.
A natural extension is to integrate FO-RI-FedAvg with \emph{asynchronous} or \emph{buffered} aggregation mechanisms,
and to study the interaction between fractional memory and delayed client contributions under realistic wireless scheduling.

\paragraph{(L5) Broader evaluation and task generalization.}
This work focuses on BEV energy regression with VED/eVED and a fixed window-based forecasting formulation.
Although architecture robustness is explored (Section~\ref{subsec:arch_generalization}),
future work should extend evaluation to additional predictive tasks (e.g., charge depletion forecasting, driver-style clustering),
alternative temporal resolutions, and different sensing modalities.
It is also valuable to validate the approach on additional real-world federated telemetry corpora
to assess robustness under unseen fleet compositions and nonstationary seasonal effects.

\paragraph{(L6) Privacy, security, and combined constraints.}
FO-RI-FedAvg is compatible with standard FL privacy boundaries (raw data remains on-device),
but it is not inherently a formal privacy mechanism.
An important research direction is combining FO-RI-FedAvg with rigorous privacy protections
(e.g., client-level DP mechanisms) and analyzing how fractional dynamics and roughness-aware control
interact with gradient clipping and noise injection.
Similarly, robustness under malicious or Byzantine clients is not explicitly addressed here and remains open.

\paragraph{Summary.}
Overall, FO-RI-FedAvg demonstrates a favorable utility--stability--overhead trade-off for connected BEV fleets,
but future work should focus on adaptive diagnostics, reduced tuning burden, broader nonstationary system modeling,
and tighter integration with privacy and security constraints.

\section{Conclusion}
\label{sec:conclusion}

 This work introduced \emph{Fractional-Order Roughness-Informed Federated Averaging (FO-RI-FedAvg)}, a unified approach that targets this instability by combining two complementary mechanisms: (i) \emph{roughness-informed regularization}, where each client adaptively adjusts the strength of a proximal pull toward the global model using a locally estimated indicator of loss-landscape roughness, and (ii) \emph{fractional-order local optimization}, where clients incorporate short-term memory of recent parameter changes to smooth noisy or conflicting update directions. Importantly, FO-RI-FedAvg preserves the standard FedAvg server pipeline and requires only lightweight client-side additions, with roughness estimation amortizable across rounds and element-wise fractional scaling that integrates cleanly into existing training code.

From a practical standpoint, FO-RI-FedAvg is attractive because it improves stability without introducing complex control variates, extra server state, or architectural changes. The method is modular: the roughness-based scaling can be used alongside other regularizers, and the fractional memory component can be enabled or disabled depending on resource constraints and the severity of heterogeneity. From a theoretical standpoint, the framework offers an analysis-ready lens that connects three quantities that are often treated separately in federated optimization: loss roughness (as a proxy for local sensitivity), client drift (as a driver of instability), and memory effects (as a stabilizer of update dynamics).

In summary, FO-RI-FedAvg provides a principled, lightweight, and compatible modification to FedAvg that improves stability under heterogeneity by jointly leveraging adaptive roughness-aware regularization and memory-augmented local updates, while offering a coherent framework for analyzing how these mechanisms reduce drift and stabilize federated training.

\section*{Acknowledgment}
The authors utilized AI tools to enhance the grammar and readability of this work.

\end{document}